\documentclass[10pt, bookmarks=true]{article} 
\usepackage[preprint]{tmlr}

\usepackage{amsmath,amsfonts,bm}

\def\eqref#1{equation~\ref{#1}}

\def\1{\bm{1}}

\DeclareMathAlphabet{\mathsfit}{\encodingdefault}{\sfdefault}{m}{sl}
\SetMathAlphabet{\mathsfit}{bold}{\encodingdefault}{\sfdefault}{bx}{n}

\usepackage{hyperref}
\usepackage{url}

\usepackage{graphicx} 
\usepackage{multicol}
\usepackage{booktabs} 
\usepackage{caption} 
\usepackage{subfigure}
\usepackage{amsmath}
\usepackage{amsfonts}

\usepackage{booktabs} 
\usepackage{array}    
\usepackage{tabularx}

\usepackage{array}
\usepackage{multirow}
\usepackage{siunitx}  

\usepackage{wrapfig}
\usepackage{subcaption}

\usepackage{hyperref}
\usepackage{xcolor}    
\usepackage{xspace}    

\usepackage{algorithmic}
\definecolor{pinkyred}{RGB}{255, 105, 180}
\usepackage{fvextra} 
\usepackage[utf8]{inputenc}
\usepackage{tcolorbox}
\tcbuselibrary{breakable}
\usepackage[linesnumbered,ruled,vlined]{algorithm2e}
\usepackage{amssymb}

\title{LAPP: Large Language Model Feedback for Preference-Driven Reinforcement Learning}

\author{}

\def\month{MM}  
\def\year{YYYY} 
\def\openreview{\url{https://openreview.net/forum?id=XXXX}} 

\begin{document}

\maketitle

\vspace{-4.5em}
\begin{center}
\textbf{Pingcheng Jian \quad Xiao Wei \quad Yanbaihui Liu \quad Samuel A. Moore \quad Michael M. Zavlanos \quad Boyuan Chen} \\[0.5ex]
Duke University \\[0.5ex]
\textcolor{orange}{\href{http://www.generalroboticslab.com/LAPP}{www.generalroboticslab.com/LAPP}}
\end{center}
\vspace{0.5em}

\begin{abstract}
We introduce Large Language Model-Assisted Preference Prediction (LAPP), a novel framework for robot learning that enables efficient, customizable, and expressive behavior acquisition with minimum human effort. Unlike prior approaches that rely heavily on reward engineering, human demonstrations, motion capture, or expensive pairwise preference labels, LAPP leverages large language models (LLMs) to automatically generate preference labels from raw state-action trajectories collected \textit{during} reinforcement learning (RL).  These labels are used to train an online preference predictor, which in turn guides the policy optimization process toward satisfying high-level behavioral specifications provided by humans. Our key technical contribution is the integration of LLMs into the RL feedback loop through trajectory-level preference prediction, enabling robots to acquire complex skills including subtle control over gait patterns and rhythmic timing. We evaluate LAPP on a diverse set of quadruped locomotion and dexterous manipulation tasks and show that it achieves efficient learning, higher final performance, faster adaptation, and precise control of high-level behaviors. Notably, LAPP enables robots to master highly dynamic and expressive tasks such as quadruped backflips, which remain out of reach for standard LLM-generated or handcrafted rewards. Our results highlight LAPP as a promising direction for scalable preference-driven robot learning.

\end{abstract}

\vspace{-10pt}
\begin{figure}[htbp]
  \centering
  \includegraphics[width=0.90\linewidth]{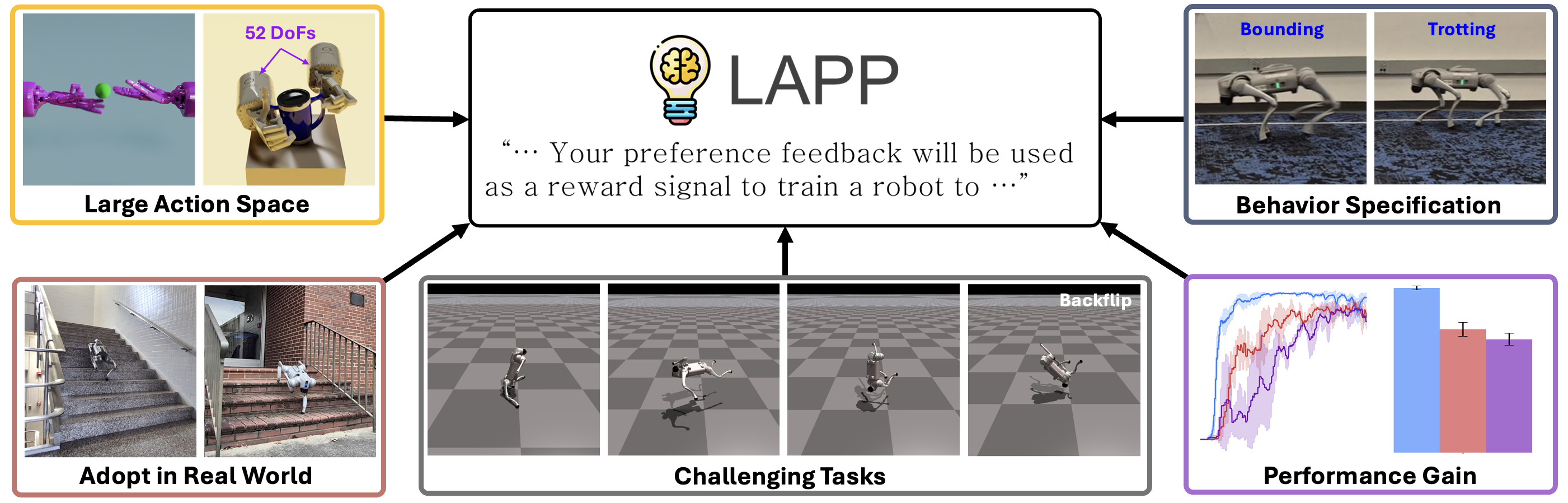}
  \caption{\textbf{Large Language Model-Assisted Preference Prediction (LAPP)} takes in language behavior instructions and generates preference feedback to guide reinforcement learning training from raw state-action robot trajectories.
  }
  \vspace{-15pt}
  \label{fig:teaser}
\end{figure}

\section{Introduction}
\label{introduction}

Designing effective reward functions remains a fundamental challenge in training robots with reinforcement learning (RL) \citep{ratner2018simplifying, dang2023clip, eschmann2021reward, sorg2010reward, evans2021reward, ng1999policy, grzes2008plan, devlin2012dynamic}. Reward functions define the objectives and constraints of the learning process, but are often hand-crafted through trial and error, a process that is labor-intensive, error-prone, and difficult to scale. Poorly designed rewards can lead to suboptimal or unsafe behaviors, making robust and expressive policy learning especially difficult in real-world robotic systems.

Several alternatives have been proposed to reduce this  burden. Inverse RL infers reward functions from expert demonstrations \citep{arora2021survey, ng2000algorithms, abbeel2004apprenticeship, zhou2021inverse}, but requires significant human effort and data collection. More recently, large language models (LLMs) and vision-language models (VLMs) have been used to automate aspects of reward design \citep{ma2023eureka, yu2023language, xie2024textreward, yu2024few, wang2024rl}. These methods typically refine reward functions by analyzing task descriptions or environment code. While promising, they often fail to capture nuanced and high-level behavioral qualities, such as rhythmic locomotion or expressive timing, which are hard to specify with explicit reward terms.

Another line of work bypasses explicit reward engineering by learning from human or AI preferences over trajectory pairs \citep{christiano2017deep, early2022non, kim2023preference, wang2024rl, venkataraman2024real}. By querying which of two behaviors is preferable, these methods convert preferences into supervision for training reward models. However, preference queries can be costly and cognitively demanding for humans, and recent VLM-based methods remain limited to relatively simple tasks with low-dimensional action spaces. These approaches also typically assume a Markovian decision process, which may not hold for long-horizon and high-dimensional control tasks.

In this work, we propose LLM-Assisted Preference Prediction (LAPP) (Fig.~\ref{fig:teaser}), a novel framework that enables robots to learn efficient, customizable, and expressive behaviors from human language specifications. The core idea of LAPP is to leverage LLMs to generate preference labels on full state-action trajectories, which are then used to train an online transformer-based reward predictor. This predictor produces dense, trajectory-informed reward estimates that guide policy optimization and are continually updated with new policy training and rollouts. Unlike prior approaches, LAPP integrates the LLM-generated feedback directly into the RL loop, enabling closed-loop refinement of learned behaviors.

We evaluate LAPP across a suite of challenging control tasks, including quadruped locomotion and dexterous manipulation with up to 52-dimensional action spaces. LAPP not only accelerates training and improves final performance compared to state-of-the-art baselines, but also enables nuanced control over high-level behavior attributes, such as gait symmetry, timing, and cadence, through simple languag inputs. LAPP also enables faster adaptation to unseen environmental conditions. Notably, LAPP successfully solves exploration-heavy tasks such as quadruped backflips, which were previously infeasible with human-designed, LLM-generated, or VLM-derived reward functions.

We summarize our key contributions as follows:
\begin{enumerate}
    \item A novel learning framework (LAPP) that uses LLM-generated preference feedback over state-action trajectories to guide reinforcement learning with language behavior instructions.
    \item A transformer-based online preference predictor that models trajectory-level feedback and integrates it as dense supervision into the RL policy learning loop through iterative policy \& reward model improvements.
    \item Empirical results showing the LAPP outperforms baselines in training speed, final performance, adaptation efficiency, and controllability of high-level behaviors across complex robotic tasks.
    \item Ablation studies dissecting architectural and algorithm design choices to reveal how trajectory-level modeling and online reward model updates contribute to LAPP's success.
    
\end{enumerate}
\section{Related Work}
\label{related_work_sec}
\textbf{Foundation Models for Robotics.}  Recent advances in foundation models have spurred applications in robotic action generation \citep{octo_2023, szot2023large, zitkovich2023rt, tang2023saytap}, simulation \citep{Genesis}, task planning \citep{lin2023text2motion, ahn2022can, singh2023progprompt, huang2024grounded, zhang2023bootstrap, hao2023reasoning, liu2023llm+, ha2023scaling, wang2023describe, wang2023voyager, ding2023task, silver2024generalized, xie2020translating, huang2023instruct2act, liang2023code}, and sim-to-real transfer \citep{ma2024dreureka}. A growing line of work focuses on using language or vision-language models to automate aspects of reward engineering \citep{ma2023eureka, yu2023language, xie2024textreward, yu2024few} or generate training environments \citep{liang2024eurekaverse, wang2023gen, wang2023robogen, faldor2024omni, wang2023gensim}. However, these approaches remain limited in specifying high-level behaviors, handling hard exploration challenges, and scaling to high-dimensional action spaces.

\textbf{Reward Signal Design for Challenging Robotic Tasks.} Reward design is a crucial component of RL \citep{ratner2018simplifying, dang2023clip, eschmann2021reward, sorg2010reward, evans2021reward, xia2024duke}. To address sparse reward issues, prior works explore reward shaping \citep{grzes2008plan, devlin2012dynamic, devidze2022exploration, marom2018belief, hu2020learning, gupta2022unpacking, grzes2017reward, zou2019reward, goyal2019using, memarian2021self, hussein2017deep, ng1999policy}. However, complex agile motions, such as rapid locomotion \citep{margolis2024rapid} and backflips \citep{tang2021learning, kim2024stage}, remain difficult to learn with a single reward function.

Reward design remains a bottleneck for complex RL tasks, especially when rewards are sparse, brittle, or difficult to engineer. Prior efforts explore reward shaping \citep{grzes2008plan, devlin2012dynamic, devidze2022exploration, marom2018belief, hu2020learning, gupta2022unpacking, grzes2017reward, zou2019reward, goyal2019using, memarian2021self, hussein2017deep, ng1999policy}, curriculum learning \citep{tang2021learning, margolis2024rapid, ryu2024curricullm}, and multi-objective optimization \citep{kim2024stage, kyriakis2022pareto, van2013scalarized, basaklar2022pd, xu2020prediction, cai2024distributional, abdolmaleki2020distributional, yang2019generalized, hayes2022practical, huang2022constrained}. Inverse RL methods aim to infer reward signals from demonstrations \citep{arora2021survey, ng2000algorithms, hadfield2016cooperative, zakka2022xirl, brown2018risk, kumar2023graph, das2021model, abbeel2004apprenticeship, zhou2021inverse}, but require curated expert data.

While recent works attempt to automate reward or curriculum generation using LLMs \citep{ma2023eureka, liang2024eurekaverse}, they still depend on explicit reward decompositions or low-level state supervision which can be difficult to obtain for complex tasks and high-level behavior specifications. Our method complements these advances by using LLMs to generate implicit preference feedback. As our experiments show, our method achieves the best performance when combined with previous state-of-the-art reward designs.

\textbf{Human-Guided Machine Learning.} Integrating human guidance into machine learning has been widely explored to improve training efficiency and model performance \citep{amershi2014power, gil2019towards, wu2022survey, zhang2019leveraging}. Various methods incorporate human demonstrations \citep{pomerleau1988alvinn, schaal1996learning}, instructions \citep{zhou2021inverse, saran2021efficiently}, and corrections \citep{chai2020human, ji2024enabling} to enhance imitation learning \citep{pomerleau1988alvinn, schaal1996learning, saran2021efficiently} or inverse RL \citep{abbeel2004apprenticeship, zhou2021inverse}. Other works model human feedback as reward functions \citep{knox2008tamer, warnell2018deep} or advantage functions \citep{macglashan2016convergent, arumugam2019deep} to guide RL. Recent advancements extend these algorithms to continuous action spaces \citep{sheidlower2022environment}, multi-agent scenarios \citep{ji2024enabling}, and real-time human feedback \citep{zhang2024crew, zhang2024guide}.

The most relevant works to ours are those that learn from human preferences \citep{wirth2017survey, akrour2011preference, daniel2015active, furnkranz2012preference, ibarz2018reward, wilson2012bayesian, wirth2016model, kim2024preference, dong2023aligndiff, liu2024pearl, aroca2024predict, akrour2012april, liu2020learning, lee2021pebble, knox2022models, ouyang2022training, park2022surf, verma2022symbol, christiano2017deep}, with applications in LLM fine-tuning \citep{brown2020language}, summarization \citep{wu2021recursively}, browser-assisted question answering \citep{nakano2021webgpt}, robotic manipulation \citep{hejna2023few}, and locomotion \citep{yuan2024preference}. \citet{yu2024few} uses human preference to select the reward functions generated by a LLM, while the human preference is not directly predicted as a preference reward to guide the policy optimization.

Our work builds on reinforcement learning from human feedback (RLHF) \citep{christiano2017deep}, where human preferences are used to train MLP-based preference predictors for Markovian rewards. Later works extend this to non-Markovian settings with LSTMs \citep{early2022non} and importance-weighted rewards using Preference Transformers \citep{kim2023preference}. However, RLHF methods require extensive human annotation, with human annotators evaluating thousands of trajectory pairs. Recent research proposes Reinforcement Learning from AI Feedback (RLAIF) \citep{bai2022constitutional, leerlaif, wang2024rl, venkataraman2024real}, replacing human annotators with AI models. However, these approaches are limited to Markovian rewards and have only been tested in low-dimensional robotic tasks. Our work not only reduces annotation cost but allows for preference-driven RL in more complex task domains than those explored in existing RLHF or RLAIF frameworks.

\textbf{Preference Feedback for Robot Learning.} Learning from human or AI preferences has emerged as an alternative to explicit reward design \citep{christiano2017deep, early2022non, kim2023preference, yuan2024preference}. These methods train reward models using preference labels over trajectory pairs, typically annotated by humans. While effective, annotation costs remain high.

More recent works adopt AI-generated feedback in place of human raters, such as RL-VLM-F \citep{wang2024rl, venkataraman2024real}, which uses vision-language models to rank state images. However, such models operate under a fixed preference criterion throughout the entire policy learning process and assume Markovian rewards, limiting them to relatively simple low-DoF tasks like CartPole or tabletop manipulation.

LAPP offers advancements in this direction in several key aspects. First, LAPP is the first work to operate on raw state-action trajectories to provide effective preference feedback from LLMs. This method avoids reliance on vision-based snapshots to query VLMs, which currently come with much higher costs than LLMs and still do not yet demonstrate strong reasoning capabilities, hence limiting the task complexity they can solve. Second, LAPP models both Markovian and non-Markovian preference rewards using transformer architectures to allow reasoning over long temporal sequences.

Moreover, LAPP enables dynamic preference shaping by prompting LLMs to evolve their evaluation criteria as training progresses, while prior works \citep{wang2024rl, venkataraman2024real} rank the states with a static standard. To our knowledge, LAPP is the first method to fully automate preference alignment via LLMs for training policies in complex and high-dimensional tasks, including quadruped backflips and dexterous hand manipulation.

\begin{figure*}[!t]
  \centering
  \includegraphics[width=0.93\linewidth]{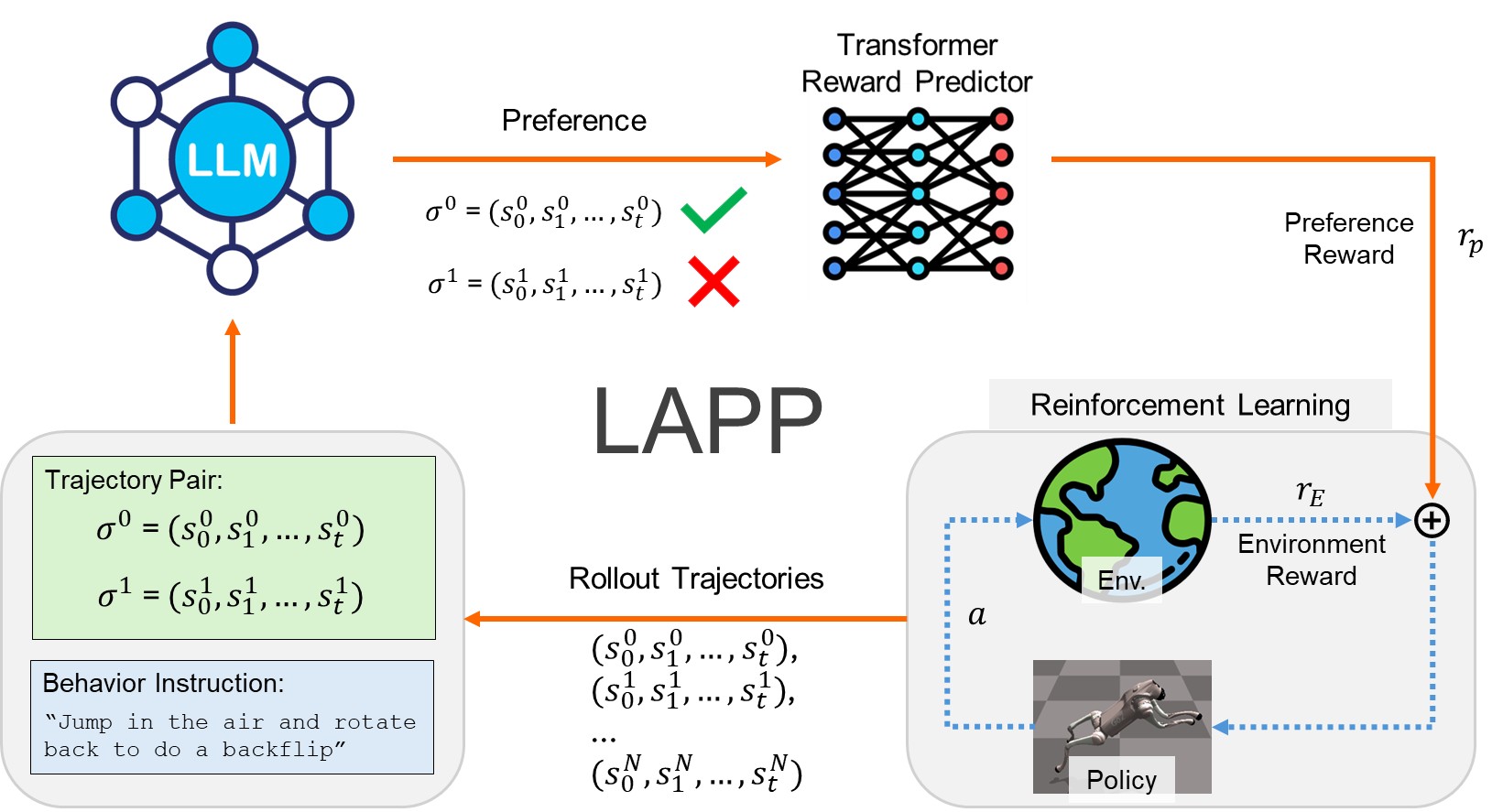}
  \caption{\textbf{LAPP} generates preference feedback from an LLM based on rollout trajectories pairs of raw state and actions as well as a high-level behavior instruction. A transformer-based reward predictor is trained using these preferences while simultaneously optimizing a robot policy to maximize a weighted sum of environment rewards and predicted preference rewards.}
  \label{fig:method}
  \vspace{-10pt}
\end{figure*}

\section{Preliminaries}
\label{preliminaries}

We consider an agent interacting with an environment over a sequence of discrete timesteps in a RL framework \citep{sutton2018reinforcement}. At each timestep $t$, the agent receives an observation $o_t$ of the current state $s_t$ and selects an action $a_t$ based on its policy $\pi$. The environment then provides a reward $r_t$ based on the pre-designed reward functions and transitions the agent to the next state $s_{t+1}$. The goal of RL is to optimize $\pi$ to maximize the expected return $\mathcal{R}_t=\sum_{k=0}^{\infty} \gamma^k r_{t+k}$ where $\gamma$ is the discount factor.

However, designing a reward function that precisely captures high-level behavioral objectives or human preferences remains challenging. To address this, preference-based RL learns a reward model that predicts human preferences instead of relying on manually defined rewards. In this setting, we consider a pair of trajectory segments $(\sigma^0, \sigma^1)$ with length $H$: $\sigma = \{ (s_1, a_1), ..., (s_H, a_H)\}$. A preference relation $\sigma^i \succ \sigma^j$ indicates that segment $\sigma^i$ is preferable over segment $\sigma^j$. Given a pair $(\sigma^0, \sigma^1)$, a human or AI provides a preference label $y \in\{0,1,0.5\}$:
 \[
y = 
\begin{cases} 
0 & \text{, } \sigma^0 \succ \sigma^1\\
1 & \text{, } \sigma^1 \succ \sigma^0\\
0.5 & \text{, } \sigma^0 \text{ and } \sigma^1 \text{ are equally preferable}
\end{cases}
\]
The preference judgments are recorded in a dataset $\mathcal{D}$ of labeled preference triples $(\sigma^0, \sigma^1, y)$.

To obtain a preference-based reward model $\hat{r}$, prior works \citep{christiano2017deep, ibarz2018reward, lee2021pebble, lee2021b, hejna2023few, park2022surf} adopt the Bradley-Terry model \citep{bradley1952rank}, assuming Markovian rewards (i.e. the reward depends only on the current state and action). The probability of preferring one segment over another is modeled as:
\begin{small}
\begin{equation}
\label{eq:markovian_preference}
\hat{P}\left[\sigma^1 \succ \sigma^2\right]=\frac{\exp \sum \hat{r}\left(s_t^1, a_t^1\right)}{\exp \sum \hat{r}\left(s_t^1, a_t^1\right)+\exp \sum \hat{r}\left(s_t^2, a_t^2\right)}
\end{equation}
\end{small}
However, Markovian rewards struggle with long-horizon tasks where preferences depend on past trajectories rather than only the current state-action pair. To address this, recent works \citep{kim2023preference, early2022non} propose non-Markovian rewards, where $\hat{r}$ considers the full preceding sub-trajectory segment $\left\{\left(\mathbf{s}_i, \mathbf{a}_i\right)\right\}_{i=1}^t$:
\begin{small}
\begin{equation}
\label{eq:non_markovian_preference}
\hat{P}[\sigma^1 \succ \sigma^0 ]=\frac{\exp \hat{r}(\{(\mathbf{s}_i^1, \mathbf{a}_i^1)\}_{i=1}^t )}{\sum_{j \in\{0,1\}} \exp \hat{r}(\{(\mathbf{s}_i^j, \mathbf{a}_i^j)\}_{i=1}^t)}
\end{equation}
\end{small}
The reward predictor $\hat{r}$ is then trained via supervised learning to fit the dataset $\mathcal{D}$ by minimizing the cross-entropy loss:
\begin{small}
\begin{equation}
\label{eq:ce_loss}
\mathcal{L}^{CE}(\hat{r}) = -\sum_{(\sigma^1, \sigma^2, y) \in \mathcal{D}} (1-y) \log \hat{P}\left[\sigma^0 \succ \sigma^1\right] + y \log \hat{P}\left[\sigma^1 \succ \sigma^0\right]
\end{equation}
\end{small}

To mitigate the noisy LLM outputs, We assume that the LLM has $\epsilon=15\%$ of chance to provide preference feedback uniformly at random. Therefore, the adjusted preference probability is:
\begin{small}
\begin{equation}
\hat{P}^{\prime}\left[\sigma^0 \succ \sigma^1\right]=(1-\epsilon) \hat{P}\left[\sigma^0 \succ \sigma^1\right]+\epsilon \cdot 0.5,
\end{equation}
where $\epsilon=0.15$ is the error rate. Consequently, We have:
\begin{equation}
\hat{P}^{\prime}\left[\sigma^1 \succ \sigma^0\right]=1-\hat{P}^{\prime}\left[\sigma^0 \succ \sigma^1\right].
\end{equation}
\end{small}
Therefore, the adjusted cross-entropy loss becomes:
\begin{small}
\begin{equation}
\label{eq:adjusted_ce_loss}
\begin{aligned}
\mathcal{L}_\epsilon^{C E}(\hat{r}) 
& =-\sum_{\left(\sigma^0, \sigma^1, y\right) \in \mathcal{D}}\left[\left(1-y\right) \log \hat{P}^{\prime}\left[\sigma^0 \succ \sigma^1\right] \right. + \left. y \log \hat{P}^{\prime}\left[\sigma^1 \succ \sigma^0\right] \right] \\
& =-\sum_{\left(\sigma^0, \sigma^1, y\right) \in \mathcal{D}}\left[\left(1-y\right) \log \left((1-\epsilon) \hat{P}\left[\sigma^0 \succ \sigma^1\right]+\epsilon \cdot 0.5\right) \right. \\
&+ \left. y \log \left(1-\left((1-\epsilon) \hat{P}\left[\sigma^0 \succ \sigma^1\right]+\epsilon \cdot 0.5\right)\right) \right].
\end{aligned}
\end{equation}
\end{small}

Once trained, the reward predictor $\hat{r}$ can be used to guide policy optimization, where an RL algorithm maximizes the expected return from the learned preference rewards.
\section{LLMs-Assisted Preference Prediction}
\label{method}

LAPP is a novel framework that enables preference-driven RL by integrating LLM-generated feedback into the policy training loop. It consists of three main components: 1) \textbf{Behavior Instruction:} a prompting strategy to elicit trajectory preferences from LLMs given language description of task objectives and preferred behaviors; 2) \textbf{Preference Predictor Training:} an ensemble of transformer-based models that learn to predict preference rewards; and 3) \textbf{Preference-Driven Reinforcement Learning:} a robot policy is optimized using both environment rewards and predicted preference rewards. An overview of LAPP is shown in Fig.~\ref{fig:method}.

\subsection{Behavior Instruction: Generating Preference Labels from State-Action Trajectories}

Conventional RLHF frameworks rely on human annotators to label trajectory preferences. However, this process is labor intensive. To reduce this burden, LAPP replaces human annotators with LLMs by prompting them to generate preference labels for pairs of trajectory segments $\sigma^0$ and $\sigma^1$.

\begin{wrapfigure}[]{r}{0.5\textwidth}
\vspace{-12pt}
  \begin{center}
  \includegraphics[width=\linewidth]{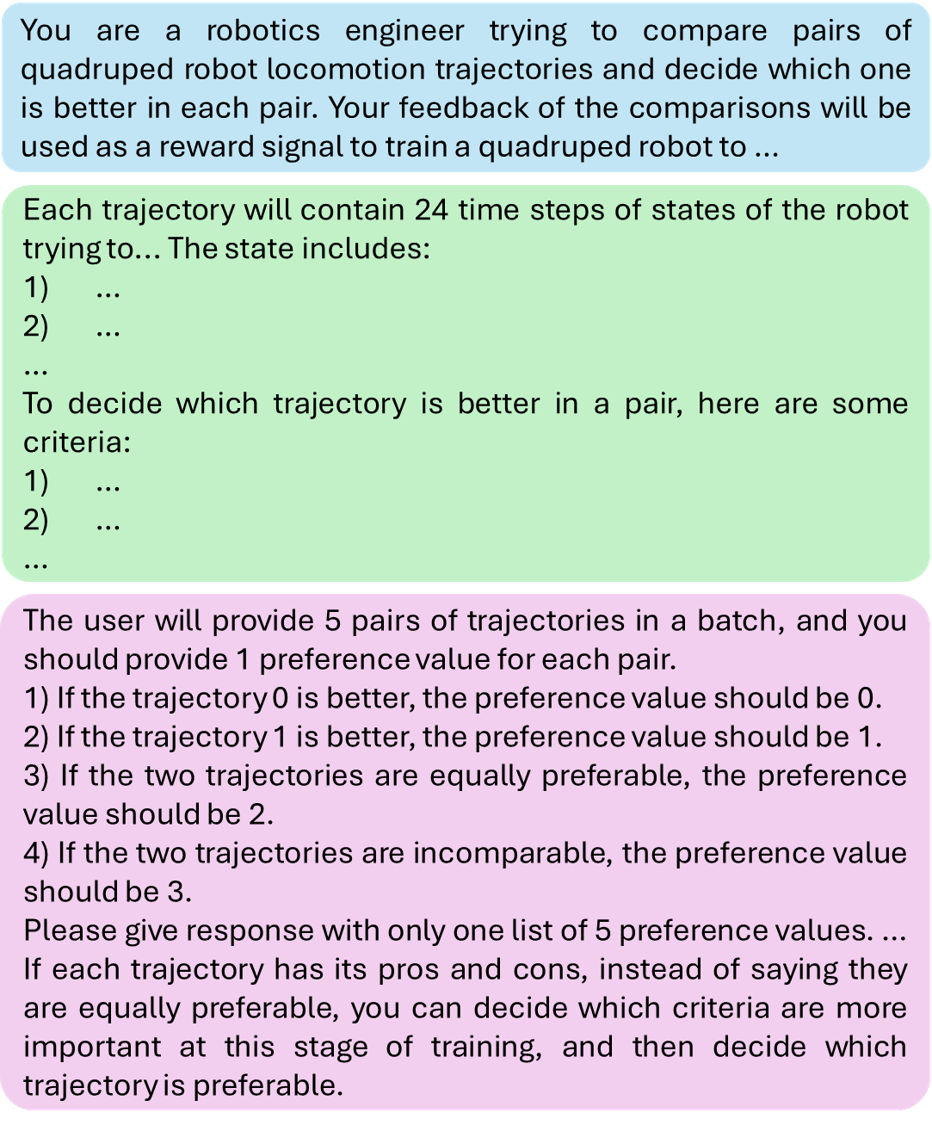}
  \end{center}
  \vspace{-13pt}
  \caption{\textbf{Behavior Instruction Prompt Example.} The LLM prompt consists of three sections: (1) defining the LLM's role and the robotic task (blue box), (2) specifying the state variables and some evaluation criteria of preference (green box), and (3) establishing rules and semantics for generating preference labels (purple box).
  }
  \vspace{-10pt} 
  \label{fig:behavior_instruction_example}
  \vspace{-13pt}
\end{wrapfigure}

Fig.~\ref{fig:behavior_instruction_example} illustrates an example of a behavior instruction prompt. The first part defines the LLM's role the robot's goal and desired behavior (e.g., ``walk forward with a bounding gait''). The second part provides numerical values and their descriptions of each trajectory (e.g., base velocity, orientation, foot contacts). The third part defines how preferences should be evaluated and formatted.

Unlike prior work that uses video clips \citep{christiano2017deep, kim2023preference} for human annotation, we feed LLMs structured numerical state-action logs, since the current multimodal foundation models such as GPT-4o lacks fine-grained video understanding with high costs and slow responses.

Notably, to enhance learning efficiency, we encourage the LLM to refer to our provided evaluation criteria and generate \textit{adaptive evaluation criteria, allowing the LLM to dynamically adjust its preferences as training progresses}. For instance, in quadruped locomotion, early-stage training should prioritize learning to stand, followed by developing stable movement, and ultimately refining gait patterns and command adherence. Instead of providing these stages explicitly by humans, our prompts ask the LLM to actively decide the important factors for different training stages by itself.

LAPP supports batched labeling of five trajectory pairs per prompt which can significantly reduce API latency and token costs. the output consists of preference labels in ${0, 1, 2, 3}$, indicating whether trajectory $\sigma^0$ is better, worse, equally preferable, or incomparable. To promote clear supervision, we encourage the LLM to avoid ambiguous labels. All labels are stored as triples $(\sigma^0, \sigma^1, y)$ in a growing preference dataset $\mathcal{D}$. Details of all state-action variables used for all tasks and full prompts can be found in Appendix \ref{apdx:prompt}.

\subsection{Preference Predictor Training: Modeling LLM Feedback}


\begin{figure}[ht]
\centering
\begin{minipage}{0.65\textwidth}
\begin{algorithm}[H]
\caption{\textsc{LAPP} \textnormal{- Preference Predictor Training}}\label{alg:lapp_preference_predictor_training}
\textbf{Require:} Ensemble of preference predictors $\{\hat{r}_i\}$, preference predictor training dataset $D_p$\\
\textbf{Hyperparameters:} Minimum iteration $N_{min}$, Maximum iteration $N_{max}$, pool of predictors number $M$, selected predictors number $C$, overfitting scale $\alpha$, LLM feedback error rate $\epsilon$.\\
\textcolor{pinkyred}{//Split into training and validation sets} \\
$D_p^{train}, D_p^{val} \leftarrow split(D_p)$ \\
$val\_loss\_list \leftarrow \left[\ \right]$\\
\For{$M$ predictors}{
Randomly initialize $\hat{r}_i$\\
\For{$m \leftarrow 0$ \KwTo $N_{max}-1$ epochs}{
\textcolor{pinkyred}{// Sample from $D_p^{train}$} \\
$(s_t^{train}, a_t^{train}) \sim D_p^{train}$ \\
\textcolor{pinkyred}{// Predict preference reward}\\
$r_t^{train} = \hat{r}_i(s_t^{train}, a_t^{train})$ \\
\textcolor{pinkyred}{// Train the preference predictor}\\
Calculate loss $\mathcal{L}^{CE}_{train}(r_t^{train})$ with Equation \ref{eq:adjusted_ce_loss} \\
$\hat{r}_i \leftarrow \mathrm{Adam}\Bigl(\hat{r}_i,\nabla_{\hat{r}_i}\mathcal{L}^{CE}_{train}(r_t^{train})\Bigr)$ \\
\textcolor{pinkyred}{// Sample from $D_p^{val}$} \\
$(s_t^{val}, a_t^{val}) \sim D_p^{val}$ \\
\textcolor{pinkyred}{// Predict preference reward}\\
$r_t^{val} = \hat{r}_i(s_t^{val}, a_t^{val})$ \\
Calculate loss $\mathcal{L}^{CE}_{val}(r_t^{val})$ with Equation \ref{eq:adjusted_ce_loss} \\
\If{$\mathcal{L}^{CE}_{val}(r_t^{val}) > \alpha \cdot \mathcal{L}^{CE}_{train}(r_t^{train})\ \textbf{and}\ n>N_{min}$}{
val\_loss\_list.append ($\mathcal{L}^{CE}_{val}(r_t^{val})$) \\
continue;
}
\If{$m == N_{max}-1$}
{val\_loss\_list.append ($\mathcal{L}^{CE}_{val}(r_t^{val})$) \\}
}
}
\textcolor{pinkyred}{// Select the C predictors with smallest validation losses} \\
\(\hat{r}_{i_1}, \hat{r}_{i_2}, ... , \hat{r}_{i_C} \leftarrow \arg\min_{\hat{r}_i \in \{\hat{r}_1,\dots,\hat{r}_M\}}^{C} val\_loss\_list[i]\) \\
\textcolor{pinkyred}{// Use the mean value of the selected predictors as the final predictor} \\
$\hat{r} = mean\left(\hat{r}_{i_1}, \hat{r}_{i_2}, ... , \hat{r}_{i_C}\right)$\\
\Return{$\hat{r}$}
\end{algorithm}
\end{minipage}
\end{figure}


LAPP models LLM-generated preferences using either Markovian or non-Markovian reward functions, depending on task complexity. For tasks like flat-ground locomotion, a Markovian reward model following the Bradley-Terry formulation (Eq.~\ref{eq:markovian_preference}) suffices. However, for more challenging tasks such as quadruped backflips or gait cadence control, a non-Markovian reward function (Eq.~\ref{eq:non_markovian_preference}) is necessary to capture long-term dependencies in behavior. Training the predictor for non-Markovian rewards requires additional computational resources, as it must process historical states to infer the reward at a given timestep. LAPP adopts the appropriate reward model based on task requirements to balance the preference prediction accuracy and the predictor training efficiency.

The preference dataset $\mathcal{D}_p = \{(\sigma^0, \sigma^1, y)\}$ is split into training ($\mathcal{D}_p^{train}$) and validation ($\mathcal{D}_p^{val}$) sets at a $9:1$ ratio. We maintain an ensemble of $M$ preference predictor networks, each trained to minimize the cross-entropy loss (Eq.~\ref{eq:ce_loss}). To prevent overfitting, training stops early if the validation loss exceeds $\alpha$ times the training loss and the training has gone through a minimum number of iterations $N_{\textrm{min}}$. If no early stopping is triggered, the training will finish after $N_{\textrm{max}}$ iterations. After training all $M$ predictors, we select the top $C$ models with the lowest validation losses and compute the final preference reward as their ensemble average. In practice, we set $M=9$, $C=3$, $N_{\textrm{min}}=30$, $N_{\textrm{max}}=90$, and $\alpha=1.3$. The full training procedure is detailed in Algorithm \ref{alg:lapp_preference_predictor_training}. This ensemble approach can help increase robustness to LLM label noise.

\subsection{Preference-Driven Reinforcement Learning}


\begin{figure}[ht]
\centering
\begin{minipage}{0.65\textwidth}
\begin{algorithm}[H]
\caption{\textsc{LAPP}}\label{alg:lapp}
\textbf{Require:} Robot behavior prompt $prompt$, preference generator LLM  $LLM$, environment $E$, policy $\pi$, preference predictor $\hat{r}$, preference predictor training dataset $D_p$, preference data buffer $B_p$ \\
\textbf{Hyperparameters:} Policy optimization epoch number $N$, preference predictors update interval epoch number $M$, per epoch trajectories pairs collection number $K$, per epoch rollouts number $S$, steps in each epoch $T$, preference reward scale $\beta$ \\
\textbf{Initialization:} Randomly initialize $\pi$, $\hat{r}$. $D_p \leftarrow \{ \text{zeros triple}_i \}_{i=1}^{|D_p|}$, $B_p \leftarrow \{ \text{zeros triple}_i \}_{i=1}^{M * K}$.\\
\textcolor{pinkyred}{// Collect initial preference dataset}\\
Rollout $\pi$ and sample $|D_p|$ trajectories pairs $\{ (\sigma^0_i, \sigma^1_i )\}^{|D_p|}_{i=1}$.\\
$\{y_i\}^{|D_p|}_{i=1} \sim \text{LLM}\left(\{ \sigma^0_i, \sigma^1_i \}^{|D_p|}_{i=1},\ prompt\right)$ \\
$D_p \leftarrow \{ (\sigma^0_i, \sigma^1_i, y_i) \}^{|D_p|}_{i=1}$\\
Update $\hat{r}$ with Algorithm \ref{alg:lapp_preference_predictor_training}.\\

$obs \sim E.reset()$ \textcolor{pinkyred}{// reset $E$, get initial observation} \\
\For{$i \leftarrow 0$ \KwTo $N-1$ epochs}{
\textcolor{pinkyred}{ //rollout $\pi$ in $E$ \\}
\For{ $T$ steps}{
$a \sim \pi(obs)$ \ \textcolor{pinkyred}{// sample action from policy}\\
$r_E \sim E(obs)$ \ \textcolor{pinkyred}{// get environment reward $r_E$}\\
$r_p \sim \hat{r}(obs)$ \ \textcolor{pinkyred}{// predict preference reward $r_p$}\\
$r = \beta \cdot r_p + r_E$ \ \textcolor{pinkyred}{// calculate weighted sum}\\
}
Update $\pi$ with PPO algorithm \citep{schulman2017proximal}\\
\textcolor{pinkyred}{// Sample $K$ pairs from $S$ rollouts} \\
$\{ (\sigma^0_k, \sigma^1_k) \}^{K}_{k=1} \sim \{ \sigma_s \}^{S}_{s=1}$ \\
\textcolor{pinkyred}{// Push into the preference data buffer} \\
Push $\{ (\sigma^0_k, \sigma^1_k, None) \}^{K}_{k=1}$ into $B_p$\\
\textcolor{pinkyred}{// Update the preference dataset \\}
\If{(i+1) \% M == 0}{
\textcolor{pinkyred}{// Generate preference labels} \\
$\{y_k\}^{M*K}_{k=1} \sim \text{LLM}\left(\{ (\sigma^0_k, \sigma^1_k) \}^{M*K}_{k=1},\ prompt\right)$ \\
\textcolor{pinkyred}{// Place preference labels into the buffer} \\
$B_p \leftarrow \{ (\sigma^0_k, \sigma^1_k, y_i) \}^{M*K}_{k=1}$ \\
\textcolor{pinkyred}{// Update preference dataset} \\
$D_p \leftarrow \{ \text{triple}_k | \text{triple}_k \in D_p \}^{|D_p|}_{k={M*K+1}} \cup B_p$\\
\textcolor{pinkyred}{// Update preference predictor} \\
Update $\hat{r}$ with Algorithm \ref{alg:lapp_preference_predictor_training}.\\
$B_p \leftarrow \{\}$\\
}
}
\Return{$\pi$}
\end{algorithm}
\end{minipage}
\end{figure}


LAPP continuously aligns robot behaviors with high-level task specifications throughout RL training by iteratively updating both the preference predictor and the policy network. Unlike previous RLHF approaches \citep{christiano2017deep, kim2023preference} that train static preference models, LAPP dynamically refines preferences during training.

Initially, the policy generates rollout trajectory pairs $\{ \sigma^0_i, \sigma^1_i \}$, which are evaluated by the LLM to generate preference labels $\{y_i\}$. To mitigate noisy outputs which could pose potential risks to training stability, we sample $15$ preference labels for each trajectory pair and calculate the mode of them as the final selected preference labels $\{y_i\}$. The labeled data is stored in an initial preference dataset $\mathcal{D}_p = \{ (\sigma^0_i, \sigma^1_i, y_i) \}^{|\mathcal{D}_p|}_{i=1}$, which is used to train the initial preference predictor via Algorithm \ref{alg:lapp_preference_predictor_training}.

During RL training, the reward at each timestep is computed as a weighted sum of the predicted preference reward $r_p$ and the environment reward $r_E$ defined by built-in explicit reward functions in our evaluation suites:
\vspace{-2pt}
\begin{small}
\begin{equation}
r = \beta r_p + r_E
\end{equation}
\end{small}
\vspace{-1pt}
where $\beta$ balances their contributions. We set the $\beta$ to be $1.0$ in all the tasks except the Backflip. The Backflip task has some reward items with large scales, so the $\beta$ is set to $50.0$ to ensure the effective influence of the preference rewards. 
We encode the key objectives from the environment rewards into LLM prompts as human languages to have LLM understand the key objectives of the task, such as following the speed commands for locomotion. Additionally, LAPP prompts LLMs to provide effective feedback on high-level behavior specifications that are difficult or impossible to ground into environment reward functions, such as``having a natural trotting gait''. 
The policy is optimized using PPO \citep{schulman2017proximal}, while new trajectory pairs are continuously collected. Every $M$ epochs, newly collected trajectories are evaluated by the LLM, added to $\mathcal{D}_p$, and used to retrain the preference predictor. LAPP only uses the latest labeled trajectories to retrain the preference predictor. As shown in our ablation studies, this design provides higher performance than including all past trajectories. LAPP's online preference learning allows the policy to progressively align with LLM preferences based on its dynamic evaluation criteria according to different learning stages. The full RL procedure is detailed in Algorithm \ref{alg:lapp}, and the environment rewards $r_E$ of all the tasks can be found in Appendix \ref{apdx:env_rewards}.

\subsection{Network Architectures}

The preference predictor is a transformer network \citep{waswani2017attention} based on the GPT architecture \citep{radford2018improving} with 6 masked self-attention layers.  Inputs are embedded into a 128-dimensional space with sinusoidal positional encodings and processed by 8-headed attention layers. Each block includes a 2-layer MLP with GELU activations \citep{hendrycks2016gaussian}  and layer normalization \citep{ba2016layer} to the output tensor from the last self-attention block. A final decoder outputs a scalar reward.

For Markovian rewards, the input sequence length is $1$ and the casual mask in the self-attention layer is removed. For non-Markovian rewards, the input sequence length is $8$, with zero-padding applied for shorter trajectories.

For quadruped tasks, the policy is an MLP with layers [512, 256, 128] and ELU activations \citep{clevert2015fast}, outputting 12 target joint angles. A PD controller computes the torque commands. For dexterous manipulation, we use the same MLP architecture. The output is a 52-dimensional joint displacement vector for two 26-DoF Shadow Hands \citep{ShadowRobot2005}.


\section{Experiments}
\label{experiments}


We evaluate LAPP on a diverse set of quadruped locomotion and dexterous manipulation tasks to assess its ability to:
\begin{enumerate}
    \item improve both training efficiency and task performance,
    \item enable high-level behavior control via language instructions, and
    \item solve highly challenging tasks that are very difficult or even infeasible with traditional reward engineering. 
\end{enumerate}

\begin{figure*}[hb]
  \centering
  \includegraphics[width=0.96\linewidth]{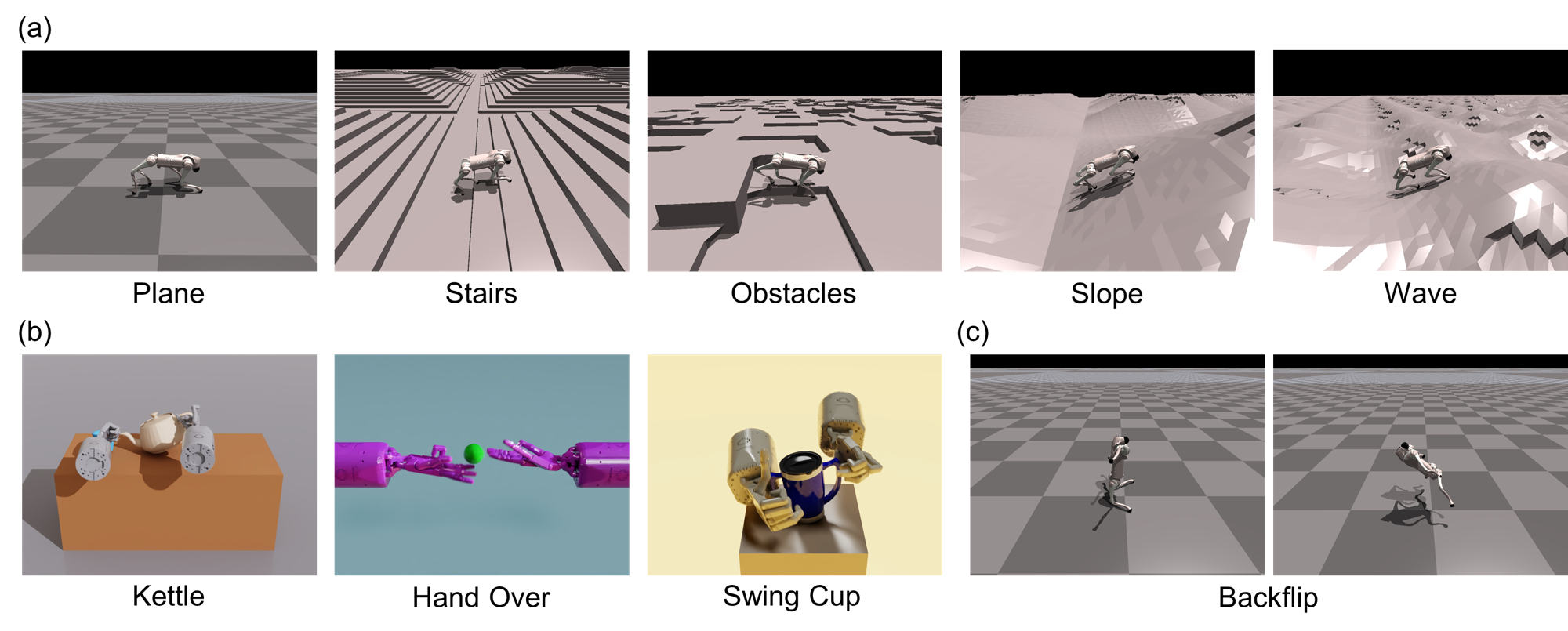}
  \caption{\textbf{Simulation Tasks.} (a) Quadruped locomotion. The robot learns to walk forward across various terrains following given velocity commands. The terrains include the flat plane, stairs pyramids, discrete obstacles, slope pyramids, and wave-pattern hills. (b) Dexterous manipulation. Each dexterous hand has $26$ degrees of freedoms. Kettle requires the robot to pick up the kettle with one hand, and the cup with another hand, and then pour water into the kettle. Hand Over requires one hand to pass a ball to another hand. Swing Cup requires two hands to hold the cup and rotate it for $180^{\circ}$. (c) Quadruped backflip. The robot jumps in the air and rotate backwards for $360^{\circ}$, and then land on the ground.
  }
  \label{fig:sim_tasks}
  \vspace{-5pt}
\end{figure*}

Additionally, we conduct ablation studies to analyze key design choices in LAPP, identifying the factors contributing to its performance gains. Finally, we deploy the trained policies on a physical quadruped robot across various terrains and tasks to demonstrate LAPP's real-world applicability.

We use GPT-4o mini \citep{achiam2023gpt} (\texttt{gpt-4o-mini-2024-07-18} variant) as the LLM backbone for LAPP. A full training run for each policy (5000 epochs) costs approximately \$2.5 to \$3, which is significantly lower than the \$40 to \$50 required for the larger GPT-4o variant, while still achieving satisfactory results. Since LAPP involves frequent online LLM queries, its ability to succeed with a smaller and cheaper LLM is crucial for broader practical adoption. For the Eureka baseline, we use GPT-4o (\texttt{gpt-4o-2024-08-06} variant) to ensure a faithful reproduction of its full capabilities from the original work. Each evolutionary reward search with Eureka costs approximately \$3.

\subsection{Baselines}

\textbf{PPO.} This baseline uses a well-tuned Proximal Policy Optimization (PPO) implementation \citep{rudin2022learning, schulman2017proximal}. In each task, PPO is trained with the same environment reward functions as LAPP. These reward functions are directly adopted from state-of-the-art policies designed by expert robot learning researchers, representing the current best outcomes from human reward engineering. For a fair comparison, PPO shares all hyperparameters with LAPP. The only difference is that PPO does not incorporate preference rewards, allowing us to isolate and analyze the effect of LAPP’s preference-guided learning design.

\textbf{Eureka.} Evolution-driven Universal Reward Kit for Agents (Eureka) \citep{ma2023eureka} is a recent LLM-based approach for automated reward function design. It prompts an LLM with reward design guidelines and environment source code to generate executable Python reward functions. Eureka then performs an evolutionary search to refine the reward function over multiple iterations based on observed training performance. By following the original implementation, we conduct $5$ evolutionary search iterations with $16$ reward samples per iteration. Out of the $80 = 16 \times 5$ reward functions, the best-discovered reward function is then used to train the policy with PPO using the same hyperparameters as the PPO baseline.

\subsection{Simulation Experiments}
\label{sec:sim_exp}

\textbf{Tasks:} We evaluate LAPP on five quadruped locomotion tasks, three dexterous manipulation tasks, and one quadruped backflip task, as shown in Fig.~\ref{fig:sim_tasks}. The Unitree Go2 robot \citep{unitree_rl_gym} is used for quadruped experiments, while the Shadow Dexterous Hand \citep{ShadowRobot2005, andrychowicz2020learning} is used for dexterous manipulation. The locomotion and manipulation tasks are established RL benchmarks from prior works \citep{ma2023eureka, ma2024dreureka, rudin2022learning}. The quadruped backflip task is an extremely challenging control problem, previously studied in multi-objective RL \citep{kim2024stage}. While some RLHF studies have explored backflips, they have primarily used the Hopper model in Gym-Mujoco \citep{christiano2017deep, kim2023preference}, which is significantly easier due to its lower degrees of freedom ($3$ DoFs) with no real-world physical counterpart.

The quadruped locomotion tasks are derived from massively parallel RL experiments in \citet{rudin2022learning}'s prior work. As shown in Fig.~\ref{fig:sim_tasks}(a), we evaluate LAPP on five terrain types including a flat plane, stairs pyramids, discrete obstacles, slope pyramids, and a periodic wave terrain (with periodic wave-pattern hills).

The dexterous manipulation tasks are from the Bidexterous Manipulation (Dexterity) benchmark \citep{chen2022towards} and are also evaluated in Eureka \citep{ma2023eureka}. As shown in Fig.~\ref{fig:sim_tasks} (b), we evaluate LAPP on the Kettle, Hand Over, and Swing Cup tasks. Kettle requires one hand to hold a kettle and pour water into a cup held in the other hand. Hand Over requires to hand over a ball from one hand to another hand. Swing Cup requires the two hands to collaborate to turn a cup for $180^{\circ}$.

Finally, the quadruped backflip task (Fig.~\ref{fig:sim_tasks} (c)) requires the Unitree Go2 robot to perform a $360^{\circ}$ backward rotation mid-air and land successfully. Unlike Hopper-based backflip tasks in prior RLHF studies \citep{christiano2017deep, kim2023preference}, which focus on low DoFs and lightweight dynamics, our setup utilizes the Go2’s official simulator \citep{unitree_rl_gym}, incorporating realistic physical parameters, which makes the task significantly more challenging for RL.

\begin{figure*}[t!]  
  \centering
  \includegraphics[width=0.96\linewidth]{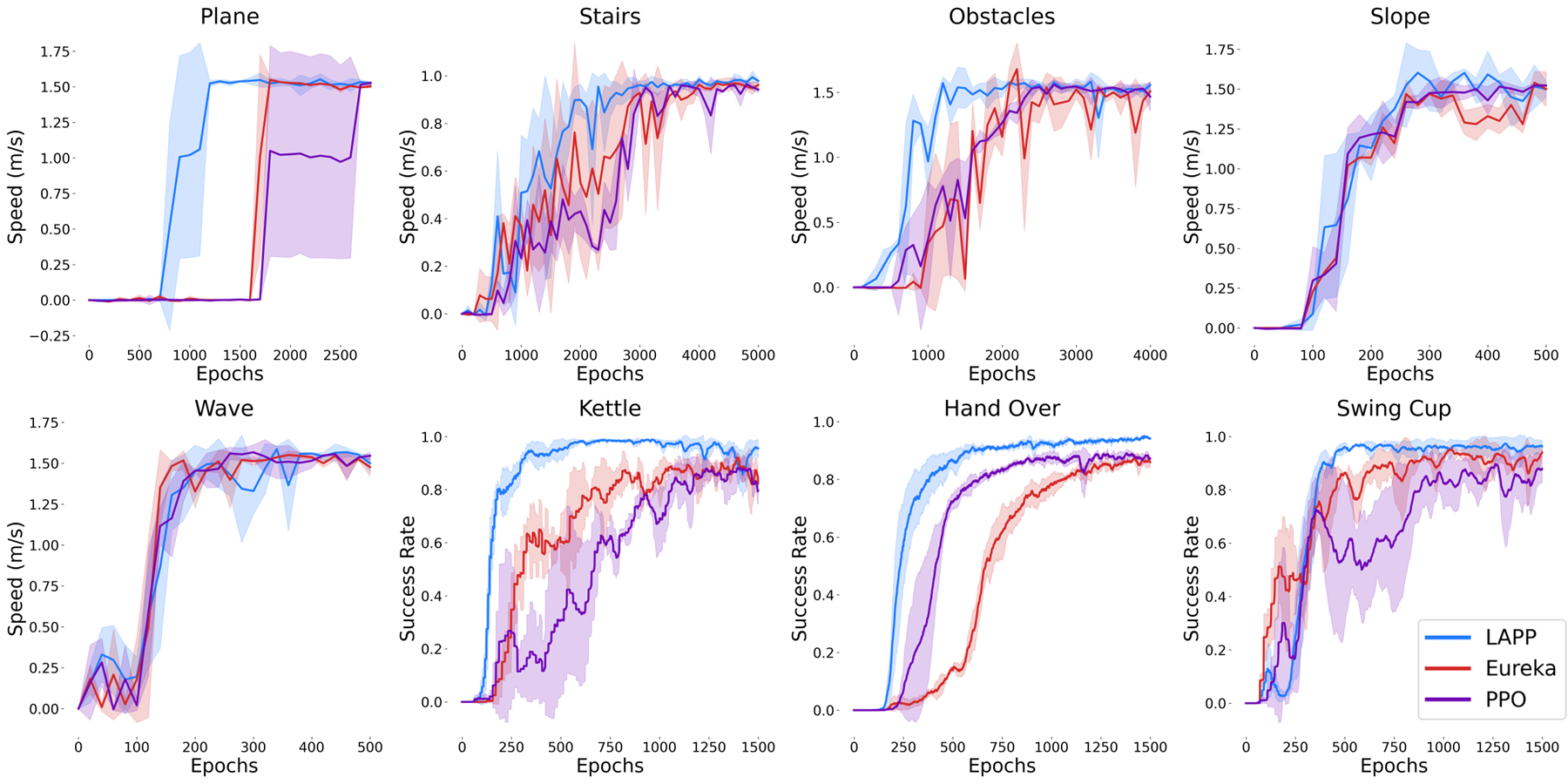}
  \caption{\textbf{Training Efficiency.} Training with LAPP converges faster in the Plane, Stairs, Obstacles, Hand Over, Swing Cup and Kettle tasks, while also exhibiting more stable performance post-convergence in Swing Cup. In the Slope and Wave tasks, LAPP performs similarly to baselines as these tasks are relatively easier for exploration, converging quickly for all algorithms.
  }
  \label{fig:training_speed}
  \vspace{-20pt}
\end{figure*}

\textbf{Lapp improves training efficiency.} Fig.~\ref{fig:training_speed} shows the learning curves of LAPP, Eureka, and PPO across five locomotion tasks and three dexterous manipulation tasks. Locomotion tasks are evaluated with a fixed velocity command of $1.0$ m/s (Stairs) or $1.5$ m/s (other terrains). For manipulation tasks, we report success rate progression throughout training.

\begin{wrapfigure}[]{r}{0.5\textwidth}
\vspace{-32pt}
  \begin{center}
  \includegraphics[width=\linewidth]{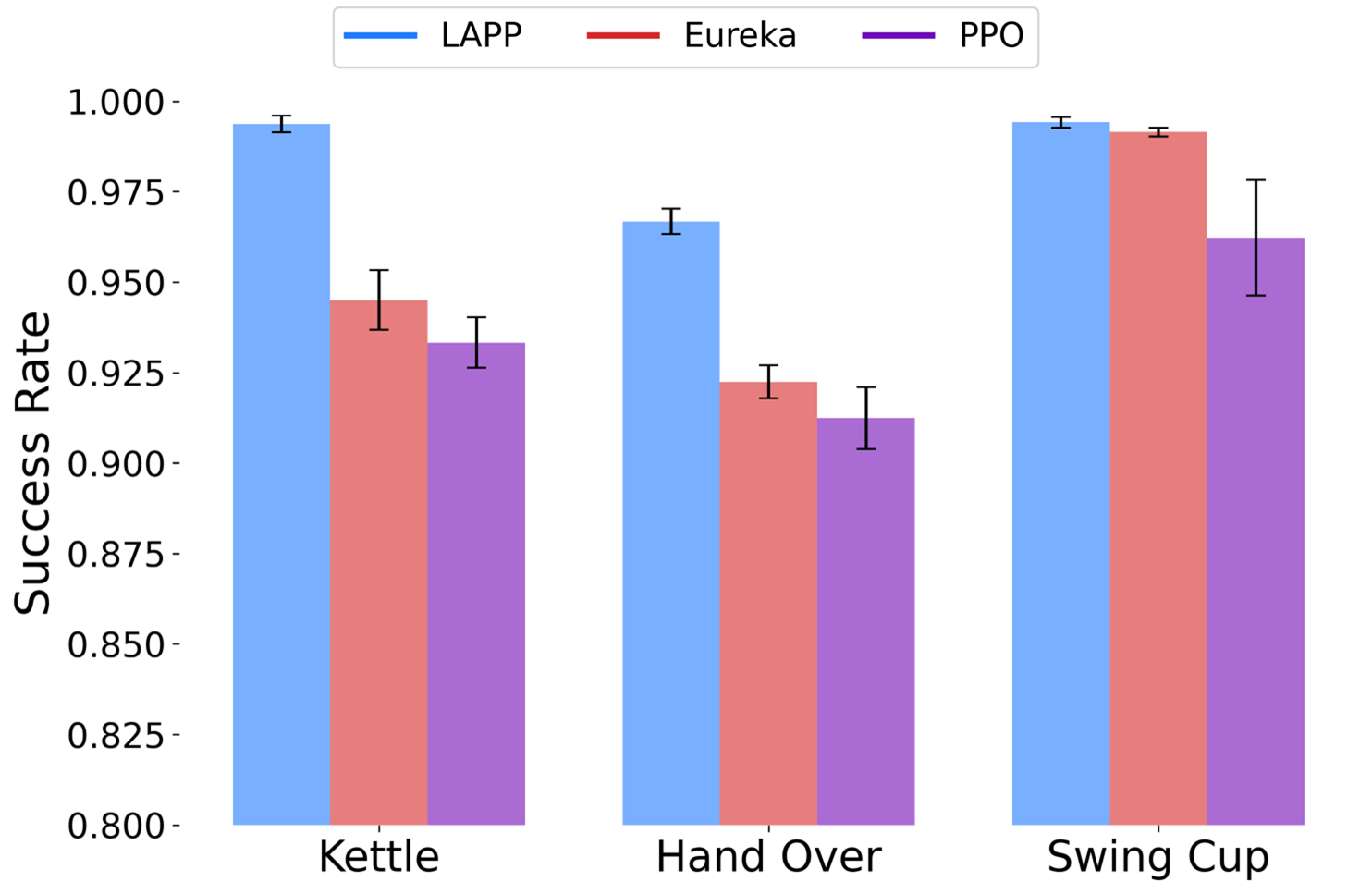}
  \end{center}
  \vspace{-13pt}
  \caption{\textbf{Convergence Success Rate.} LAPP achieves higher success rates in Kettle, Hand Over, and Swing Cup after the training converges. It shows that the preference rewards can continuously refine the robot motions to improve the performance beyond the reach of explicit reward shaping.
  }
  \vspace{-10pt} 
  \label{fig:max_success_rates}
  \vspace{-3pt}
\end{wrapfigure}

LAPP demonstrates faster convergence in flat-plane, stairs, and discrete obstacle locomotion, as well as all manipulation tasks, achieving higher final success rates. These tasks pose non-trivial exploration challenges, where LAPP accelerates learning by dynamically adjusting preference rewards. This flexibility prioritizes different behaviors at different training stages, so that policy exploration can be guided more effectively. In contrast, Eureka struggles with reward balancing, as it relies on a static reward function throughout the training process. While this ensures a well-calibrated reward function, it often results in inferior performance compared to LAPP.

Interestingly, in the Hand Over task, Eureka converges slower than PPO with human-designed rewards. This occurs because Eureka’s evolutionary search optimizes for final performance as the fitness score rather than training efficiency. Therefore, this method can help improve the policy performance, but it may not help with training efficiency.

For relatively easier tasks like Slope and Wave, LAPP exhibits similar performance to baselines. This is because randomized robot initialization on smooth slopes can naturally lead to sliding motions to facilitate early exploration of velocity tracking rewards. As a result, all methods converge within $300$ epochs in these tasks, suggesting no significant advantage for LAPP in environments where task exploration is inherently easier.

\textbf{Lapp achieves higher convergence performance.} Although well-designed reward functions from human experts or LLMs can effectively train quadruped robots to follow velocity commands across various terrains, they often fail to reach optimal performance in more complex dexterous manipulation tasks such as Kettle, Hand Over, and Swing Cup. As shown in Fig.~\ref{fig:max_success_rates}, LAPP significantly improves success rates over PPO desipte sharing the same environment reward functions, increasing from $92\%$ to $99\%$ in Kettle, $91\%$ to $97\%$ in Hand Over, and $96\%$ to $99\%$ in Swing Cup. These gains stem from the continuous motion refinement enabled by LAPP’s dynamically updated preference predictor.

Compared to Eureka, LAPP achieves a $6\%$ higher success rate in Kettle and a $5\%$ improvement in Hand Over. In the Swing Cup task, both LAPP and Eureka reach near-optimal 99\% success rates, but LAPP converges faster and maintains more stable performance over extended training epochs as in Fig.~\ref{fig:training_speed}.

\begin{figure*}[htbp]
  \centering
  \includegraphics[width=0.999\linewidth]{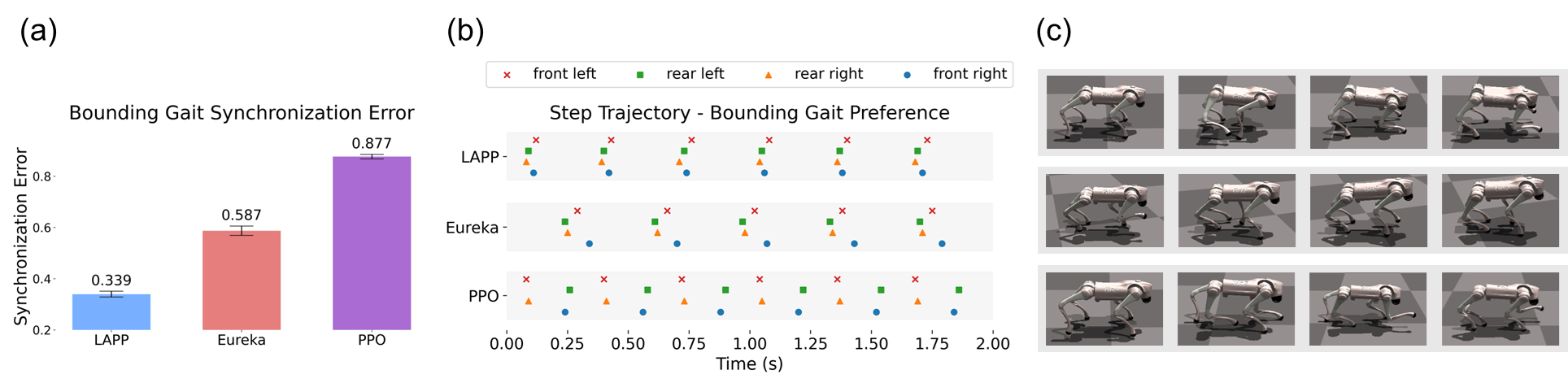}
  \caption{\textbf{Bounding Gait Pattern Control.} (a) Feet synchronization error calculated with Eq.~\ref{eq:sync_error}. LAPP achieves the lowest synchronization error, indicating its closest adherence to a bounding gait. (b) Step trajectories of the robots. LAPP synchronizes both front and rear feet, while Eureka aligns only the rear feet, and PPO fails to produce a bounding gait. (c) Motion frames of the robots.}
  \label{fig:gait_pattern_control}
\vspace{-5pt}
\end{figure*}

\textbf{LAPP enables behavior control via instruction.} Traditional RL can train robots to complete tasks but cannot typically control how they perform them in a way that aligns with high-level human preferences. Can LAPP guide robot behaviors using high-level specifications in the behavior instruction prompt? To investigate this question, we design two experiments: 1) enforcing a bounding gait in quadruped forward locomotion, and 2) controlling gait cadence to be either higher or lower in quadruped forward locomotion.

A bounding gait requires the quadruped’s front and rear feet to make simultaneous ground contact in pairs. To quantify how closely a robot’s gait adheres to this pattern, we adopt a synchronization error definition as in Eq.~\ref{eq:sync_error}:
\begin{equation}
\label{eq:sync_error}
\text{ sync\_error} =\frac{1}{N} \sum_{t=1}^N\left(\left|\mathrm{FL}_t-\mathrm{FR}_t\right|+\left|\mathrm{RL}_t-\mathrm{RR}_t\right|\right) ,
\end{equation}
where $\mathrm{FL}_t$, $\mathrm{FR}_t$, $\mathrm{RL}_t$, and $\mathrm{RR}_t$ represent front left feet, front right feet, rear left feet, rear right feet contacts at time $t$. These are binary values: $1$ if the foot is in contact with the ground and $0$ if it is in the air. A lower synchronization error indicates a gait pattern closer to bounding.

As shown in Fig.~\ref{fig:gait_pattern_control} (a), LAPP trains the robot to achieve a bounding gait with the lowest synchronization error. While Eureka also encourages a bounding gait through reward shaping, its synchronization error remains higher than LAPP. PPO fails to enforce a bounding gait effectively.

\begin{figure*}[htbp]
  \centering
  \includegraphics[width=0.999\linewidth]{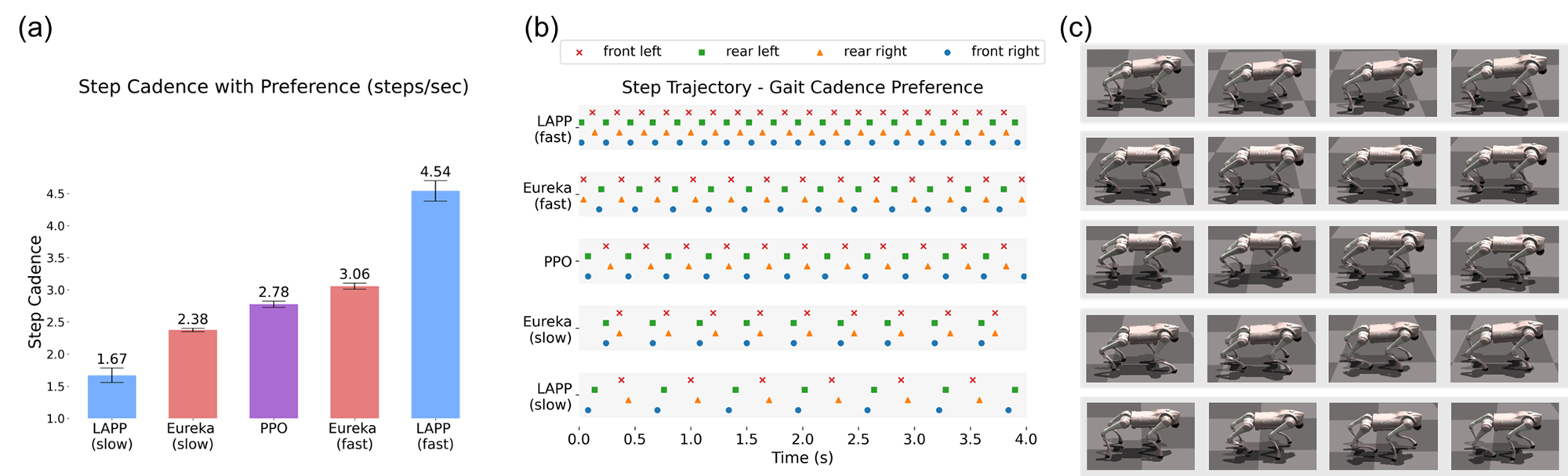}
  \caption{\textbf{Gait Cadence Control.} (a) Step trajectories under different cadence instructions. LAPP effectively modulates step frequency based on the high-level prompts by following faster and slower gaits. Eureka can adjust the cadence slightly, but it is less effective.(b) Step cadence comparison. LAPP provides precise control over cadence, while Eureka has limited effect. (c) Motion frames illustrating cadence variation. Robots with high cadence take quick and shallow steps, while those with low cadence take larger and higher steps.}
  \label{fig:gait_cadence_control}
\vspace{-10pt}
\end{figure*}

To further illustrate gait patterns, Fig.~\ref{fig:gait_pattern_control} (b) presents the step trajectories of all methods, where each dot represents a foot contacting the ground. LAPP successfully trains the robot to synchronize its front and rear feet, ensuring that both front feet land simultaneously, followed by both rear feet. In contrast, Eureka achieves partial synchronization, aligning only the rear feet. PPO fails to learn a bounding gait with unsynchronized foot contacts.

Fig.~\ref{fig:gait_pattern_control} (c) provides motion frames of the robots trained with LAPP and the baselines. These visualizations complement the trajectory plots, clearly demonstrating that LAPP exhibits stronger behavior control through effective implicit reward shaping to follow the high-level gait patterns.

\begin{figure*}[t!]
  \centering
  \includegraphics[width=0.98\linewidth]{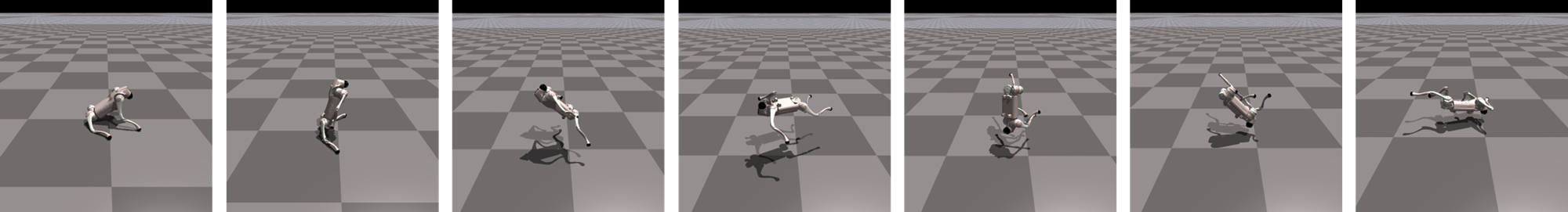}
  \caption{\textbf{Quadruped Robot Backflip.} LAPP successfully trains the Unitree Go2 robot to acomplish the backflip task. No baseline including PPO, curriculum learning, or Eureka is able to solve this task.
  }
  \label{fig:sim_backflip}
\vspace{-10pt}
\end{figure*}

We also evaluate LAPP’s ability to control step cadence through high-level instructions. In this experiment, the behavior instruction prompt specifies a preference for either faster or slower stepping frequency. We compare the effectiveness of LAPP and Eureka in enforcing these behaviors. Eureka is also prompted to generate reward functions that encourage the desired cadence. During testing, all robots follow a velocity command of $1.5$ m/s. As shown in Fig.~\ref{fig:gait_cadence_control} (a), with a high-cadence instruction, LAPP achieves 4.54 steps/sec, significantly exceeding Eureka (3.06 steps/sec) and PPO (2.78 steps/sec). Similarly, with a low-cadence instruction, LAPP produces 1.67 steps/sec, notably lower than Eureka (2.38 steps/sec) and PPO (2.78 steps/sec). While Eureka can influence cadence through reward shaping, its effect is far weaker than LAPP’s.

The step trajectories in Fig.~\ref{fig:gait_cadence_control} (b) further highlight LAPP’s superior cadence control. Compared to all other methods, LAPP produces the densest step trajectory under a high-cadence instruction and the sparsest trajectory under a low-cadence instruction. Qualitatively, Fig.~\ref{fig:gait_cadence_control} (c) presents motion frames illustrating the impact of cadence control. With fast cadence, LAPP trains the robot to take small and rapid steps with minimal foot lift. Conversely, with slow cadence, the robot takes larger strides, keeping its feet in the air for extended periods.

Notably, this experiment uses the non-Markovian reward model from Eq.~\ref{eq:non_markovian_preference}. We set the transformer preference predictor’s input sequence length to 8. To control the step cadence, the preference predictor needs to consider the history states of the feet contacts to determine the latent reward value of the current state.

\textbf{LAPP solves challenging tasks.} Quadruped backflips have long been considered a challenging RL problem due to the need for precise whole-body coordination, complex dynamics, and controlled landing. Some previous RLHF works have trained a Hopper model to perform a backflip in the Gym-Mujoco simulator \citep{arumugam2019deep}, but the Hopper has only three joints, is lightweight, and lacks a real-world counterpart.

\begin{wrapfigure}[]{r}{0.4\textwidth}
\vspace{-10pt}
  \begin{center}
  \includegraphics[width=\linewidth]{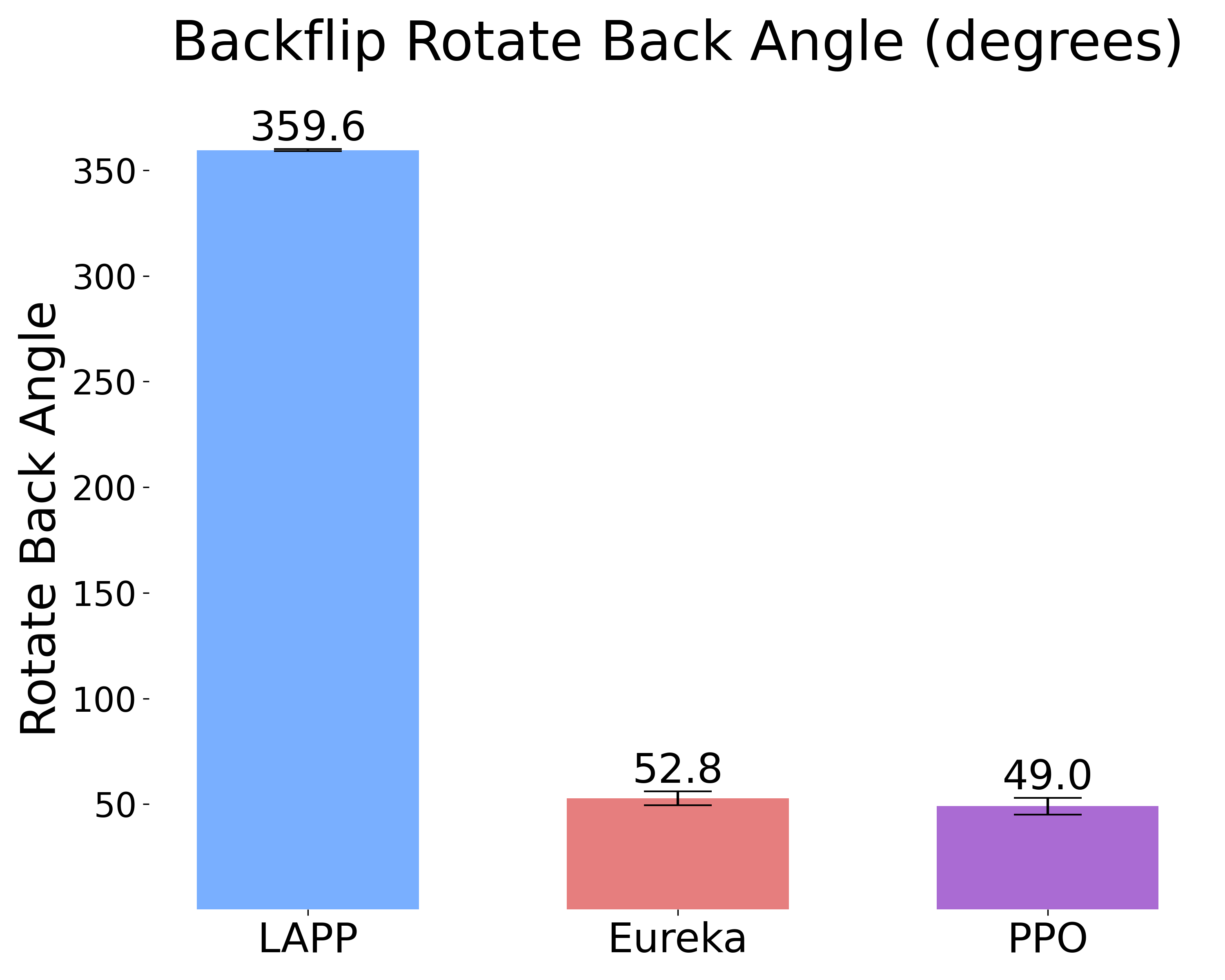}
  \end{center}
  \vspace{-13pt}
  \caption{\textbf{Backflip rotate angle.} LAPP successfully enables the quadruped robot to complete backflips with full $360^{\circ}$ rotation. In contrast, PPO and Eureka fail to generate sufficient rotation, highlighting the advantage of preference-driven learning in solving highly complex and dynamic tasks.
  }
  \label{fig:backflip_angle}
\end{wrapfigure}

A recent multi-objective RL (MORL) approach solves the quadruped backflip by dividing the motion into five stages, designing a separate handcrafted reward function for each stage \citep{kim2024stage}. However, this method requires significant expert knowledge, as practitioners must manually define stage transitions and fine-tune rewards for each specific robot.

In contrast, LAPP solves the backflip without intensive human labor for preference feedback or manual reward tuning. We train a Unitree Go2 robot using a weighted combination of a human-designed environment reward and a predicted preference reward. Our process still requires an initial warm-up to encourage the exploration, but we limit our process with simple reward designs for each step. Specifically, we first pre-train the robot to jump vertically. We then randomly initialize the robot in the air. We also reduce the robot's weight during training but restore its real-world weight for testing.

\begin{wrapfigure}[]{r}{0.5\textwidth}
\vspace{-20pt}
  \begin{center}
  \includegraphics[width=\linewidth]{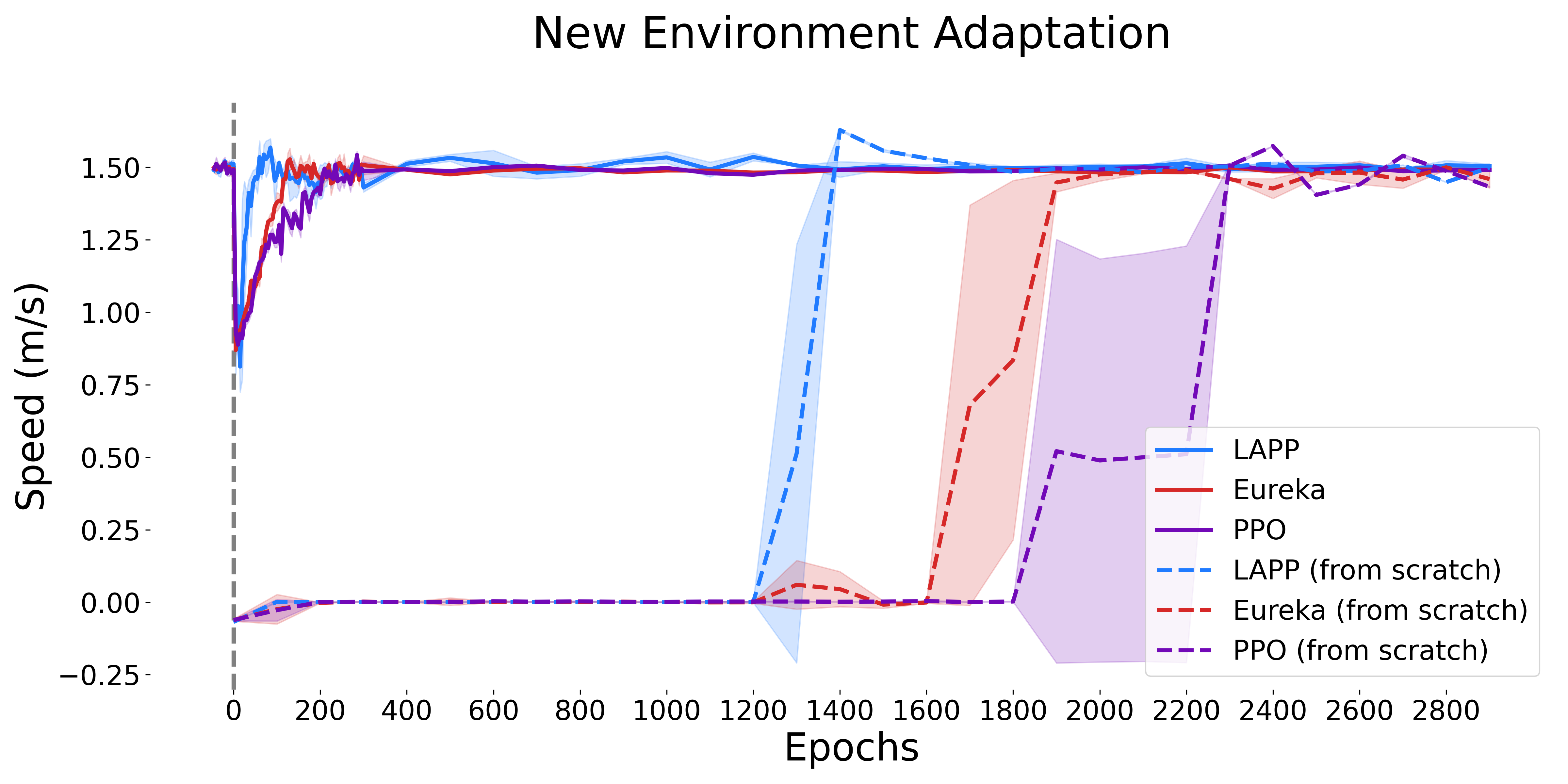}
  \end{center}
  \vspace{-13pt}
  \caption{\textbf{Transfer Learning.} For the Plane locomotion task, the robot is transferred to a new environment with different friction and restitution. The solid curves show that LAPP enables the robot to adapt to the new environment faster. Compared with the dashed curves, transfer learning is generally faster than training from scratch.
  }
  \label{fig:new_env_adapt}
\end{wrapfigure}

As shown in Fig.~\ref{fig:sim_backflip} and Fig.~\ref{fig:backflip_angle}, LAPP successfully trains the robot to jump, rotate backward $360^{\circ}$, and land safely. We also evaluate PPO and Eureka on the same task, using identical exploration strategies (i.e., pre-training to jump, random air initialization, and reduced training weight). PPO follows the same human-designed backflip reward as LAPP but lacks a preference reward, while Eureka generates its own reward function via GPT-4o. Neither PPO nor Eureka succeeds in training the robot for a full backflip. Instead, PPO learns to jump and oscillate the robot's torso up and down but struggles to flip for over $180^{\circ}$ with an average maximum rotation of $49.0^{\circ}$. Despite the iterative reward search mechanism of Eureka, it produces similar behavior to PPO with an average maximum rotation of $52.8^{\circ}$. These results demonstrate that explicit reward engineering struggles to capture the complex dynamics of a backflip, while LAPP’s preference-driven learning and the adaptive online predictor updates enable successful learning of this highly dynamic capability. We believe that LAPP can shed light on automatically solving many difficult tasks in the near future that are previously unsolvable by conventional RL.

\textbf{LAPP enables faster transfer learning.} We evaluate the transfer learning performance of LAPP. For the Plane task with a forward speed command of $1.5$ m/s, the Go2 robot is first trained on a flat ground with static friction of $1.0$, dynamic friction of $1.0$, and a restitution of $0.0$. Then, it is transferred to a different flat ground with static friction of $0.01$, dynamic friction of $0.01$, and a restitution of $0.9$. We fine-tune the policy from the source environment in the target environment with LAPP and other baselines. As shown in Fig. ~\ref{fig:new_env_adapt}, the robot speed drops to about $0.9$ m/s due to the more slippery ground surface, and LAPP trains the robot to adapt to this new environment faster than the other two baselines. The dashed curves show the learning processes of the robots trained in the target environment from scratch. The results show that transfer learning with a pre-trained policy in a different environment is much faster than training from scratch, and LAPP trains the robot to adapt faster than other baselines.

\subsection{Ablation Study}

\begin{figure*}[t!]
  \centering
  \includegraphics[width=0.96\linewidth]{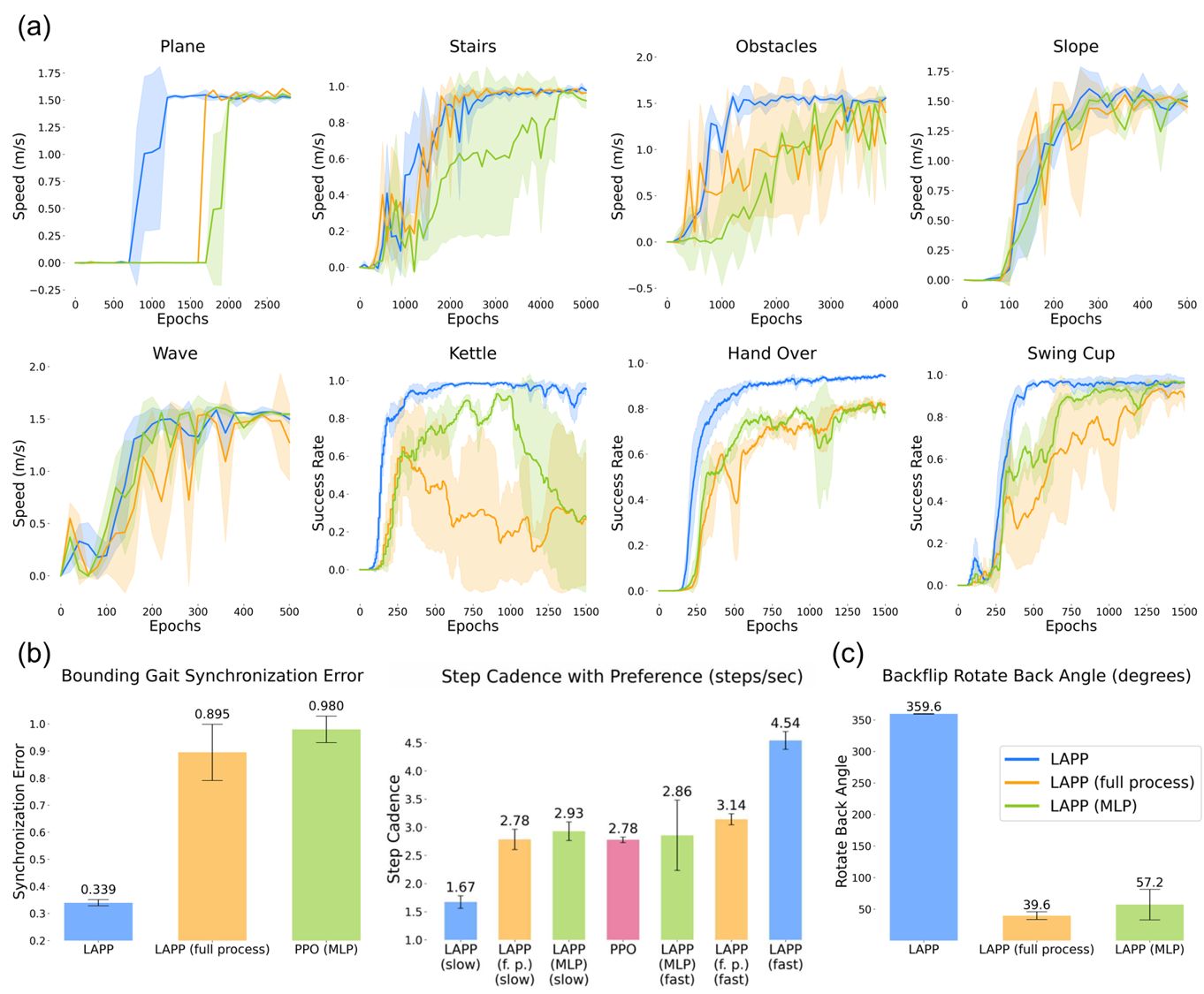}
  \caption{\textbf{Ablation Study.} We evaluate two key design choices of LAPP: (1) updating the preference predictor with the latest rollout trajectories (\textbf{LAPP} vs. LAPP (full process)), and (2) using a transformer-based reward predictor (\textbf{LAPP} vs. LAPP (MLP)). (a) Training Speed: LAPP generally converges faster, except in the Stairs task, where LAPP (full process) achieves similar speed, and in the Slope and Wave tasks, where both tasks are simple enough for LAPP and its ablations to have similar performance. In the Kettle task, both ablations fail at different stages, suggesting that a suboptimal training method for the preference predictor can disrupt policy learning. (b) Behavior Control: Only LAPP successfully controls gait pattern and cadence, while both ablations fail. (c) Challenging Task: LAPP enables successful backflips, achieving a full $360^{\circ}$ rotation. In contrast, LAPP (MLP) and LAPP (full process) reach only $57.2^{\circ}$ and $39.6^{\circ}$, respectively, and fail to complete the task.
  }
  \label{fig:ablation}
\vspace{-15pt}
\end{figure*}

We carry out the ablation study to analyze the contributions of two key design decisions of LAPP: 1) updating the preference predictor with the latest rollout trajectories instead of trajectories sampled from the full RL process, and 2) adopting a transformer architecture for the reward predictor network instead of a simple MLP.

To investigate the first design choice, we introduce LAPP (full process), which stores all rollout trajectories in a pool and samples from the entire RL process rather than only the latest epochs (set to $500$ epochs in LAPP). The dataset size remains identical between LAPP and LAPP (full process), ensuring a controlled comparison. However, unlike LAPP, this variant may use trajectory data from early training, which potentially introduces biases to suboptimal policy rollouts. Note that we do not use the full trajectory pool for predictor training, as its continuous expansion would slow down online updates.

For the second ablation, LAPP (MLP) replaces the transformer-based predictor with an MLP, limiting it to Markovian preference rewards (Eq.~\ref{eq:markovian_preference}). In contrast, LAPP supports both Markovian and non-Markovian rewards (Eq.~\ref{eq:non_markovian_preference}), which are essential for modeling tasks like step cadence control and backflips.

We compare the performance of LAPP and its two ablations in the simulation tasks in Sec.~\ref{sec:sim_exp}. Fig.~\ref{fig:ablation} (a) shows that replacing the transformer with an MLP reduces training speed across most locomotion tasks except for the Slope and Wave tasks, which are relatively simple and lead to similar performance of LAPP and its ablations. Moreover, sampling preference data from the full RL process does not affect speed in Stairs, Slope, and Wave, but slows down the training in other locomotion tasks. In Kettle, both ablations initially improve performance but then drop to $20\%$, suggesting that biased preference data or a predictor network that does not capture long-term dependencies in past states can introduce misleading preference rewards, which ultimately destabilize the policy training.

Fig.~\ref{fig:ablation} (b) evaluates the ability to control gait pattern and step cadence in quadruped locomotion. Only LAPP succeeds, while both ablations fail. This supports the hypothesis that preference rewards must evolve dynamically with training. LAPP (full process) struggles because sampling from the full trajectory pool prevents the predictor from adapting to the current learning stage. LAPP (MLP) fails because MLPs lack the capacity to model non-Markovian rewards, which are essential for cadence control. Specifically, bounding gait requires analyzing foot synchronization from step history, and cadence control depends on tracking past states to predict step timing rewards. However, MLPs cannot effectively capture these temporal dependencies, leading to failure in high-level behavior control.

Fig.~\ref{fig:ablation} (c) evaluates the quadruped backflip task. Neither ablation succeeds. LAPP (full process) reaches $39.6^{\circ}$ degrees of backward rotation, while LAPP (MLP) improves slightly to $57.2^{\circ}$. LAPP completes the full $360^{\circ}$ backflip, demonstrating the importance of both preference data sampling and transformer-based prediction for capturing complex dynamics.

\subsection{Real World Experiments}

To evaluate the feasibility of deploying LAPP-trained policies on real robots, we conduct experiments with the Unitree Go2 quadruped. The policy is first trained in the IsaacGym simulator \citep{makoviychuk2021isaac} and then directly deployed in the real world. To mitigate the sim-to-real gap, we apply domain randomization, varying ground friction, robot mass, observation noise, and external disturbances during training.

\begin{wrapfigure}[]{r}{0.5\textwidth}
\vspace{-12pt}
  \begin{center}
  \includegraphics[width=\linewidth]{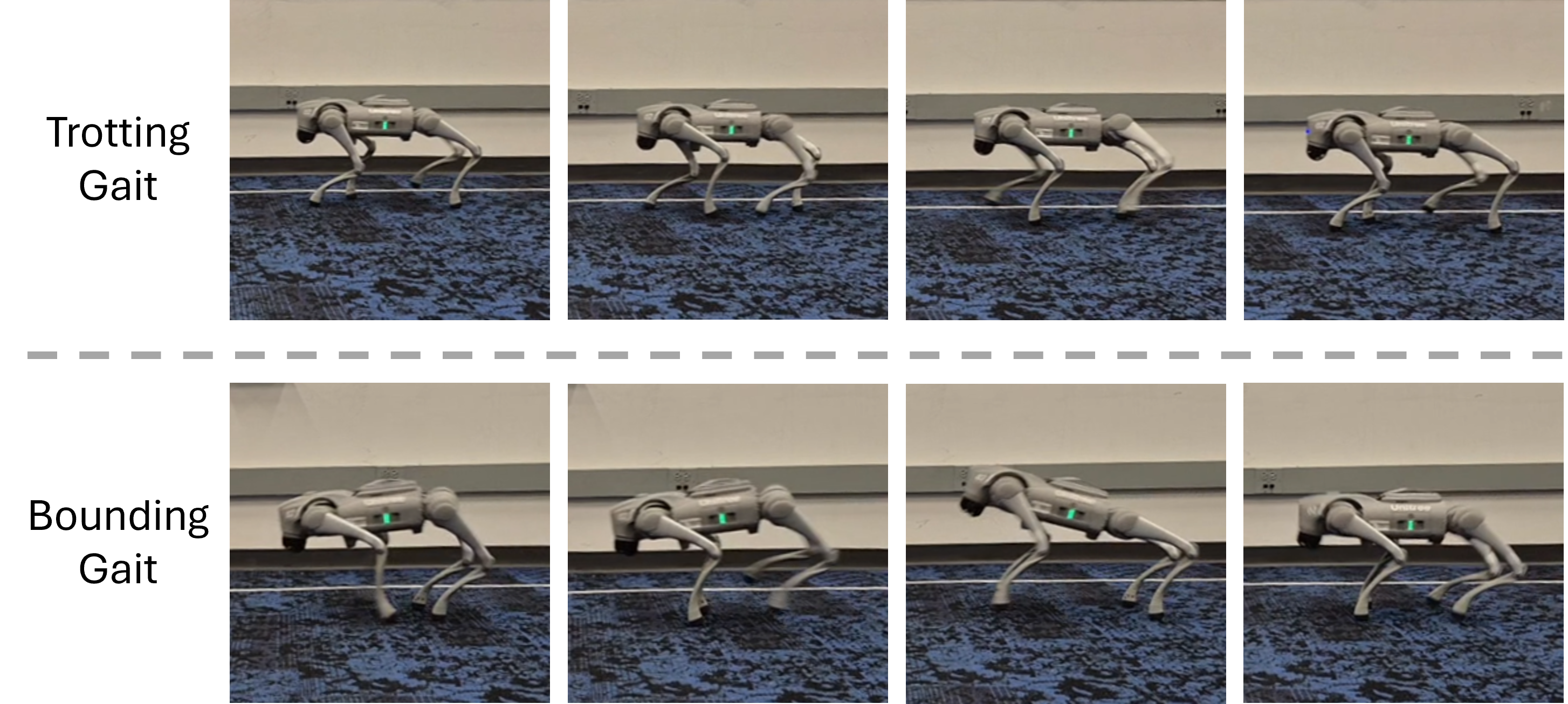}
  \end{center}
  \vspace{-13pt}
  \caption{\textbf{Gait Pattern Control.} We can directly deploy the policy trained by LAPP on trotting and bounding gait control, specified via high-level language instructions in the behavior prompts.
  }
  \vspace{-10pt} 
  \label{fig:real_gait_pattern}
  \vspace{-3pt}
\end{wrapfigure}

We first test gait pattern control. By specifying a preferred gait pattern in the behavior instruction prompt, LAPP successfully trains the robot to walk forward using either a trotting or bounding gait as shown in Fig.~\ref{fig:real_gait_pattern}. 

\begin{wrapfigure}[]{r}{0.5\textwidth}
\vspace{-12pt}
  \begin{center}
  \includegraphics[width=\linewidth]{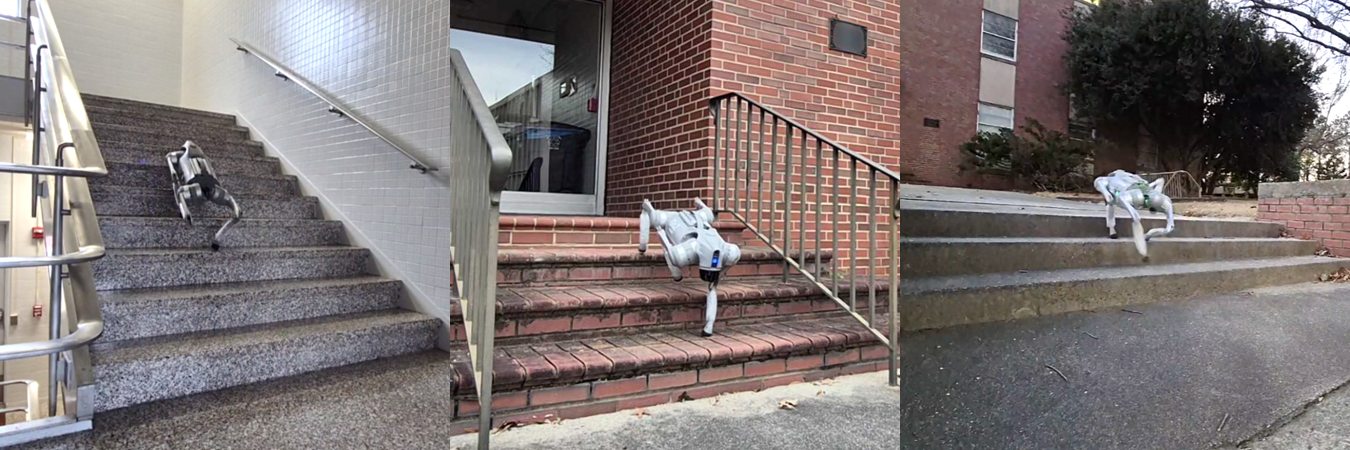}
  \end{center}
  \vspace{-5pt}
  \caption{\textbf{Stairs Climbing.} LAPP-trained policies enable the Unitree Go2 robot to climb up and down stairs in a real-world deployment. The stairs are approximately $17$ cm in height.
  }
  \vspace{-10pt} 
  \label{fig:real_stairs}
  \vspace{-3pt}
\end{wrapfigure}

We then evaluate stair climbing. In Sec.~\ref{sec:sim_exp}, Fig.~\ref{fig:training_speed} has shown that LAPP accelerates stair-climbing training in simulation. Here, we deploy the trained policy at epoch 2200 onto a real 17 cm-high staircase. As shown in Fig.~\ref{fig:real_stairs}, LAPP-trained robot successfully climbs both up and down stairs. In contrast, policies trained with PPO or Eureka under the same number of epochs often fail to maintain stability, stumble, and fall while navigating stairs. These results demonstrate LAPP’s robustness in real-world deployment, effectively translating simulation-trained behaviors to real hardware while reducing manual reward engineering efforts.

\section{Conclusion}
\label{sec:conclusion}

We introduced LAPP, a novel framework that leverages LLM for preference feedback from raw state-action trajectories to guide reinforcement learning. Given only high-level behavior specifications in natural language, LAPP automatically learns a preference reward predictor using LLM-generated feedback and continuously refines robot motions to align with the specified behavior through the adaptive reward predictor.

Compared to conventional RL approaches that rely on explicitly shaped reward functions, our experiments in both simulation and real-world deployment demonstrate that LAPP achieves faster training convergence while maintaining superior performance. Moreover, LAPP enables customizable high-level behavior controls. Finally, LAPP generalizes to complex and non-Markovian preference rewards, surpassing traditional reward engineering methods and LLM-generated reward functions. Notably, LAPP successfully solves the quadruped backflip task for the first time under a basic RL setting, aided only by simple exploration warm-up steps. In contrast, all other baselines fail under the same conditions.

These results showcase LAPP’s potential to push the boundaries of RL for robot learning, expanding robot capabilities through foundation model-driven behavior guidance.
\section{Limitations}
\label{sec:Limitations}

We note several areas for future improvements of LAPP:

\textbf{Frequent LLM queries.} Despite our batch query process, LAPP still queries LLM very frequently due to its online adaptive training scheme for the preference predictor. Future work could explore more efficient data utilization to maintain or improve performance without biasing the predictor training while reducing LLM query frequency.
    
\textbf{Manual selection of state variables.} Our trajectory state representation is curated for LLM queries to include appropriate amount of the information. Including too many variables results in long prompts leads to higher costs, and potential confusion in LLM's responses, while too few variables may not provide sufficient context for accurate reasoning. Future research could develop automated methods for state variable selection, optimizing the balance between prompt length, cost, and label accuracy. One possible method is to warm up the training with different variable selections and choose the best set based on the warm-up performance.

\textbf{Absence of visual inputs.} 
LAPP currently does not consider the tasks that requires visual inputs to the policy networks such as reactive motion planning in locomotion tasks, and its feasibility to handle state trajectories with images remains unexplored. With the reasoning capability of VLMs continuously improving, future work can explore the similar idea for tasks with visual trajectories.
This could enhance preference prediction accuracy and enable richer robot capabilities.

\section*{Acknowledgments}
This work is supported by ARL STRONG program under awards W911NF2320182, W911NF2220113 and W911NF2420215, by DARPA FoundSci program under award HR00112490372, by DARPA TIAMAT HR00112490419, and by AFOSR under award \#FA9550-19-1-0169.

Copyright 2025. Duke University. This paper is hereby licensed publicly under the CC BY-NC-ND 4.0 License. Duke University has filed patent rights for the technology associated with this article. For further license rights, including using the patent rights for commercial purposes, please contact Duke’s Office for Translation and Commercialization (otcquestions@duke.edu) and reference OTC File 8724.

\bibliography{main}

\begin{thebibliography}{132}
\providecommand{\natexlab}[1]{#1}
\providecommand{\url}[1]{\texttt{#1}}
\expandafter\ifx\csname urlstyle\endcsname\relax
  \providecommand{\doi}[1]{doi: #1}\else
  \providecommand{\doi}{doi: \begingroup \urlstyle{rm}\Url}\fi

\bibitem[Abbeel \& Ng(2004)Abbeel and Ng]{abbeel2004apprenticeship}
Pieter Abbeel and Andrew~Y Ng.
\newblock Apprenticeship learning via inverse reinforcement learning.
\newblock In \emph{Proceedings of the twenty-first international conference on Machine learning}, pp.\ ~1, 2004.

\bibitem[Abdolmaleki et~al.(2020)Abdolmaleki, Huang, Hasenclever, Neunert, Song, Zambelli, Martins, Heess, Hadsell, and Riedmiller]{abdolmaleki2020distributional}
Abbas Abdolmaleki, Sandy Huang, Leonard Hasenclever, Michael Neunert, Francis Song, Martina Zambelli, Murilo Martins, Nicolas Heess, Raia Hadsell, and Martin Riedmiller.
\newblock A distributional view on multi-objective policy optimization.
\newblock In \emph{International conference on machine learning}, pp.\  11--22. PMLR, 2020.

\bibitem[Achiam et~al.(2023)Achiam, Adler, Agarwal, Ahmad, Akkaya, Aleman, Almeida, Altenschmidt, Altman, Anadkat, et~al.]{achiam2023gpt}
Josh Achiam, Steven Adler, Sandhini Agarwal, Lama Ahmad, Ilge Akkaya, Florencia~Leoni Aleman, Diogo Almeida, Janko Altenschmidt, Sam Altman, Shyamal Anadkat, et~al.
\newblock Gpt-4 technical report.
\newblock \emph{arXiv preprint arXiv:2303.08774}, 2023.

\bibitem[Ahn et~al.(2022)Ahn, Brohan, Brown, Chebotar, Cortes, David, Finn, Fu, Gopalakrishnan, Hausman, et~al.]{ahn2022can}
Michael Ahn, Anthony Brohan, Noah Brown, Yevgen Chebotar, Omar Cortes, Byron David, Chelsea Finn, Chuyuan Fu, Keerthana Gopalakrishnan, Karol Hausman, et~al.
\newblock Do as i can, not as i say: Grounding language in robotic affordances.
\newblock \emph{arXiv preprint arXiv:2204.01691}, 2022.

\bibitem[Akrour et~al.(2011)Akrour, Schoenauer, and Sebag]{akrour2011preference}
Riad Akrour, Marc Schoenauer, and Michele Sebag.
\newblock Preference-based policy learning.
\newblock In \emph{Machine Learning and Knowledge Discovery in Databases: European Conference, ECML PKDD 2011, Athens, Greece, September 5-9, 2011. Proceedings, Part I 11}, pp.\  12--27. Springer, 2011.

\bibitem[Akrour et~al.(2012)Akrour, Schoenauer, and Sebag]{akrour2012april}
Riad Akrour, Marc Schoenauer, and Mich{\`e}le Sebag.
\newblock April: Active preference learning-based reinforcement learning.
\newblock In \emph{Machine Learning and Knowledge Discovery in Databases: European Conference, ECML PKDD 2012, Bristol, UK, September 24-28, 2012. Proceedings, Part II 23}, pp.\  116--131. Springer, 2012.

\bibitem[Amershi et~al.(2014)Amershi, Cakmak, Knox, and Kulesza]{amershi2014power}
Saleema Amershi, Maya Cakmak, William~Bradley Knox, and Todd Kulesza.
\newblock Power to the people: The role of humans in interactive machine learning.
\newblock \emph{AI magazine}, 35\penalty0 (4):\penalty0 105--120, 2014.

\bibitem[Andrychowicz et~al.(2020)Andrychowicz, Baker, Chociej, Jozefowicz, McGrew, Pachocki, Petron, Plappert, Powell, Ray, et~al.]{andrychowicz2020learning}
OpenAI:~Marcin Andrychowicz, Bowen Baker, Maciek Chociej, Rafal Jozefowicz, Bob McGrew, Jakub Pachocki, Arthur Petron, Matthias Plappert, Glenn Powell, Alex Ray, et~al.
\newblock Learning dexterous in-hand manipulation.
\newblock \emph{The International Journal of Robotics Research}, 39\penalty0 (1):\penalty0 3--20, 2020.

\bibitem[Aroca-Ouellette et~al.(2024)Aroca-Ouellette, Mackraz, Theobald, and Metcalf]{aroca2024predict}
Stephane Aroca-Ouellette, Natalie Mackraz, Barry-John Theobald, and Katherine Metcalf.
\newblock Predict: Preference reasoning by evaluating decomposed preferences inferred from candidate trajectories.
\newblock \emph{arXiv preprint arXiv:2410.06273}, 2024.

\bibitem[Arora \& Doshi(2021)Arora and Doshi]{arora2021survey}
Saurabh Arora and Prashant Doshi.
\newblock A survey of inverse reinforcement learning: Challenges, methods and progress.
\newblock \emph{Artificial Intelligence}, 297:\penalty0 103500, 2021.

\bibitem[Arumugam et~al.(2019)Arumugam, Lee, Saskin, and Littman]{arumugam2019deep}
Dilip Arumugam, Jun~Ki Lee, Sophie Saskin, and Michael~L Littman.
\newblock Deep reinforcement learning from policy-dependent human feedback.
\newblock \emph{arXiv preprint arXiv:1902.04257}, 2019.

\bibitem[Authors(2024)]{Genesis}
Genesis Authors.
\newblock Genesis: A universal and generative physics engine for robotics and beyond, December 2024.
\newblock URL \url{https://github.com/Genesis-Embodied-AI/Genesis}.

\bibitem[Ba(2016)]{ba2016layer}
Jimmy~Lei Ba.
\newblock Layer normalization.
\newblock \emph{arXiv preprint arXiv:1607.06450}, 2016.

\bibitem[Bai et~al.(2022)Bai, Kadavath, Kundu, Askell, Kernion, Jones, Chen, Goldie, Mirhoseini, McKinnon, et~al.]{bai2022constitutional}
Yuntao Bai, Saurav Kadavath, Sandipan Kundu, Amanda Askell, Jackson Kernion, Andy Jones, Anna Chen, Anna Goldie, Azalia Mirhoseini, Cameron McKinnon, et~al.
\newblock Constitutional ai: Harmlessness from ai feedback.
\newblock \emph{arXiv preprint arXiv:2212.08073}, 2022.

\bibitem[Basaklar et~al.(2022)Basaklar, Gumussoy, and Ogras]{basaklar2022pd}
Toygun Basaklar, Suat Gumussoy, and Umit~Y Ogras.
\newblock Pd-morl: Preference-driven multi-objective reinforcement learning algorithm.
\newblock \emph{arXiv preprint arXiv:2208.07914}, 2022.

\bibitem[Bradley \& Terry(1952)Bradley and Terry]{bradley1952rank}
Ralph~Allan Bradley and Milton~E Terry.
\newblock Rank analysis of incomplete block designs: I. the method of paired comparisons.
\newblock \emph{Biometrika}, 39\penalty0 (3/4):\penalty0 324--345, 1952.

\bibitem[Brown et~al.(2018)Brown, Cui, and Niekum]{brown2018risk}
Daniel~S Brown, Yuchen Cui, and Scott Niekum.
\newblock Risk-aware active inverse reinforcement learning.
\newblock In \emph{Conference on Robot Learning}, pp.\  362--372. PMLR, 2018.

\bibitem[Brown et~al.(2020)Brown, Mann, Ryder, Subbiah, Kaplan, Dhariwal, Neelakantan, Shyam, Sastry, Askell, et~al.]{brown2020language}
Tom Brown, Benjamin Mann, Nick Ryder, Melanie Subbiah, Jared~D Kaplan, Prafulla Dhariwal, Arvind Neelakantan, Pranav Shyam, Girish Sastry, Amanda Askell, et~al.
\newblock Language models are few-shot learners.
\newblock \emph{Advances in neural information processing systems}, 33:\penalty0 1877--1901, 2020.

\bibitem[Cai et~al.(2024)Cai, Zhang, Zhao, Bian, Sugiyama, and Llorens]{cai2024distributional}
Xin-Qiang Cai, Pushi Zhang, Li~Zhao, Jiang Bian, Masashi Sugiyama, and Ashley Llorens.
\newblock Distributional pareto-optimal multi-objective reinforcement learning.
\newblock \emph{Advances in Neural Information Processing Systems}, 36, 2024.

\bibitem[Chai \& Li(2020)Chai and Li]{chai2020human}
Chengliang Chai and Guoliang Li.
\newblock Human-in-the-loop techniques in machine learning.
\newblock \emph{IEEE Data Eng. Bull.}, 43\penalty0 (3):\penalty0 37--52, 2020.

\bibitem[Chen et~al.(2022)Chen, Wu, Wang, Feng, Jiang, Lu, McAleer, Dong, Zhu, and Yang]{chen2022towards}
Yuanpei Chen, Tianhao Wu, Shengjie Wang, Xidong Feng, Jiechuan Jiang, Zongqing Lu, Stephen McAleer, Hao Dong, Song-Chun Zhu, and Yaodong Yang.
\newblock Towards human-level bimanual dexterous manipulation with reinforcement learning.
\newblock \emph{Advances in Neural Information Processing Systems}, 35:\penalty0 5150--5163, 2022.

\bibitem[Christiano et~al.(2017)Christiano, Leike, Brown, Martic, Legg, and Amodei]{christiano2017deep}
Paul~F Christiano, Jan Leike, Tom Brown, Miljan Martic, Shane Legg, and Dario Amodei.
\newblock Deep reinforcement learning from human preferences.
\newblock \emph{Advances in neural information processing systems}, 30, 2017.

\bibitem[Clevert(2015)]{clevert2015fast}
Djork-Arn{\'e} Clevert.
\newblock Fast and accurate deep network learning by exponential linear units (elus).
\newblock \emph{arXiv preprint arXiv:1511.07289}, 2015.

\bibitem[Dang et~al.(2023)Dang, Edelkamp, and Ribault]{dang2023clip}
Xuzhe Dang, Stefan Edelkamp, and Nicolas Ribault.
\newblock Clip-motion: Learning reward functions for robotic actions using consecutive observations.
\newblock \emph{arXiv preprint arXiv:2311.03485}, 2023.

\bibitem[Daniel et~al.(2015)Daniel, Kroemer, Viering, Metz, and Peters]{daniel2015active}
Christian Daniel, Oliver Kroemer, Malte Viering, Jan Metz, and Jan Peters.
\newblock Active reward learning with a novel acquisition function.
\newblock \emph{Autonomous Robots}, 39:\penalty0 389--405, 2015.

\bibitem[Das et~al.(2021)Das, Bechtle, Davchev, Jayaraman, Rai, and Meier]{das2021model}
Neha Das, Sarah Bechtle, Todor Davchev, Dinesh Jayaraman, Akshara Rai, and Franziska Meier.
\newblock Model-based inverse reinforcement learning from visual demonstrations.
\newblock In \emph{Conference on Robot Learning}, pp.\  1930--1942. PMLR, 2021.

\bibitem[Devidze et~al.(2022)Devidze, Kamalaruban, and Singla]{devidze2022exploration}
Rati Devidze, Parameswaran Kamalaruban, and Adish Singla.
\newblock Exploration-guided reward shaping for reinforcement learning under sparse rewards.
\newblock \emph{Advances in Neural Information Processing Systems}, 35:\penalty0 5829--5842, 2022.

\bibitem[Devlin \& Kudenko(2012)Devlin and Kudenko]{devlin2012dynamic}
Sam~Michael Devlin and Daniel Kudenko.
\newblock Dynamic potential-based reward shaping.
\newblock In \emph{11th International Conference on Autonomous Agents and Multiagent Systems (AAMAS 2012)}, pp.\  433--440. IFAAMAS, 2012.

\bibitem[Ding et~al.(2023)Ding, Zhang, Paxton, and Zhang]{ding2023task}
Yan Ding, Xiaohan Zhang, Chris Paxton, and Shiqi Zhang.
\newblock Task and motion planning with large language models for object rearrangement.
\newblock In \emph{2023 IEEE/RSJ International Conference on Intelligent Robots and Systems (IROS)}, pp.\  2086--2092. IEEE, 2023.

\bibitem[Dong et~al.(2023)Dong, Yuan, Hao, Ni, Mu, Zheng, Hu, Lv, Fan, and Hu]{dong2023aligndiff}
Zibin Dong, Yifu Yuan, Jianye Hao, Fei Ni, Yao Mu, Yan Zheng, Yujing Hu, Tangjie Lv, Changjie Fan, and Zhipeng Hu.
\newblock Aligndiff: Aligning diverse human preferences via behavior-customisable diffusion model.
\newblock \emph{arXiv preprint arXiv:2310.02054}, 2023.

\bibitem[Early et~al.(2022)Early, Bewley, Evers, and Ramchurn]{early2022non}
Joseph Early, Tom Bewley, Christine Evers, and Sarvapali Ramchurn.
\newblock Non-markovian reward modelling from trajectory labels via interpretable multiple instance learning.
\newblock \emph{Advances in Neural Information Processing Systems}, 35:\penalty0 27652--27663, 2022.

\bibitem[Eschmann(2021)]{eschmann2021reward}
Jonas Eschmann.
\newblock Reward function design in reinforcement learning.
\newblock \emph{Reinforcement Learning Algorithms: Analysis and Applications}, pp.\  25--33, 2021.

\bibitem[Evans et~al.(2021)Evans, Engelbrecht, and Jordaan]{evans2021reward}
Benjamin Evans, Herman~A Engelbrecht, and Hendrik~W Jordaan.
\newblock Reward signal design for autonomous racing.
\newblock In \emph{2021 20th International Conference on Advanced Robotics (ICAR)}, pp.\  455--460. IEEE, 2021.

\bibitem[Faldor et~al.(2024)Faldor, Zhang, Cully, and Clune]{faldor2024omni}
Maxence Faldor, Jenny Zhang, Antoine Cully, and Jeff Clune.
\newblock Omni-epic: Open-endedness via models of human notions of interestingness with environments programmed in code.
\newblock \emph{arXiv preprint arXiv:2405.15568}, 2024.

\bibitem[F{\"u}rnkranz et~al.(2012)F{\"u}rnkranz, H{\"u}llermeier, Cheng, and Park]{furnkranz2012preference}
Johannes F{\"u}rnkranz, Eyke H{\"u}llermeier, Weiwei Cheng, and Sang-Hyeun Park.
\newblock Preference-based reinforcement learning: a formal framework and a policy iteration algorithm.
\newblock \emph{Machine learning}, 89:\penalty0 123--156, 2012.

\bibitem[Gil et~al.(2019)Gil, Honaker, Gupta, Ma, D'Orazio, Garijo, Gadewar, Yang, and Jahanshad]{gil2019towards}
Yolanda Gil, James Honaker, Shikhar Gupta, Yibo Ma, Vito D'Orazio, Daniel Garijo, Shruti Gadewar, Qifan Yang, and Neda Jahanshad.
\newblock Towards human-guided machine learning.
\newblock In \emph{Proceedings of the 24th international conference on intelligent user interfaces}, pp.\  614--624, 2019.

\bibitem[Goyal et~al.(2019)Goyal, Niekum, and Mooney]{goyal2019using}
Prasoon Goyal, Scott Niekum, and Raymond~J Mooney.
\newblock Using natural language for reward shaping in reinforcement learning.
\newblock \emph{arXiv preprint arXiv:1903.02020}, 2019.

\bibitem[Grze{\'s}(2017)]{grzes2017reward}
Marek Grze{\'s}.
\newblock Reward shaping in episodic reinforcement learning.
\newblock In \emph{Proceedings of the 16th Conference on Autonomous Agents and MultiAgent Systems}, pp.\  565--573, 2017.

\bibitem[Grzes \& Kudenko(2008)Grzes and Kudenko]{grzes2008plan}
Marek Grzes and Daniel Kudenko.
\newblock Plan-based reward shaping for reinforcement learning.
\newblock In \emph{2008 4th International IEEE Conference Intelligent Systems}, volume~2, pp.\  10--22. IEEE, 2008.

\bibitem[Gupta et~al.(2022)Gupta, Pacchiano, Zhai, Kakade, and Levine]{gupta2022unpacking}
Abhishek Gupta, Aldo Pacchiano, Yuexiang Zhai, Sham Kakade, and Sergey Levine.
\newblock Unpacking reward shaping: Understanding the benefits of reward engineering on sample complexity.
\newblock \emph{Advances in Neural Information Processing Systems}, 35:\penalty0 15281--15295, 2022.

\bibitem[Ha et~al.(2023)Ha, Florence, and Song]{ha2023scaling}
Huy Ha, Pete Florence, and Shuran Song.
\newblock Scaling up and distilling down: Language-guided robot skill acquisition.
\newblock In \emph{Conference on Robot Learning}, pp.\  3766--3777. PMLR, 2023.

\bibitem[Hadfield-Menell et~al.(2016)Hadfield-Menell, Russell, Abbeel, and Dragan]{hadfield2016cooperative}
Dylan Hadfield-Menell, Stuart~J Russell, Pieter Abbeel, and Anca Dragan.
\newblock Cooperative inverse reinforcement learning.
\newblock \emph{Advances in neural information processing systems}, 29, 2016.

\bibitem[Hao et~al.(2023)Hao, Gu, Ma, Hong, Wang, Wang, and Hu]{hao2023reasoning}
Shibo Hao, Yi~Gu, Haodi Ma, Joshua Hong, Zhen Wang, Daisy Wang, and Zhiting Hu.
\newblock Reasoning with language model is planning with world model.
\newblock In \emph{Proceedings of the 2023 Conference on Empirical Methods in Natural Language Processing}, pp.\  8154--8173, 2023.

\bibitem[Hayes et~al.(2022)Hayes, R{\u{a}}dulescu, Bargiacchi, K{\"a}llstr{\"o}m, Macfarlane, Reymond, Verstraeten, Zintgraf, Dazeley, Heintz, et~al.]{hayes2022practical}
Conor~F Hayes, Roxana R{\u{a}}dulescu, Eugenio Bargiacchi, Johan K{\"a}llstr{\"o}m, Matthew Macfarlane, Mathieu Reymond, Timothy Verstraeten, Luisa~M Zintgraf, Richard Dazeley, Fredrik Heintz, et~al.
\newblock A practical guide to multi-objective reinforcement learning and planning.
\newblock \emph{Autonomous Agents and Multi-Agent Systems}, 36\penalty0 (1):\penalty0 26, 2022.

\bibitem[Hejna~III \& Sadigh(2023)Hejna~III and Sadigh]{hejna2023few}
Donald~Joseph Hejna~III and Dorsa Sadigh.
\newblock Few-shot preference learning for human-in-the-loop rl.
\newblock In \emph{Conference on Robot Learning}, pp.\  2014--2025. PMLR, 2023.

\bibitem[Hendrycks \& Gimpel(2016)Hendrycks and Gimpel]{hendrycks2016gaussian}
Dan Hendrycks and Kevin Gimpel.
\newblock Gaussian error linear units (gelus).
\newblock \emph{arXiv preprint arXiv:1606.08415}, 2016.

\bibitem[Hu et~al.(2020)Hu, Wang, Jia, Wang, Chen, Hao, Wu, and Fan]{hu2020learning}
Yujing Hu, Weixun Wang, Hangtian Jia, Yixiang Wang, Yingfeng Chen, Jianye Hao, Feng Wu, and Changjie Fan.
\newblock Learning to utilize shaping rewards: A new approach of reward shaping.
\newblock \emph{Advances in Neural Information Processing Systems}, 33:\penalty0 15931--15941, 2020.

\bibitem[Huang et~al.(2022)Huang, Abdolmaleki, Vezzani, Brakel, Mankowitz, Neunert, Bohez, Tassa, Heess, Riedmiller, et~al.]{huang2022constrained}
Sandy Huang, Abbas Abdolmaleki, Giulia Vezzani, Philemon Brakel, Daniel~J Mankowitz, Michael Neunert, Steven Bohez, Yuval Tassa, Nicolas Heess, Martin Riedmiller, et~al.
\newblock A constrained multi-objective reinforcement learning framework.
\newblock In \emph{Conference on Robot Learning}, pp.\  883--893. PMLR, 2022.

\bibitem[Huang et~al.(2023)Huang, Jiang, Dong, Qiao, Gao, and Li]{huang2023instruct2act}
Siyuan Huang, Zhengkai Jiang, Hao Dong, Yu~Qiao, Peng Gao, and Hongsheng Li.
\newblock Instruct2act: Mapping multi-modality instructions to robotic actions with large language model.
\newblock \emph{arXiv preprint arXiv:2305.11176}, 2023.

\bibitem[Huang et~al.(2024)Huang, Xia, Shah, Driess, Zeng, Lu, Florence, Mordatch, Levine, Hausman, et~al.]{huang2024grounded}
Wenlong Huang, Fei Xia, Dhruv Shah, Danny Driess, Andy Zeng, Yao Lu, Pete Florence, Igor Mordatch, Sergey Levine, Karol Hausman, et~al.
\newblock Grounded decoding: Guiding text generation with grounded models for embodied agents.
\newblock \emph{Advances in Neural Information Processing Systems}, 36, 2024.

\bibitem[Hussein et~al.(2017)Hussein, Elyan, Gaber, and Jayne]{hussein2017deep}
Ahmed Hussein, Eyad Elyan, Mohamed~Medhat Gaber, and Chrisina Jayne.
\newblock Deep reward shaping from demonstrations.
\newblock In \emph{2017 International Joint Conference on Neural Networks (IJCNN)}, pp.\  510--517. IEEE, 2017.

\bibitem[Ibarz et~al.(2018)Ibarz, Leike, Pohlen, Irving, Legg, and Amodei]{ibarz2018reward}
Borja Ibarz, Jan Leike, Tobias Pohlen, Geoffrey Irving, Shane Legg, and Dario Amodei.
\newblock Reward learning from human preferences and demonstrations in atari.
\newblock \emph{Advances in neural information processing systems}, 31, 2018.

\bibitem[Ji et~al.(2024)Ji, Zhang, Sajda, and Chen]{ji2024enabling}
Zhengran Ji, Lingyu Zhang, Paul Sajda, and Boyuan Chen.
\newblock Enabling multi-robot collaboration from single-human guidance.
\newblock \emph{arXiv preprint arXiv:2409.19831}, 2024.

\bibitem[Kim et~al.(2023)Kim, Park, Shin, Lee, Abbeel, and Lee]{kim2023preference}
Changyeon Kim, Jongjin Park, Jinwoo Shin, Honglak Lee, Pieter Abbeel, and Kimin Lee.
\newblock Preference transformer: Modeling human preferences using transformers for rl.
\newblock \emph{arXiv preprint arXiv:2303.00957}, 2023.

\bibitem[Kim et~al.(2024{\natexlab{a}})Kim, Kwon, Kim, Lee, and Oh]{kim2024stage}
Dohyeong Kim, Hyeokjin Kwon, Junseok Kim, Gunmin Lee, and Songhwai Oh.
\newblock Stage-wise reward shaping for acrobatic robots: A constrained multi-objective reinforcement learning approach.
\newblock \emph{arXiv preprint arXiv:2409.15755}, 2024{\natexlab{a}}.

\bibitem[Kim et~al.(2024{\natexlab{b}})Kim, Lee, Kang, Oh, Chong, and Yun]{kim2024preference}
Minu Kim, Yongsik Lee, Sehyeok Kang, Jihwan Oh, Song Chong, and Se-Young Yun.
\newblock Preference alignment with flow matching.
\newblock \emph{arXiv preprint arXiv:2405.19806}, 2024{\natexlab{b}}.

\bibitem[Knox \& Stone(2008)Knox and Stone]{knox2008tamer}
W~Bradley Knox and Peter Stone.
\newblock Tamer: Training an agent manually via evaluative reinforcement.
\newblock In \emph{2008 7th IEEE international conference on development and learning}, pp.\  292--297. IEEE, 2008.

\bibitem[Knox et~al.(2022)Knox, Hatgis-Kessell, Booth, Niekum, Stone, and Allievi]{knox2022models}
W~Bradley Knox, Stephane Hatgis-Kessell, Serena Booth, Scott Niekum, Peter Stone, and Alessandro Allievi.
\newblock Models of human preference for learning reward functions.
\newblock \emph{arXiv preprint arXiv:2206.02231}, 2022.

\bibitem[Kumar et~al.(2023)Kumar, Zamora, Hansen, Jangir, and Wang]{kumar2023graph}
Sateesh Kumar, Jonathan Zamora, Nicklas Hansen, Rishabh Jangir, and Xiaolong Wang.
\newblock Graph inverse reinforcement learning from diverse videos.
\newblock In \emph{Conference on Robot Learning}, pp.\  55--66. PMLR, 2023.

\bibitem[Kyriakis \& Deshmukh(2022)Kyriakis and Deshmukh]{kyriakis2022pareto}
Panagiotis Kyriakis and Jyotirmoy Deshmukh.
\newblock Pareto policy adaptation.
\newblock In \emph{International Conference on Learning Representations}, volume 2022, 2022.

\bibitem[Lee et~al.(2024)Lee, Phatale, Mansoor, Mesnard, Ferret, Lu, Bishop, Hall, Carbune, Rastogi, et~al.]{leerlaif}
Harrison Lee, Samrat Phatale, Hassan Mansoor, Thomas Mesnard, Johan Ferret, Kellie~Ren Lu, Colton Bishop, Ethan Hall, Victor Carbune, Abhinav Rastogi, et~al.
\newblock Rlaif vs. rlhf: Scaling reinforcement learning from human feedback with ai feedback.
\newblock In \emph{Forty-first International Conference on Machine Learning}, 2024.

\bibitem[Lee et~al.(2021{\natexlab{a}})Lee, Smith, Dragan, and Abbeel]{lee2021b}
Kimin Lee, Laura Smith, Anca Dragan, and Pieter Abbeel.
\newblock B-pref: Benchmarking preference-based reinforcement learning.
\newblock \emph{arXiv preprint arXiv:2111.03026}, 2021{\natexlab{a}}.

\bibitem[Lee et~al.(2021{\natexlab{b}})Lee, Smith, and Abbeel]{lee2021pebble}
Kimin Lee, Laura~M Smith, and Pieter Abbeel.
\newblock Pebble: Feedback-efficient interactive reinforcement learning via relabeling experience and unsupervised pre-training.
\newblock In \emph{International Conference on Machine Learning}, pp.\  6152--6163. PMLR, 2021{\natexlab{b}}.

\bibitem[Liang et~al.(2023)Liang, Huang, Xia, Xu, Hausman, Ichter, Florence, and Zeng]{liang2023code}
Jacky Liang, Wenlong Huang, Fei Xia, Peng Xu, Karol Hausman, Brian Ichter, Pete Florence, and Andy Zeng.
\newblock Code as policies: Language model programs for embodied control.
\newblock In \emph{2023 IEEE International Conference on Robotics and Automation (ICRA)}, pp.\  9493--9500. IEEE, 2023.

\bibitem[Liang et~al.(2024)Liang, Wang, Wang, Bastani, Jayaraman, and Ma]{liang2024eurekaverse}
William Liang, Sam Wang, Hung-Ju Wang, Osbert Bastani, Dinesh Jayaraman, and Yecheng~Jason Ma.
\newblock Environment curriculum generation via large language models.
\newblock In \emph{Conference on Robot Learning (CoRL)}, 2024.

\bibitem[Lin et~al.(2023)Lin, Agia, Migimatsu, Pavone, and Bohg]{lin2023text2motion}
Kevin Lin, Christopher Agia, Toki Migimatsu, Marco Pavone, and Jeannette Bohg.
\newblock Text2motion: From natural language instructions to feasible plans.
\newblock \emph{Autonomous Robots}, 47\penalty0 (8):\penalty0 1345--1365, 2023.

\bibitem[Liu et~al.(2023)Liu, Jiang, Zhang, Liu, Zhang, Biswas, and Stone]{liu2023llm+}
Bo~Liu, Yuqian Jiang, Xiaohan Zhang, Qiang Liu, Shiqi Zhang, Joydeep Biswas, and Peter Stone.
\newblock Llm+ p: Empowering large language models with optimal planning proficiency.
\newblock \emph{arXiv preprint arXiv:2304.11477}, 2023.

\bibitem[Liu et~al.(2020)]{liu2020learning}
Fei Liu et~al.
\newblock Learning to summarize from human feedback.
\newblock In \emph{Proceedings of the 58th Annual Meeting of the Association for Computational Linguistics}, 2020.

\bibitem[Liu et~al.(2024)Liu, Du, Bai, Lyu, and Li]{liu2024pearl}
Runze Liu, Yali Du, Fengshuo Bai, Jiafei Lyu, and Xiu Li.
\newblock Pearl: zero-shot cross-task preference alignment and robust reward learning for robotic manipulation.
\newblock In \emph{International Conference on Machine Learning}, 2024.

\bibitem[Ma et~al.(2023)Ma, Liang, Wang, Huang, Bastani, Jayaraman, Zhu, Fan, and Anandkumar]{ma2023eureka}
Yecheng~Jason Ma, William Liang, Guanzhi Wang, De-An Huang, Osbert Bastani, Dinesh Jayaraman, Yuke Zhu, Linxi Fan, and Anima Anandkumar.
\newblock Eureka: Human-level reward design via coding large language models.
\newblock \emph{arXiv preprint arXiv: Arxiv-2310.12931}, 2023.

\bibitem[Ma et~al.(2024)Ma, Liang, Wang, Wang, Zhu, Fan, Bastani, and Jayaraman]{ma2024dreureka}
Yecheng~Jason Ma, William Liang, Hungju Wang, Sam Wang, Yuke Zhu, Linxi Fan, Osbert Bastani, and Dinesh Jayaraman.
\newblock Dreureka: Language model guided sim-to-real transfer.
\newblock In \emph{Robotics: Science and Systems (RSS)}, 2024.

\bibitem[MacGlashan et~al.(2016)MacGlashan, Littman, Roberts, Loftin, Peng, and Taylor]{macglashan2016convergent}
James MacGlashan, Michael~L Littman, David~L Roberts, Robert Loftin, Bei Peng, and Matthew~E Taylor.
\newblock Convergent actor critic by humans.
\newblock In \emph{International Conference on Intelligent Robots and Systems}, 2016.

\bibitem[Makoviychuk et~al.(2021)Makoviychuk, Wawrzyniak, Guo, Lu, Storey, Macklin, Hoeller, Rudin, Allshire, Handa, et~al.]{makoviychuk2021isaac}
Viktor Makoviychuk, Lukasz Wawrzyniak, Yunrong Guo, Michelle Lu, Kier Storey, Miles Macklin, David Hoeller, Nikita Rudin, Arthur Allshire, Ankur Handa, et~al.
\newblock Isaac gym: High performance gpu-based physics simulation for robot learning.
\newblock \emph{arXiv preprint arXiv:2108.10470}, 2021.

\bibitem[Margolis et~al.(2024)Margolis, Yang, Paigwar, Chen, and Agrawal]{margolis2024rapid}
Gabriel~B Margolis, Ge~Yang, Kartik Paigwar, Tao Chen, and Pulkit Agrawal.
\newblock Rapid locomotion via reinforcement learning.
\newblock \emph{The International Journal of Robotics Research}, 43\penalty0 (4):\penalty0 572--587, 2024.

\bibitem[Marom \& Rosman(2018)Marom and Rosman]{marom2018belief}
Ofir Marom and Benjamin Rosman.
\newblock Belief reward shaping in reinforcement learning.
\newblock In \emph{Proceedings of the AAAI conference on artificial intelligence}, volume~32, 2018.

\bibitem[Memarian et~al.(2021)Memarian, Goo, Lioutikov, Niekum, and Topcu]{memarian2021self}
Farzan Memarian, Wonjoon Goo, Rudolf Lioutikov, Scott Niekum, and Ufuk Topcu.
\newblock Self-supervised online reward shaping in sparse-reward environments.
\newblock In \emph{2021 IEEE/RSJ International Conference on Intelligent Robots and Systems (IROS)}, pp.\  2369--2375. IEEE, 2021.

\bibitem[Nakano et~al.(2021)Nakano, Hilton, Balaji, Wu, Ouyang, Kim, Hesse, Jain, Kosaraju, Saunders, et~al.]{nakano2021webgpt}
Reiichiro Nakano, Jacob Hilton, Suchir Balaji, Jeff Wu, Long Ouyang, Christina Kim, Christopher Hesse, Shantanu Jain, Vineet Kosaraju, William Saunders, et~al.
\newblock Webgpt: Browser-assisted question-answering with human feedback.
\newblock \emph{arXiv preprint arXiv:2112.09332}, 2021.

\bibitem[Ng et~al.(1999)Ng, Harada, and Russell]{ng1999policy}
Andrew~Y Ng, Daishi Harada, and Stuart Russell.
\newblock Policy invariance under reward transformations: Theory and application to reward shaping.
\newblock In \emph{Icml}, volume~99, pp.\  278--287, 1999.

\bibitem[Ng et~al.(2000)Ng, Russell, et~al.]{ng2000algorithms}
Andrew~Y Ng, Stuart Russell, et~al.
\newblock Algorithms for inverse reinforcement learning.
\newblock In \emph{Icml}, volume~1, pp.\ ~2, 2000.

\bibitem[{Octo Model Team} et~al.(2024){Octo Model Team}, Ghosh, Walke, Pertsch, Black, Mees, Dasari, Hejna, Xu, Luo, Kreiman, Tan, Chen, Sanketi, Vuong, Xiao, Sadigh, Finn, and Levine]{octo_2023}
{Octo Model Team}, Dibya Ghosh, Homer Walke, Karl Pertsch, Kevin Black, Oier Mees, Sudeep Dasari, Joey Hejna, Charles Xu, Jianlan Luo, Tobias Kreiman, {You Liang} Tan, Lawrence~Yunliang Chen, Pannag Sanketi, Quan Vuong, Ted Xiao, Dorsa Sadigh, Chelsea Finn, and Sergey Levine.
\newblock Octo: An open-source generalist robot policy.
\newblock In \emph{Proceedings of Robotics: Science and Systems}, Delft, Netherlands, 2024.

\bibitem[Ouyang et~al.(2022)Ouyang, Wu, Jiang, Almeida, Wainwright, Mishkin, Zhang, Agarwal, Slama, Ray, et~al.]{ouyang2022training}
Long Ouyang, Jeffrey Wu, Xu~Jiang, Diogo Almeida, Carroll Wainwright, Pamela Mishkin, Chong Zhang, Sandhini Agarwal, Katarina Slama, Alex Ray, et~al.
\newblock Training language models to follow instructions with human feedback.
\newblock \emph{Advances in neural information processing systems}, 35:\penalty0 27730--27744, 2022.

\bibitem[Park et~al.(2022)Park, Seo, Shin, Lee, Abbeel, and Lee]{park2022surf}
Jongjin Park, Younggyo Seo, Jinwoo Shin, Honglak Lee, Pieter Abbeel, and Kimin Lee.
\newblock Surf: Semi-supervised reward learning with data augmentation for feedback-efficient preference-based reinforcement learning.
\newblock \emph{arXiv preprint arXiv:2203.10050}, 2022.

\bibitem[Pomerleau(1988)]{pomerleau1988alvinn}
Dean~A Pomerleau.
\newblock Alvinn: An autonomous land vehicle in a neural network.
\newblock \emph{Advances in neural information processing systems}, 1, 1988.

\bibitem[Radford(2018)]{radford2018improving}
Alec Radford.
\newblock Improving language understanding by generative pre-training.
\newblock 2018.

\bibitem[Ratner et~al.(2018)Ratner, Hadfield-Menell, and Dragan]{ratner2018simplifying}
Ellis Ratner, Dylan Hadfield-Menell, and Anca~D Dragan.
\newblock Simplifying reward design through divide-and-conquer.
\newblock \emph{arXiv preprint arXiv:1806.02501}, 2018.

\bibitem[Robotics(2023)]{unitree_rl_gym}
Unitree Robotics.
\newblock Unitree rl gym.
\newblock \url{https://github.com/unitreerobotics/unitree_rl_gym}, 2023.
\newblock Accessed: 2025-01-22.

\bibitem[Rudin et~al.(2022)Rudin, Hoeller, Reist, and Hutter]{rudin2022learning}
Nikita Rudin, David Hoeller, Philipp Reist, and Marco Hutter.
\newblock Learning to walk in minutes using massively parallel deep reinforcement learning.
\newblock In \emph{Conference on Robot Learning}, pp.\  91--100. PMLR, 2022.

\bibitem[Ryu et~al.(2024)Ryu, Liao, Li, Sreenath, and Mehr]{ryu2024curricullm}
Kanghyun Ryu, Qiayuan Liao, Zhongyu Li, Koushil Sreenath, and Negar Mehr.
\newblock Curricullm: Automatic task curricula design for learning complex robot skills using large language models.
\newblock \emph{arXiv preprint arXiv:2409.18382}, 2024.

\bibitem[Saran et~al.(2021)Saran, Zhang, Short, and Niekum]{saran2021efficiently}
Akanksha Saran, Ruohan Zhang, E~Short, and Scott Niekum.
\newblock Efficiently guiding imitation learning algorithms with human gaze.
\newblock In \emph{International Conference on Autonomous Agents and Multiagent Systems}, 2021.

\bibitem[Schaal(1996)]{schaal1996learning}
Stefan Schaal.
\newblock Learning from demonstration.
\newblock \emph{Advances in neural information processing systems}, 9, 1996.

\bibitem[Schulman et~al.(2017)Schulman, Wolski, Dhariwal, Radford, and Klimov]{schulman2017proximal}
John Schulman, Filip Wolski, Prafulla Dhariwal, Alec Radford, and Oleg Klimov.
\newblock Proximal policy optimization algorithms.
\newblock \emph{arXiv preprint arXiv:1707.06347}, 2017.

\bibitem[ShadowRobot(2005)]{ShadowRobot2005}
ShadowRobot.
\newblock Shadowrobot dexterous hand, 2005.
\newblock URL \url{https://www.shadowrobot.com/products/dexterous-hand/}.
\newblock Accessed: 2025-01-31.

\bibitem[Sheidlower et~al.(2022)Sheidlower, Short, and Moore]{sheidlower2022environment}
Isaac Sheidlower, Elaine~Schaertl Short, and Allison Moore.
\newblock Environment guided interactive reinforcement learning: Learning from binary feedback in high-dimensional robot task environments.
\newblock In \emph{Proceedings of the 21st International Conference on Autonomous Agents and Multiagent Systems}, pp.\  1726--1728, 2022.

\bibitem[Silver et~al.(2024)Silver, Dan, Srinivas, Tenenbaum, Kaelbling, and Katz]{silver2024generalized}
Tom Silver, Soham Dan, Kavitha Srinivas, Joshua~B Tenenbaum, Leslie Kaelbling, and Michael Katz.
\newblock Generalized planning in pddl domains with pretrained large language models.
\newblock In \emph{Proceedings of the AAAI Conference on Artificial Intelligence}, volume~38, pp.\  20256--20264, 2024.

\bibitem[Singh et~al.(2023)Singh, Blukis, Mousavian, Goyal, Xu, Tremblay, Fox, Thomason, and Garg]{singh2023progprompt}
Ishika Singh, Valts Blukis, Arsalan Mousavian, Ankit Goyal, Danfei Xu, Jonathan Tremblay, Dieter Fox, Jesse Thomason, and Animesh Garg.
\newblock Progprompt: Generating situated robot task plans using large language models.
\newblock In \emph{2023 IEEE International Conference on Robotics and Automation (ICRA)}, pp.\  11523--11530. IEEE, 2023.

\bibitem[Sorg et~al.(2010)Sorg, Lewis, and Singh]{sorg2010reward}
Jonathan Sorg, Richard~L Lewis, and Satinder Singh.
\newblock Reward design via online gradient ascent.
\newblock \emph{Advances in Neural Information Processing Systems}, 23, 2010.

\bibitem[Sutton \& Barto(2018)Sutton and Barto]{sutton2018reinforcement}
Richard~S Sutton and Andrew~G Barto.
\newblock \emph{Reinforcement learning: An introduction}.
\newblock MIT press, 2018.

\bibitem[Szot et~al.(2023)Szot, Schwarzer, Agrawal, Mazoure, Metcalf, Talbott, Mackraz, Hjelm, and Toshev]{szot2023large}
Andrew Szot, Max Schwarzer, Harsh Agrawal, Bogdan Mazoure, Rin Metcalf, Walter Talbott, Natalie Mackraz, R~Devon Hjelm, and Alexander~T Toshev.
\newblock Large language models as generalizable policies for embodied tasks.
\newblock In \emph{The Twelfth International Conference on Learning Representations}, 2023.

\bibitem[Tang et~al.(2023)Tang, Yu, Tan, Zen, Faust, and Harada]{tang2023saytap}
Yujin Tang, Wenhao Yu, Jie Tan, Heiga Zen, Aleksandra Faust, and Tatsuya Harada.
\newblock Saytap: Language to quadrupedal locomotion.
\newblock In \emph{Conference on Robot Learning}, pp.\  3556--3570. PMLR, 2023.

\bibitem[Tang et~al.(2021)Tang, Kim, and Ha]{tang2021learning}
Zuoxin Tang, Donghyun Kim, and Sehoon Ha.
\newblock Learning agile motor skills on quadrupedal robots using curriculum learning.
\newblock In \emph{International Conference on Robot Intelligence Technology and Applications}, volume~3, 2021.

\bibitem[Van~Moffaert et~al.(2013)Van~Moffaert, Drugan, and Now{\'e}]{van2013scalarized}
Kristof Van~Moffaert, Madalina~M Drugan, and Ann Now{\'e}.
\newblock Scalarized multi-objective reinforcement learning: Novel design techniques.
\newblock In \emph{2013 IEEE symposium on adaptive dynamic programming and reinforcement learning (ADPRL)}, pp.\  191--199. IEEE, 2013.

\bibitem[Venkataraman et~al.(2024)Venkataraman, Wang, Wang, Erickson, and Held]{venkataraman2024real}
Sreyas Venkataraman, Yufei Wang, Ziyu Wang, Zackory Erickson, and David Held.
\newblock Real-world offline reinforcement learning from vision language model feedback.
\newblock \emph{arXiv preprint arXiv:2411.05273}, 2024.

\bibitem[Verma \& Metcalf(2022)Verma and Metcalf]{verma2022symbol}
Mudit Verma and Katherine Metcalf.
\newblock Symbol guided hindsight priors for reward learning from human preferences.
\newblock \emph{arXiv preprint arXiv:2210.09151}, 2022.

\bibitem[Wang et~al.(2023{\natexlab{a}})Wang, Xie, Jiang, Mandlekar, Xiao, Zhu, Fan, and Anandkumar]{wang2023voyager}
Guanzhi Wang, Yuqi Xie, Yunfan Jiang, Ajay Mandlekar, Chaowei Xiao, Yuke Zhu, Linxi Fan, and Anima Anandkumar.
\newblock Voyager: An open-ended embodied agent with large language models.
\newblock \emph{arXiv preprint arXiv: Arxiv-2305.16291}, 2023{\natexlab{a}}.

\bibitem[Wang et~al.(2023{\natexlab{b}})Wang, Ling, Yuan, Shridhar, Bao, Qin, Wang, Xu, and Wang]{wang2023gen}
Lirui Wang, Yiyang Ling, Zhecheng Yuan, Mohit Shridhar, Chen Bao, Yuzhe Qin, Bailin Wang, Huazhe Xu, and Xiaolong Wang.
\newblock Gensim: Generating robotic simulation tasks via large language models.
\newblock In \emph{Arxiv}, 2023{\natexlab{b}}.

\bibitem[Wang et~al.(2023{\natexlab{c}})Wang, Ling, Yuan, Shridhar, Bao, Qin, Wang, Xu, and Wang]{wang2023gensim}
Lirui Wang, Yiyang Ling, Zhecheng Yuan, Mohit Shridhar, Chen Bao, Yuzhe Qin, Bailin Wang, Huazhe Xu, and Xiaolong Wang.
\newblock Gensim: Generating robotic simulation tasks via large language models.
\newblock \emph{arXiv preprint arXiv:2310.01361}, 2023{\natexlab{c}}.

\bibitem[Wang et~al.(2023{\natexlab{d}})Wang, Xian, Chen, Wang, Wang, Fragkiadaki, Erickson, Held, and Gan]{wang2023robogen}
Yufei Wang, Zhou Xian, Feng Chen, Tsun-Hsuan Wang, Yian Wang, Katerina Fragkiadaki, Zackory Erickson, David Held, and Chuang Gan.
\newblock Robogen: Towards unleashing infinite data for automated robot learning via generative simulation.
\newblock \emph{arXiv preprint arXiv:2311.01455}, 2023{\natexlab{d}}.

\bibitem[Wang et~al.(2024)Wang, Sun, Zhang, Xian, Biyik, Held, and Erickson]{wang2024rl}
Yufei Wang, Zhanyi Sun, Jesse Zhang, Zhou Xian, Erdem Biyik, David Held, and Zackory Erickson.
\newblock Rl-vlm-f: Reinforcement learning from vision language foundation model feedback.
\newblock \emph{arXiv preprint arXiv:2402.03681}, 2024.

\bibitem[Wang et~al.(2023{\natexlab{e}})Wang, Cai, Chen, Liu, Ma, Liang, and CraftJarvis]{wang2023describe}
Zihao Wang, Shaofei Cai, Guanzhou Chen, Anji Liu, Xiaojian Ma, Yitao Liang, and Team CraftJarvis.
\newblock Describe, explain, plan and select: interactive planning with large language models enables open-world multi-task agents.
\newblock In \emph{Proceedings of the 37th International Conference on Neural Information Processing Systems}, pp.\  34153--34189, 2023{\natexlab{e}}.

\bibitem[Warnell et~al.(2018)Warnell, Waytowich, Lawhern, and Stone]{warnell2018deep}
Garrett Warnell, Nicholas Waytowich, Vernon Lawhern, and Peter Stone.
\newblock Deep tamer: Interactive agent shaping in high-dimensional state spaces.
\newblock In \emph{Proceedings of the AAAI conference on artificial intelligence}, volume~32, 2018.

\bibitem[Waswani et~al.(2017)Waswani, Shazeer, Parmar, Uszkoreit, Jones, Gomez, Kaiser, and Polosukhin]{waswani2017attention}
A~Waswani, N~Shazeer, N~Parmar, J~Uszkoreit, L~Jones, A~Gomez, L~Kaiser, and I~Polosukhin.
\newblock Attention is all you need.
\newblock In \emph{NIPS}, 2017.

\bibitem[Wilson et~al.(2012)Wilson, Fern, and Tadepalli]{wilson2012bayesian}
Aaron Wilson, Alan Fern, and Prasad Tadepalli.
\newblock A bayesian approach for policy learning from trajectory preference queries.
\newblock \emph{Advances in neural information processing systems}, 25, 2012.

\bibitem[Wirth et~al.(2016)Wirth, F{\"u}rnkranz, and Neumann]{wirth2016model}
Christian Wirth, Johannes F{\"u}rnkranz, and Gerhard Neumann.
\newblock Model-free preference-based reinforcement learning.
\newblock In \emph{Proceedings of the AAAI conference on artificial intelligence}, volume~30, 2016.

\bibitem[Wirth et~al.(2017)Wirth, Akrour, Neumann, and F{\"u}rnkranz]{wirth2017survey}
Christian Wirth, Riad Akrour, Gerhard Neumann, and Johannes F{\"u}rnkranz.
\newblock A survey of preference-based reinforcement learning methods.
\newblock \emph{Journal of Machine Learning Research}, 18\penalty0 (136):\penalty0 1--46, 2017.

\bibitem[Wu et~al.(2021)Wu, Ouyang, Ziegler, Stiennon, Lowe, Leike, and Christiano]{wu2021recursively}
Jeff Wu, Long Ouyang, Daniel~M Ziegler, Nisan Stiennon, Ryan Lowe, Jan Leike, and Paul Christiano.
\newblock Recursively summarizing books with human feedback.
\newblock \emph{arXiv preprint arXiv:2109.10862}, 2021.

\bibitem[Wu et~al.(2022)Wu, Xiao, Sun, Zhang, Ma, and He]{wu2022survey}
Xingjiao Wu, Luwei Xiao, Yixuan Sun, Junhang Zhang, Tianlong Ma, and Liang He.
\newblock A survey of human-in-the-loop for machine learning.
\newblock \emph{Future Generation Computer Systems}, 135:\penalty0 364--381, 2022.

\bibitem[Xia et~al.(2024)Xia, Li, Lee, Scutari, and Chen]{xia2024duke}
Boxi Xia, Bokuan Li, Jacob Lee, Michael Scutari, and Boyuan Chen.
\newblock The duke humanoid: Design and control for energy efficient bipedal locomotion using passive dynamics.
\newblock \emph{arXiv preprint arXiv:2409.19795}, 2024.

\bibitem[Xie et~al.(2024)Xie, Zhao, Wu, Liu, Luo, Zhong, Yang, and Yu]{xie2024textreward}
Tianbao Xie, Siheng Zhao, Chen~Henry Wu, Yitao Liu, Qian Luo, Victor Zhong, Yanchao Yang, and Tao Yu.
\newblock Text2reward: Reward shaping with language models for reinforcement learning.
\newblock In \emph{The Twelfth International Conference on Learning Representations}, 2024.
\newblock URL \url{https://openreview.net/forum?id=tUM39YTRxH}.

\bibitem[Xie(2020)]{xie2020translating}
Yaqi Xie.
\newblock Translating natural language to planning goals with large-language models.
\newblock \emph{The International Journal of Robotics Research}, 2019:\penalty0 1, 2020.

\bibitem[Xu et~al.(2020)Xu, Tian, Ma, Rus, Sueda, and Matusik]{xu2020prediction}
Jie Xu, Yunsheng Tian, Pingchuan Ma, Daniela Rus, Shinjiro Sueda, and Wojciech Matusik.
\newblock Prediction-guided multi-objective reinforcement learning for continuous robot control.
\newblock In \emph{International conference on machine learning}, pp.\  10607--10616. PMLR, 2020.

\bibitem[Yang et~al.(2019)Yang, Sun, and Narasimhan]{yang2019generalized}
Runzhe Yang, Xingyuan Sun, and Karthik Narasimhan.
\newblock A generalized algorithm for multi-objective reinforcement learning and policy adaptation.
\newblock \emph{Advances in neural information processing systems}, 32, 2019.

\bibitem[Yu et~al.(2024)Yu, Lu, Gao, Tan, Yang, Wang, Wu, and Vinitsky]{yu2024few}
Chao Yu, Hong Lu, Jiaxuan Gao, Qixin Tan, Xinting Yang, Yu~Wang, Yi~Wu, and Eugene Vinitsky.
\newblock Few-shot in-context preference learning using large language models.
\newblock \emph{arXiv preprint arXiv:2410.17233}, 2024.

\bibitem[Yu et~al.(2023)Yu, Gileadi, Fu, Kirmani, Lee, Arenas, Chiang, Erez, Hasenclever, Humplik, et~al.]{yu2023language}
Wenhao Yu, Nimrod Gileadi, Chuyuan Fu, Sean Kirmani, Kuang-Huei Lee, Montserrat~Gonzalez Arenas, Hao-Tien~Lewis Chiang, Tom Erez, Leonard Hasenclever, Jan Humplik, et~al.
\newblock Language to rewards for robotic skill synthesis.
\newblock In \emph{Conference on Robot Learning}, pp.\  374--404. PMLR, 2023.

\bibitem[Yuan et~al.(2024)Yuan, Shang, Wang, Wang, Shan, Qi, Zhu, Bai, and Li]{yuan2024preference}
Xinyi Yuan, Zhiwei Shang, Zifan Wang, Chenkai Wang, Zhao Shan, Zhenchao Qi, Meixin Zhu, Chenjia Bai, and Xuelong Li.
\newblock Preference aligned diffusion planner for quadrupedal locomotion control.
\newblock \emph{arXiv preprint arXiv:2410.13586}, 2024.

\bibitem[Zakka et~al.(2022)Zakka, Zeng, Florence, Tompson, Bohg, and Dwibedi]{zakka2022xirl}
Kevin Zakka, Andy Zeng, Pete Florence, Jonathan Tompson, Jeannette Bohg, and Debidatta Dwibedi.
\newblock Xirl: Cross-embodiment inverse reinforcement learning.
\newblock In \emph{Conference on Robot Learning}, pp.\  537--546. PMLR, 2022.

\bibitem[Zhang et~al.(2023)Zhang, Zhang, Pertsch, Liu, Ren, Chang, Sun, and Lim]{zhang2023bootstrap}
Jesse Zhang, Jiahui Zhang, Karl Pertsch, Ziyi Liu, Xiang Ren, Minsuk Chang, Shao-Hua Sun, and Joseph~J Lim.
\newblock Bootstrap your own skills: Learning to solve new tasks with large language model guidance.
\newblock In \emph{Conference on Robot Learning}, pp.\  302--325. PMLR, 2023.

\bibitem[Zhang et~al.(2024{\natexlab{a}})Zhang, Ji, and Chen]{zhang2024crew}
Lingyu Zhang, Zhengran Ji, and Boyuan Chen.
\newblock {CREW}: Facilitating human-ai teaming research.
\newblock \emph{Transactions on Machine Learning Research}, 2024{\natexlab{a}}.

\bibitem[Zhang et~al.(2024{\natexlab{b}})Zhang, Ji, Waytowich, and Chen]{zhang2024guide}
Lingyu Zhang, Zhengran Ji, Nicholas~R Waytowich, and Boyuan Chen.
\newblock {GUIDE}: Real-time human-shaped agents.
\newblock In \emph{The Thirty-eighth Annual Conference on Neural Information Processing Systems}, 2024{\natexlab{b}}.

\bibitem[Zhang et~al.(2019)Zhang, Torabi, Guan, Ballard, and Stone]{zhang2019leveraging}
Ruohan Zhang, Faraz Torabi, Lin Guan, Dana~H Ballard, and Peter Stone.
\newblock Leveraging human guidance for deep reinforcement learning tasks.
\newblock \emph{arXiv preprint arXiv:1909.09906}, 2019.

\bibitem[Zhou \& Small(2021)Zhou and Small]{zhou2021inverse}
Li~Zhou and Kevin Small.
\newblock Inverse reinforcement learning with natural language goals.
\newblock In \emph{Proceedings of the AAAI Conference on Artificial Intelligence}, volume~35, pp.\  11116--11124, 2021.

\bibitem[Zitkovich et~al.(2023)Zitkovich, Yu, Xu, Xu, Xiao, Xia, Wu, Wohlhart, Welker, Wahid, et~al.]{zitkovich2023rt}
Brianna Zitkovich, Tianhe Yu, Sichun Xu, Peng Xu, Ted Xiao, Fei Xia, Jialin Wu, Paul Wohlhart, Stefan Welker, Ayzaan Wahid, et~al.
\newblock Rt-2: Vision-language-action models transfer web knowledge to robotic control.
\newblock In \emph{Conference on Robot Learning}, pp.\  2165--2183. PMLR, 2023.

\bibitem[Zou et~al.(2019)Zou, Ren, Yan, Su, and Zhu]{zou2019reward}
Haosheng Zou, Tongzheng Ren, Dong Yan, Hang Su, and Jun Zhu.
\newblock Reward shaping via meta-learning.
\newblock \emph{arXiv preprint arXiv:1901.09330}, 2019.

\end{thebibliography}
\bibliographystyle{tmlr}

\appendix

\onecolumn
\newenvironment{myverbatim}{\obeylines}{}
\DefineVerbatimEnvironment{mycode}{Verbatim}{breaklines=true, breakanywhere=true}
\appendix

\section{Appendix}
\subsection{Full Prompts}
\label{apdx:prompt}

 In this section,we provide all the prompts for training with LAPP.

 \begin{tcolorbox}[colback=gray!10, colframe=black, title=Prompt for Flat Plane, sharp corners, breakable]
 \begin{myverbatim}
You are a robotics engineer trying to compare pairs of quadruped robot locomotion trajectories and decide which one is better in each pair.
Your feedback of the comparisons will be used as a reward signal (for reinforcement learning) to train a quadruped robot (Unitree Go2) to walk forward at some speed given by the commands, and the velocity range of the speed command is [0.0, 2.2] m/s.
The training method is similar to that in the paper "Deep Reinforcement Learning from Human Preferences", where humans provide preference of trajectories in different pairs of comparisons,

but now you will take the role of the humans to provide feedback on which one trajectory is better in a pair of trajectories.

Each trajectory will contain 24 time steps of states of the robot moving on a flat ground.

The state includes:

1) "commands": the linear velocity command along x axis that the robot needs to follow. its length is 24, standing for the 24 steps of a trajectory. its value range at each step is [0.0, 2.2] m/s. Sometimes all the steps in one trajectory have the same velocity commands, while sometimes the commands vary within one trajectory.

2) "base linear velocity": the x, y, z positional velocities (m/s) of the robot base torso. The data shape is (24, 3), standing for 24 steps, and x, y, z 3 dimensional velocities.

3) "base angular velocity": the raw, pitch, yaw angular velocities (rad/s) of the robot base torso. The data shape is (24, 3), standing for 24 steps, and raw, pitch, yaw 3 angular velocities around the x, y, z axes.

4) "base height": the z position (height) of the robot base torso. The data shape is (24, ), standing for the 24 steps of a trajectory.

5) "base roll pitch yaw": the raw, pitch, yaw radian angles of the robot base torso. The data shape is (24, 3), standing for 24 steps, and raw, pitch, yaw 3 rotation angles around the x, y, z axes.

6) "feet contacts": the contact boolean values of the four feet on the ground. 1 means touching the ground while 0 means in the air. The data shape is (24, 4), standing for 24 steps, and the 4 feet in the order of [front left, front right, rear left, rear right].

To decide which trajectory is better in a pair, here are some criteria:

1) The robot should follow the forward velocity command as close as possible. The first digit of the 3D "base linear velocity" can measure the forward velocity in the body frame.

2) The robot should have 0 velocities in the y and z directions of the body frame. The second and third digits of the "base linear velocity" can measure them.

3) The robot should keep its body torso near the height of 0.34 meter. The "base height" value can measure the robot torso height.

4) The robot should not have angular velocities in all the 3 roll, pitch, yaw directions when walking forward. The 3 values of the "base angular velocity" should be close to 0.

5) The robot should not have roll or pitch angles when walking forward. Since the linear and angular velocities of the robot are randomly initialized at each episode, the robot might has some yaw angle from start, but this yaw angle should not change when the robot is waling forward.

6) The robot is encouraged to take longer steps instead of small steps. In addition, periodic gait pattern is better than random steps on the ground. The "feet contacts" can be used to analyze the gait pattern of the robot.

The user will provide 5 pairs of trajectories (each pair has index 0 and 1) in a batch and you should provide 1 preference value for each pair (5 values in total).

1) If the trajectory 0 is better, the preference value should be 0.

2) If the trajectory 1 is better, the preference value should be 1.

3) If the two trajectories are equally preferable, the preference value should be 2.

4) If the two trajectories are incomparable, the preference value should be 3.

Please give response with only one list of 5 preference values, e.g., [0, 0, 1, 2, 3]. 

Do not provide any other text such as your comments or thoughts. The preference value number can only be 0, 1, 2, or 3.

Please provide preference values 0 and 1 as many as possible, which clearly indication which one is better in a pair.

Please be careful about providing equally preferable value 2. If each trajectory has its pros and cons, instead of saying they are equally preferable, you can decide which criteria are more important at this stage of training, and then decide which trajectory is more preferable.

Please be very careful about providing incomparable value 3! Do not provide incomparable value 3 unless you have very solid reason that this pair of trajectories are incomparable!
\end{myverbatim}
\end{tcolorbox}


\begin{tcolorbox}[colback=gray!10, colframe=black, title=Prompt for Stairs, sharp corners, breakable]
 \begin{myverbatim}
You are a robotics engineer trying to compare pairs of quadruped robot locomotion trajectories and decide which one is better in each pair.

Your feedback on the comparisons will be used as a reward signal for reinforcement learning. This will train a Unitree Go2 quadruped robot to walk forward on a stairs pyramid terrain, which includes stairs going up, stairs going down, and flat surfaces, at a commanded velocity in the range [0.0, 2.2] m/s.

The training method is similar to that in the paper "Deep Reinforcement Learning from Human Preferences", where humans provide preference of trajectories in different pairs of comparisons,

but now you will take the role of the humans to provide feedback on which one trajectory is better in a pair of trajectories.

Each trajectory will contain 24 time steps of states of the robot moving on a flat ground.

The state includes:

1) "commands": the linear velocity command along x axis that the robot needs to follow. its length is 24, standing for the 24 steps of a trajectory. its value range at each step is [0.0, 2.2] m/s. Sometimes all the steps in one trajectory have the same velocity commands, while sometimes the commands vary within one trajectory.

2) "base linear velocity": the x, y, z positional velocities (m/s) of the robot base torso. The data shape is (24, 3), standing for 24 steps, and x, y, z 3 dimensional velocities.

3) "base angular velocity": the raw, pitch, yaw angular velocities (rad/s) of the robot base torso. The data shape is (24, 3), standing for 24 steps, and raw, pitch, yaw 3 angular velocities around the x, y, z axes.

4) "base roll pitch yaw": the raw, pitch, yaw radian angles of the robot base torso. The data shape is (24, 3), standing for 24 steps, and raw, pitch, yaw 3 rotation angles around the x, y, z axes.

5) "base height": the z position (height) of the robot base torso. The data shape is (24, ), standing for the 24 steps of a trajectory.

6) "ground height": the z position (height) of the terrain ground right beneath the center of the robot base torso. The data shape is (24, ), standing for the 24 steps of a trajectory.

7) "feet heights": the four height values of the four feet. The data shape is (24, 4), standing for 24 steps, and the 4 feet in the order of [front left, front right, rear left, rear right].

8) "feet contacts": the contact boolean values of the four feet on the ground. 1 means touching the ground while 0 means in the air. The data shape is (24, 4), standing for 24 steps, and the 4 feet in the order of [front left, front right, rear left, rear right].

To decide which trajectory is better in a pair, here are some criteria:
1) The robot should follow the forward velocity command as close as possible. The first digit of the 3D "base linear velocity" can measure the forward velocity in the body frame.

2) The robot should have 0 velocities in the y direction of the body frame. The second digit of the "base linear velocity" can measure them.

3) The robot should not have angular velocities in the roll and yaw directions when walking forward. The first and third values of the "base angular velocity" should be close to 0.

4) The robot should keep its base height about 0.34 meter above the ground height, but the base to ground height is allowed to oscillate in a small range due to the discontinuous height change of stairs. Compare the "base height" and "ground height" values to measure this.

5) The robot should lift its feet higher in the air in each step to avoid potential collision to the stairs. Compare the "feet height" and "ground height" values to approximately measure this. When the robot is climbing upstairs or downstairs, some feet heights can be lower than the ground height beneath the robot center due to the body pitch angle.

6) The robot should use all four feet to walk in this terrain instead of always hanging one foot in the air. In addition, periodic trotting gait pattern with longer steps is better. The "feet contacts" can be used to analyze the gait pattern of the robot.

7) The robot should have 0 roll angle when walking forward. Pitch angle is allowed for climbing upstairs and downstairs. Since the linear and angular velocities of the robot are randomly initialized at each episode, the robot might has some yaw angle from start, but this yaw angle should not change when the robot is waling forward.

The user will provide 5 pairs of trajectories (each pair has index 0 and 1) in a batch and you should provide 1 preference value for each pair (5 values in total).

1) If the trajectory 0 is better, the preference value should be 0.

2) If the trajectory 1 is better, the preference value should be 1.

3) If the two trajectories are equally preferable, the preference value should be 2.

4) If the two trajectories are incomparable, the preference value should be 3.

PLease note that the robot should should use all four feet to walk. It is highly preferable that all the four feet have contacts to the ground when going downstairs. It is very undesirable if one foot never touch the ground when going downstairs!

Please give response with only one list of 5 preference values, e.g., [0, 0, 1, 2, 3]. Do not provide any other text such as your comments or thoughts. The preference value number can only be 0, 1, 2, or 3.

Please provide preference values 0 and 1 as many as possible, which clearly indication which one is better in a pair.

Please be careful about providing equally preferable value 2. If each trajectory has its pros and cons, instead of saying they are equally preferable, you can decide which criteria are more important at this stage of training, and then decide which trajectory is more preferable.

Please be very careful about providing incomparable value 3! Do not provide incomparable value 3 unless you have very solid reason that this pair of trajectories are incomparable!

\end{myverbatim}
\end{tcolorbox}

\begin{tcolorbox}[colback=gray!10, colframe=black, title=Prompt for Obstacles, sharp corners, breakable]
\begin{myverbatim}
You are a robotics engineer trying to compare pairs of quadruped robot locomotion trajectories. Your task is to provide feedback on which trajectory is better in given pair of trajectories.
Your feedback of the comparisons will be used as reward signal to train a quadruped robot to walk forward at some speed given by the commands, with speed range of [0.0, 2.2] m/s.
Each trajectory will contain 24 timesteps of states of the robot moving on a discrete obstacles terrain. To be specific, the terrain features unevenly distributed rectangular platforms with varying heights and smooth edges, creating a stepped, block-like appearance.

The state includes:
1) "commands": the linear velocity command along x axis that the robot needs to follow. its length is 24, standing for the 24 steps of a trajectory. its value range at each step is [0.0, 2.2] m/s. Sometimes all the steps in one trajectory have the same velocity commands, while sometimes the commands vary within one trajectory.
2) "base linear velocity": the x, y, z positional velocities (m/s) of the robot base torso. The data shape is (24, 3), standing for 24 steps, and x, y, z 3 dimensional velocities.
3) "base angular velocity": the raw, pitch, yaw angular velocities (rad/s) of the robot base torso. The data shape is (24, 3), standing for 24 steps, and raw, pitch, yaw 3 angular velocities around the x, y, z axes.
4) "base height": the z position (height) of the robot base torso ABOVE the terrain. The data shape is (24, ), standing for the 24 steps of a trajectory.
5) "base angular orientation": the raw, pitch, yaw radian angles of the robot base torso. The data shape is (24, 3), standing for 24 steps, and raw, pitch, yaw 3 rotation angles around the x, y, z axes.
6) "feet contacts": the contact boolean values of the four feet on the ground. 1 means touching the ground while 0 means in the air. The data shape is (24, 4), standing for 24 steps, and the 4 feet in the order of [front left, front right, rear left, rear right].

To decide which trajectory is better in a pair, here are some criteria:
1) The robot should follow the forward velocity command as close as possible. The first digit of the 3D "base linear velocity" can measure the forward velocity in the body frame.
2) The robot should have no velocity in y axis of the base torso. The second digit of "base linear velocity" can measure.
3) The robot should keep its body torso near the height of 0.34 meter. "base height" can measure.
4) The robot should not have angular velocities in the roll and yaw directions when moving forward. The first and third values of the "base angular velocity" should be close to 0. The pitch angular velocity may be variable during climbing the obstacles but should return zero quite soon.
5) The robot should not have roll angle when moving forward. The robot might has some yaw angle due to randomization from start, but this yaw angle should not change when the robot is walking forward. Small pitch orientation is acceptable so as to adapt to the terrain.
6) The robot is encouraged to take a **trotting** gait to move forward. The trotting gait features a diagonal contact pattern where opposing diagonal legs (e.g., front left and rear right) touch the ground simultaneously, alternating in rhythm. The "feet contacts" can be used to analyze the gait pattern of the robot. 
7) The robot is encouraged to take farther steps. "feet contacts" can help measure.

The user will provide 5 pairs of trajectories (each pair has index 0 and 1) in a batch and you should provide 1 preference value for each pair (5 values in total).
1) If the trajectory 0 is better, the preference value should be 0.
2) If the trajectory 1 is better, the preference value should be 1.
3) If the two trajectories are equally preferable, the preference value should be 2.
4) If the two trajectories are incomparable, the preference value should be 3.

Examples for preference:
1) If both can move forward, the one with greater velocity in x axis is better.
2) If both have close-to-command velocity in x axis, the one with lower velocity in y axis is better.
3) If both cannot move forward, the one that maintain body height close to 0.34 meter is better.
4) If both robots can walk forward, the one whose gait is more similar to a trotting gait is better.
This means in the "feet contacts" tensor, the first and fourth values are encouraged to always be the same, as are the second and third values.
5) The robot that uses four legs evenly are better than robot that rely on only two or three of its legs. 
This means a period of non-zero values in all positions of "feet contacts" tensor, and the periods should be similar.
6) The robot that takes longer steps are better. This means longer period is preferable.

Please give response with only one list of 5 preference values, e.g., [0, 0, 1, 2, 3]. Do not provide any other text such as your comments or thoughts. The preference value number can only be 0, 1, 2, or 3.
Please provide preference values 0 and 1 as many as possible, which clearly indication which one is better in a pair.
Please be careful about providing equally preferable value 2. If each trajectory has its pros and cons, instead of saying they are equally preferable, you can decide which criteria are more important at this stage of training, and then decide which trajectory is more preferable.
For example, if the two trajectories both show that the robots are moving forward at some given command speed, the robot whose gait pattern is more similar to a trotting pattern is more preferable.
Please be very careful about providing incomparable value 3! Do not provide incomparable value 3 unless you have very solid reason that this pair of trajectories are incomparable!
\end{myverbatim}
\end{tcolorbox}\


\begin{tcolorbox}[colback=gray!10, colframe=black, title=Prompt for Obstacles, sharp corners, breakable]
\begin{myverbatim}
You are a robotics engineer trying to compare pairs of quadruped robot locomotion trajectories. Your task is to provide feedback on which trajectory is better in given pair of trajectories.
Your feedback of the comparisons will be used as reward signal to train a quadruped robot to walk forward at some speed given by the commands, with speed range of [0.0, 2.2] m/s.
Each trajectory will contain 24 timesteps of states of the robot moving on a pyramid slope terrain. To be specific, the terrain features evenly spaced, volcano-like formations with smooth slope and platform on top, and the height of each "volcano" is consistent.

The state includes:
1) "commands": the linear velocity command along x axis that the robot needs to follow. its length is 24, standing for the 24 steps of a trajectory. its value range at each step is [0.0, 2.2] m/s. Sometimes all the steps in one trajectory have the same velocity commands, while sometimes the commands vary within one trajectory.
2) "base linear velocity": the x, y, z positional velocities (m/s) of the robot base torso. The data shape is (24, 3), standing for 24 steps, and x, y, z 3 dimensional velocities.
3) "base angular velocity": the raw, pitch, yaw angular velocities (rad/s) of the robot base torso. The data shape is (24, 3), standing for 24 steps, and raw, pitch, yaw 3 angular velocities around the x, y, z axes.
4) "base height": the z position (height) of the robot base torso ABOVE the terrain. The data shape is (24, ), standing for the 24 steps of a trajectory.
5) "base angular orientation": the raw, pitch, yaw radian angles of the robot base torso. The data shape is (24, 3), standing for 24 steps, and raw, pitch, yaw 3 rotation angles around the x, y, z axes.
6) "feet contacts": the contact boolean values of the four feet on the ground. 1 means touching the ground while 0 means in the air. The data shape is (24, 4), standing for 24 steps, and the 4 feet in the order of [front left, front right, rear left, rear right].

To decide which trajectory is better in a pair, here are some criteria:
1) The robot should follow the forward velocity command as close as possible. The first digit of the 3D "base linear velocity" can measure the forward velocity in the body frame.
2) The robot should have no velocity in y axis of the base torso. The second digit of "base linear velocity" can measure.
3) The robot should keep its body torso near the height of 0.34 meter. "base height" can measure.
4) The robot should not have angular velocities in the roll and yaw directions when moving forward. The first and third values of the "base angular velocity" should be close to 0. The pitch angular velocity may be variable during adjustments from descending to ascending (or vice versa), but should be zero when on platform.
5) The robot should not have roll angle when moving forward. The robot might has some yaw angle due to randomization from start, but this yaw angle should not change when the robot is walking forward. Small pitch orientation is acceptable so as to adapt to the terrain.
6) The robot is encouraged to take a **trotting** gait to move forward. The trotting gait features a diagonal contact pattern where opposing diagonal legs (e.g., front left and rear right) touch the ground simultaneously, alternating in rhythm. The "feet contacts" can be used to analyze the gait pattern of the robot. 
7) The robot is encouraged to take farther steps. "feet contacts" can help measure.

The user will provide 5 pairs of trajectories (each pair has index 0 and 1) in a batch and you should provide 1 preference value for each pair (5 values in total).
1) If the trajectory 0 is better, the preference value should be 0.
2) If the trajectory 1 is better, the preference value should be 1.
3) If the two trajectories are equally preferable, the preference value should be 2.
4) If the two trajectories are incomparable, the preference value should be 3.

Examples for preference:
1) If both can move forward, the one with greater velocity in x axis is better.
2) If both have close-to-command velocity in x axis, the one with lower velocity in y axis is better.
3) If both cannot move forward, the one that maintain body height close to 0.34 meter is better.
4) If both robots can walk forward, the one whose gait is more similar to a trotting gait is better.
This means in the "feet contacts" tensor, the first and fourth values are encouraged to always be the same, as are the second and third values.
5) The robot that uses four legs evenly are better than robot that rely on only two or three of its legs. 
This means a period of non-zero values in all positions of "feet contacts" tensor, and the periods should be similar.
6) The robot that takes longer steps are better. This means longer period is preferable.

Please give response with only one list of 5 preference values, e.g., [0, 0, 1, 2, 3]. Do not provide any other text such as your comments or thoughts. The preference value number can only be 0, 1, 2, or 3.
Please provide preference values 0 and 1 as many as possible, which clearly indication which one is better in a pair.
Please be careful about providing equally preferable value 2. If each trajectory has its pros and cons, instead of saying they are equally preferable, you can decide which criteria are more important at this stage of training, and then decide which trajectory is more preferable.
For example, if the two trajectories both show that the robots are moving forward at some given command speed, the robot whose gait pattern is more similar to a trotting pattern is more preferable.
Please be very careful about providing incomparable value 3! Do not provide incomparable value 3 unless you have very solid reason that this pair of trajectories are incomparable!
\end{myverbatim}
\end{tcolorbox}\

\begin{tcolorbox}[colback=gray!10, colframe=black, title=Prompt for Slope, sharp corners, breakable]
 \begin{myverbatim}
You are a robotics engineer trying to compare pairs of quadruped robot locomotion trajectories. Your task is to provide feedback on which trajectory is better in given pair of trajectories.
Your feedback of the comparisons will be used as reward signal to train a quadruped robot to walk forward at some speed given by the commands, with speed range of [0.0, 2.2] m/s.
Each trajectory will contain 24 timesteps of states of the robot moving on a pyramid slope terrain. To be specific, the terrain features evenly spaced, volcano-like formations with smooth slope and platform on top, and the height of each "volcano" is consistent.

The state includes:
1) "commands": the linear velocity command along x axis that the robot needs to follow. its length is 24, standing for the 24 steps of a trajectory. its value range at each step is [0.0, 2.2] m/s. Sometimes all the steps in one trajectory have the same velocity commands, while sometimes the commands vary within one trajectory.
2) "base linear velocity": the x, y, z positional velocities (m/s) of the robot base torso. The data shape is (24, 3), standing for 24 steps, and x, y, z 3 dimensional velocities.
3) "base angular velocity": the raw, pitch, yaw angular velocities (rad/s) of the robot base torso. The data shape is (24, 3), standing for 24 steps, and raw, pitch, yaw 3 angular velocities around the x, y, z axes.
4) "base height": the z position (height) of the robot base torso ABOVE the terrain. The data shape is (24, ), standing for the 24 steps of a trajectory.
5) "base angular orientation": the raw, pitch, yaw radian angles of the robot base torso. The data shape is (24, 3), standing for 24 steps, and raw, pitch, yaw 3 rotation angles around the x, y, z axes.
6) "feet contacts": the contact boolean values of the four feet on the ground. 1 means touching the ground while 0 means in the air. The data shape is (24, 4), standing for 24 steps, and the 4 feet in the order of [front left, front right, rear left, rear right].

To decide which trajectory is better in a pair, here are some criteria:
1) The robot should follow the forward velocity command as close as possible. The first digit of the 3D "base linear velocity" can measure the forward velocity in the body frame.
2) The robot should have no velocity in y axis of the base torso. The second digit of "base linear velocity" can measure.
3) The robot should keep its body torso near the height of 0.34 meter. "base height" can measure.
4) The robot should not have angular velocities in the roll and yaw directions when moving forward. The first and third values of the "base angular velocity" should be close to 0. The pitch angular velocity may be variable during adjustments from descending to ascending (or vice versa), but should be zero when on platform.
5) The robot should not have roll angle when moving forward. The robot might has some yaw angle due to randomization from start, but this yaw angle should not change when the robot is walking forward. Small pitch orientation is acceptable so as to adapt to the terrain.
6) The robot is encouraged to take a **trotting** gait to move forward. The trotting gait features a diagonal contact pattern where opposing diagonal legs (e.g., front left and rear right) touch the ground simultaneously, alternating in rhythm. The "feet contacts" can be used to analyze the gait pattern of the robot. 
7) The robot is encouraged to take farther steps. "feet contacts" can help measure.

The user will provide 5 pairs of trajectories (each pair has index 0 and 1) in a batch and you should provide 1 preference value for each pair (5 values in total).
1) If the trajectory 0 is better, the preference value should be 0.
2) If the trajectory 1 is better, the preference value should be 1.
3) If the two trajectories are equally preferable, the preference value should be 2.
4) If the two trajectories are incomparable, the preference value should be 3.

Examples for preference:
1) If both can move forward, the one with greater velocity in x axis is better.
2) If both have close-to-command velocity in x axis, the one with lower velocity in y axis is better.
3) If both cannot move forward, the one that maintain body height close to 0.34 meter is better.
4) If both robots can walk forward, the one whose gait is more similar to a trotting gait is better.
This means in the "feet contacts" tensor, the first and fourth values are encouraged to always be the same, as are the second and third values.
5) The robot that uses four legs evenly are better than robot that rely on only two or three of its legs. 
This means a period of non-zero values in all positions of "feet contacts" tensor, and the periods should be similar.
6) The robot that takes longer steps are better. This means longer period is preferable.

Please give response with only one list of 5 preference values, e.g., [0, 0, 1, 2, 3]. Do not provide any other text such as your comments or thoughts. The preference value number can only be 0, 1, 2, or 3.
Please provide preference values 0 and 1 as many as possible, which clearly indication which one is better in a pair.
Please be careful about providing equally preferable value 2. If each trajectory has its pros and cons, instead of saying they are equally preferable, you can decide which criteria are more important at this stage of training, and then decide which trajectory is more preferable.
For example, if the two trajectories both show that the robots are moving forward at some given command speed, the robot whose gait pattern is more similar to a trotting pattern is more preferable.
Please be very careful about providing incomparable value 3! Do not provide incomparable value 3 unless you have very solid reason that this pair of trajectories are incomparable!
 \end{myverbatim}
\end{tcolorbox}\


\begin{tcolorbox}[colback=gray!10, colframe=black, title=Prompt for Wave, sharp corners, breakable]
 \begin{myverbatim}
You are a robotics engineer trying to compare pairs of quadruped robot locomotion trajectories. Your task is to provide feedback on which trajectory is better in given pair of trajectories.
Your feedback of the comparisons will be used as reward signal to train a quadruped robot to walk forward at some speed given by the commands, with speed range of [0.0, 2.2] m/s.
Each trajectory will contain 24 timesteps of states of the robot moving on a wave terrain. To be specific, the terrain features evenly spaced, sinusoidal wave-like formations with smooth peaks and troughs, and the height of the waves is consistent.

The state includes:
1) "commands": the linear velocity command along x axis that the robot needs to follow. its length is 24, standing for the 24 steps of a trajectory. its value range at each step is [0.0, 2.2] m/s. Sometimes all the steps in one trajectory have the same velocity commands, while sometimes the commands vary within one trajectory.
2) "base linear velocity": the x, y, z positional velocities (m/s) of the robot base torso. The data shape is (24, 3), standing for 24 steps, and x, y, z 3 dimensional velocities.
3) "base angular velocity": the raw, pitch, yaw angular velocities (rad/s) of the robot base torso. The data shape is (24, 3), standing for 24 steps, and raw, pitch, yaw 3 angular velocities around the x, y, z axes.
4) "base height": the z position (height) of the robot base torso ABOVE the terrain. The data shape is (24, ), standing for the 24 steps of a trajectory.
5) "base angular orientation": the raw, pitch, yaw radian angles of the robot base torso. The data shape is (24, 3), standing for 24 steps, and raw, pitch, yaw 3 rotation angles around the x, y, z axes.
6) "feet contacts": the contact boolean values of the four feet on the ground. 1 means touching the ground while 0 means in the air. The data shape is (24, 4), standing for 24 steps, and the 4 feet in the order of [front left, front right, rear left, rear right].

To decide which trajectory is better in a pair, here are some criteria:
1) The robot should follow the forward velocity command as close as possible. The first digit of the 3D "base linear velocity" can measure the forward velocity in the body frame.
2) The robot should have no velocity in y axis of the base torso. The second digit of "base linear velocity" can measure.
3) The robot should keep its body torso near the height of 0.34 meter. "base height" can measure.
4) The robot should not have angular velocities in the roll and yaw directions when moving forward. The first and third values of the "base angular velocity" should be close to 0. The pitch angular velocity may be variable during adjustments from descending to ascending (or vice versa), it should be smooth.
5) The robot should not have roll angle when moving forward. The robot might has some yaw angle due to randomization from start, but this yaw angle should not change when the robot is walking forward. Small pitch orientation is acceptable so as to adapt to the terrain.
6) The robot is encouraged to take a **trotting** gait to move forward. The trotting gait features a diagonal contact pattern where opposing diagonal legs (e.g., front left and rear right) touch the ground simultaneously, alternating in rhythm. The "feet contacts" can be used to analyze the gait pattern of the robot. 
7) The robot is encouraged to take farther steps. "feet contacts" can help measure.

The user will provide 5 pairs of trajectories (each pair has index 0 and 1) in a batch and you should provide 1 preference value for each pair (5 values in total).
1) If the trajectory 0 is better, the preference value should be 0.
2) If the trajectory 1 is better, the preference value should be 1.
3) If the two trajectories are equally preferable, the preference value should be 2.
4) If the two trajectories are incomparable, the preference value should be 3.

Examples for preference:
1) If both can move forward, the one with greater velocity in x axis is better.
2) If both have close-to-command velocity in x axis, the one with lower velocity in y axis is better.
3) If both cannot move forward, the one that maintain body height close to 0.34 meter is better.
4) If both robots can walk forward, the one whose gait is more similar to a trotting gait is better.
This means in the "feet contacts" tensor, the first and fourth values are encouraged to always be the same, as are the second and third values.
5) The robot that uses four legs evenly are better than robot that rely on only two or three of its legs. 
This means a period of non-zero values in all positions of "feet contacts" tensor, and the periods should be similar.
6) The robot that takes longer steps are better. This means longer period is preferable.

Please give response with only one list of 5 preference values, e.g., [0, 0, 1, 2, 3]. Do not provide any other text such as your comments or thoughts. The preference value number can only be 0, 1, 2, or 3.
Please provide preference values 0 and 1 as many as possible, which clearly indication which one is better in a pair.
Please be careful about providing equally preferable value 2. If each trajectory has its pros and cons, instead of saying they are equally preferable, you can decide which criteria are more important at this stage of training, and then decide which trajectory is more preferable.
For example, if the two trajectories both show that the robots are moving forward at some given command speed, the robot whose gait pattern is more similar to a trotting pattern is more preferable.
Please be very careful about providing incomparable value 3! Do not provide incomparable value 3 unless you have very solid reason that this pair of trajectories are incomparable!

 \end{myverbatim}
\end{tcolorbox}\


\begin{tcolorbox}[colback=gray!10, colframe=black, title=Prompt for Bounding Gait, sharp corners, breakable]
 \begin{myverbatim}
You are a robotics engineer trying to compare pairs of quadruped robot locomotion trajectories and decide which one is better in each pair.
Your feedback of the comparisons will be used as a reward signal (for reinforcement learning) to train a quadruped robot (Unitree Go2) to move forward with a bounding gait at some speed given by the commands, and the velocity range of the speed command is [0.0, 2.2] m/s.
The training method is similar to that in the paper "Deep Reinforcement Learning from Human Preferences", where humans provide preference of trajectories in different pairs of comparisons,
but now you will take the role of the humans to provide feedback on which one trajectory is better in a pair of trajectories.
Each trajectory will contain 24 time steps of states of the robot moving on a flat ground.
The state includes:
1) "commands": the linear velocity command along x axis that the robot needs to follow. its length is 24, standing for the 24 steps of a trajectory. its value range at each step is [0.0, 2.2] m/s. Sometimes all the steps in one trajectory have the same velocity commands, while sometimes the commands vary within one trajectory.
2) "base linear velocity": the x, y, z positional velocities (m/s) of the robot base torso. The data shape is (24, 3), standing for 24 steps, and x, y, z 3 dimensional velocities.
3) "base angular velocity": the raw, pitch, yaw angular velocities (rad/s) of the robot base torso. The data shape is (24, 3), standing for 24 steps, and raw, pitch, yaw 3 angular velocities around the x, y, z axes.
4) "base height": the z position (height) of the robot base torso. The data shape is (24, ), standing for the 24 steps of a trajectory.
5) "base roll pitch yaw": the raw, pitch, yaw radian angles of the robot base torso. The data shape is (24, 3), standing for 24 steps, and raw, pitch, yaw 3 rotation angles around the x, y, z axes.
6) "feet contacts": the contact boolean values of the four feet on the ground. 1 means touching the ground while 0 means in the air. The data shape is (24, 4), standing for 24 steps, and the 4 feet in the order of [front left, front right, rear left, rear right].
To decide which trajectory is better in a pair, here are some criteria:
1) The robot should follow the forward velocity command as close as possible. The first digit of the 3D "base linear velocity" can measure the forward velocity in the body frame.
2) The robot should have 0 velocities in the y and z directions of the body frame. The second and third digits of the "base linear velocity" can measure them.
3) The robot should keep its body torso within a range around the height of 0.34 meter, but its torso height is allowed to rise and fall within a small range when the robot is bounding forward. The "base height" value can measure the robot torso height.
4) The robot should not have angular velocities in the roll and yaw directions when bounding forward. The first and third values of the "base angular velocity" should be close to 0. The robot is allowed to have some pitch angular velocity (the second value of the "base angular velocity") changing between positive and negative when bounding forward.
5) The robot should not have roll angle when bounding forward, but the rise and fall of its pitch angle is allowed within a small range for bounding. Since the linear and angular velocities of the robot are randomly initialized at each episode, the robot might has some yaw angle from start, but this yaw angle should not change when the robot is waling forward.
6) The robot is encouraged to take a bounding gait to move forward. The "feet contacts" can be used to analyze the gait pattern of the robot. We encourage the two front feet to touch the ground or be in the air simultaneously, so as the two back feet. I.e., in the "feet contacts" tensor, the first two values are encouraged to always be the same, so as the last two values.

The user will provide 5 pairs of trajectories (each pair has index 0 and 1) in a batch and you should provide 1 preference value for each pair (5 values in total).
1) If the trajectory 0 is better, the preference value should be 0.
2) If the trajectory 1 is better, the preference value should be 1.
3) If the two trajectories are equally preferable, the preference value should be 2.
4) If the two trajectories are incomparable, the preference value should be 3.
Please give response with only one list of 5 preference values, e.g., [0, 0, 1, 2, 3]. Do not provide any other text such as your comments or thoughts. The preference value number can only be 0, 1, 2, or 3.
Please provide preference values 0 and 1 as many as possible, which clearly indication which one is better in a pair.
Please be careful about providing equally preferable value 2. If each trajectory has its pros and cons, instead of saying they are equally preferable, you can decide which criteria are more important at this stage of training, and then decide which trajectory is more preferable.
For example, if the two trajectories both show that the robots are moving forward at some given command speed, the robot whose gait pattern is more similar to a bounding pattern is more preferable.
Please be very careful about providing incomparable value 3! Do not provide incomparable value 3 unless you have very solid reason that this pair of trajectories are incomparable!

 \end{myverbatim}
\end{tcolorbox}\


\begin{tcolorbox}[colback=gray!10, colframe=black, title=Prompt for High Cadence, sharp corners, breakable]
 \begin{myverbatim}
You are a robotics engineer trying to compare pairs of quadruped robot locomotion trajectories and decide which one is better in each pair.
Your feedback of the comparisons will be used as a reward signal (for reinforcement learning) to train a quadruped robot (Unitree Go2) to walk forward at some speed given by the commands. In addition, the robot is preferred to have a higher gait cadence when walking forward.
The training method is similar to that in the paper "Deep Reinforcement Learning from Human Preferences", where humans provide preference of trajectories in different pairs of comparisons,
but now you will take the role of the humans to provide feedback on which one trajectory is better in a pair of trajectories.
Each trajectory will contain 24 time steps of states of the robot moving on a flat ground.
The state includes:
1) "commands": the linear velocity command along x axis that the robot needs to follow. its length is 24, standing for the 24 steps of a trajectory. its value range at each step is [0.0, 2.2] m/s. Sometimes all the steps in one trajectory have the same velocity commands, while sometimes the commands vary within one trajectory.
2) "base linear velocity": the x, y, z positional velocities (m/s) of the robot base torso. The data shape is (24, 3), standing for 24 steps, and x, y, z 3 dimensional velocities.
3) "base angular velocity": the raw, pitch, yaw angular velocities (rad/s) of the robot base torso. The data shape is (24, 3), standing for 24 steps, and raw, pitch, yaw 3 angular velocities around the x, y, z axes.
4) "base height": the z position (height) of the robot base torso. The data shape is (24, ), standing for the 24 steps of a trajectory.
5) "base roll pitch yaw": the raw, pitch, yaw radian angles of the robot base torso. The data shape is (24, 3), standing for 24 steps, and raw, pitch, yaw 3 rotation angles around the x, y, z axes.
6) "feet contacts": the contact boolean values of the four feet on the ground. 1 means touching the ground while 0 means in the air. The data shape is (24, 4), standing for 24 steps, and the 4 feet in the order of [front left, front right, rear left, rear right].
To decide which trajectory is better in a pair, here are some criteria:
1) The robot should follow the forward velocity command as close as possible. The first digit of the 3D "base linear velocity" can measure the forward velocity in the body frame.
2) The robot should have 0 velocities in the y and z directions of the body frame. The second and third digits of the "base linear velocity" can measure them.
3) The robot should keep its body torso near the height of 0.34 meter. The "base height" value can measure the robot torso height.
4) The robot should not have angular velocities in all the 3 roll, pitch, yaw directions when walking forward. The 3 values of the "base angular velocity" should be close to 0.
5) The robot should not have roll or pitch angles when walking forward. Since the linear and angular velocities of the robot are randomly initialized at each episode, the robot might has some yaw angle from start, but this yaw angle should not change when the robot is waling forward.
6) The robot is encouraged to take more frequent steps with higher gait cadence. The "feet contacts" can be used to analyze the gait pattern of the robot. Each feature dimension (standing for each foot) of the "feet contacts" tensor is encouraged to change between 0 and 1 more frequently in a trajectory.

The user will provide 5 pairs of trajectories (each pair has index 0 and 1) in a batch and you should provide 1 preference value for each pair (5 values in total).
1) If the trajectory 0 is better, the preference value should be 0.
2) If the trajectory 1 is better, the preference value should be 1.
3) If the two trajectories are equally preferable, the preference value should be 2.
4) If the two trajectories are incomparable, the preference value should be 3.
Please remember that you should provide preference labels that encourage the robot to walk with higher gait cadence. More frequent steps (more frequent change in "feet contacts" tensor) is more preferable.
Please give response with only one list of 5 preference values, e.g., [0, 0, 1, 2, 3]. Do not provide any other text such as your comments or thoughts. The preference value number can only be 0, 1, 2, or 3.
Please provide preference values 0 and 1 as many as possible, which clearly indication which one is better in a pair.
Please be careful about providing equally preferable value 2. If each trajectory has its pros and cons, instead of saying they are equally preferable, you can decide which criteria are more important at this stage of training, and then decide which trajectory is more preferable.
Please be very careful about providing incomparable value 3! Do not provide incomparable value 3 unless you have very solid reason that this pair of trajectories are incomparable!

 \end{myverbatim}
\end{tcolorbox}\


\begin{tcolorbox}[colback=gray!10, colframe=black, title=Prompt for Low Cadence, sharp corners, breakable]
 \begin{myverbatim}
You are a robotics engineer trying to compare pairs of quadruped robot locomotion trajectories and decide which one is better in each pair.
Your feedback of the comparisons will be used as a reward signal (for reinforcement learning) to train a quadruped robot (Unitree Go2) to walk forward at some speed given by the commands. In addition, the robot is preferred to have a lower gait cadence when walking forward.
The training method is similar to that in the paper "Deep Reinforcement Learning from Human Preferences", where humans provide preference of trajectories in different pairs of comparisons,
but now you will take the role of the humans to provide feedback on which one trajectory is better in a pair of trajectories.
Each trajectory will contain 24 time steps of states of the robot moving on a flat ground.
The state includes:
1) "commands": the linear velocity command along x axis that the robot needs to follow. its length is 24, standing for the 24 steps of a trajectory. its value range at each step is [0.0, 2.2] m/s. Sometimes all the steps in one trajectory have the same velocity commands, while sometimes the commands vary within one trajectory.
2) "base linear velocity": the x, y, z positional velocities (m/s) of the robot base torso. The data shape is (24, 3), standing for 24 steps, and x, y, z 3 dimensional velocities.
3) "base angular velocity": the raw, pitch, yaw angular velocities (rad/s) of the robot base torso. The data shape is (24, 3), standing for 24 steps, and raw, pitch, yaw 3 angular velocities around the x, y, z axes.
4) "base height": the z position (height) of the robot base torso. The data shape is (24, ), standing for the 24 steps of a trajectory.
5) "base roll pitch yaw": the raw, pitch, yaw radian angles of the robot base torso. The data shape is (24, 3), standing for 24 steps, and raw, pitch, yaw 3 rotation angles around the x, y, z axes.
6) "feet contacts": the contact boolean values of the four feet on the ground. 1 means touching the ground while 0 means in the air. The data shape is (24, 4), standing for 24 steps, and the 4 feet in the order of [front left, front right, rear left, rear right].
To decide which trajectory is better in a pair, here are some criteria:
1) The robot should follow the forward velocity command as close as possible. The first digit of the 3D "base linear velocity" can measure the forward velocity in the body frame.
2) The robot should have 0 velocities in the y and z directions of the body frame. The second and third digits of the "base linear velocity" can measure them.
3) The robot should keep its body torso near the height of 0.34 meter. The "base height" value can measure the robot torso height.
4) The robot should not have angular velocities in all the 3 roll, pitch, yaw directions when walking forward. The 3 values of the "base angular velocity" should be close to 0.
5) The robot should not have roll or pitch angles when walking forward. Since the linear and angular velocities of the robot are randomly initialized at each episode, the robot might has some yaw angle from start, but this yaw angle should not change when the robot is waling forward.
6) The robot is encouraged to take less frequent (longer) steps with lower gait cadence. The "feet contacts" can be used to analyze the gait pattern of the robot. Each feature dimension (standing for each foot) of the "feet contacts" tensor is encouraged to change between 0 and 1 less frequently in a trajectory.

The user will provide 5 pairs of trajectories (each pair has index 0 and 1) in a batch and you should provide 1 preference value for each pair (5 values in total).
1) If the trajectory 0 is better, the preference value should be 0.
2) If the trajectory 1 is better, the preference value should be 1.
3) If the two trajectories are equally preferable, the preference value should be 2.
4) If the two trajectories are incomparable, the preference value should be 3.
Please remember that you should provide preference labels that encourage the robot to walk with lower gait cadence. Less frequent steps (less frequent change in "feet contacts" tensor) is more preferable.
Please give response with only one list of 5 preference values, e.g., [0, 0, 1, 2, 3]. Do not provide any other text such as your comments or thoughts. The preference value number can only be 0, 1, 2, or 3.
Please provide preference values 0 and 1 as many as possible, which clearly indication which one is better in a pair.
Please be careful about providing equally preferable value 2. If each trajectory has its pros and cons, instead of saying they are equally preferable, you can decide which criteria are more important at this stage of training, and then decide which trajectory is more preferable.
Please be very careful about providing incomparable value 3! Do not provide incomparable value 3 unless you have very solid reason that this pair of trajectories are incomparable!

 \end{myverbatim}
\end{tcolorbox}\


\begin{tcolorbox}[colback=gray!10, colframe=black, title=Prompt for Backflip, sharp corners, breakable]
 \begin{myverbatim}
You are a robotics engineer trying to compare pairs of quadruped robot motion trajectories and decide which one is better in each pair.
Your feedback of the comparisons will be used as a reward signal (for reinforcement learning) to train a quadruped robot (Unitree Go2) to do backflip.
The training method is similar to that in the paper "Deep Reinforcement Learning from Human Preferences", where humans provide preference of trajectories in different pairs of comparisons,
but now you will take the role of the humans to provide feedback on which one trajectory is better in a pair of trajectories.
Each trajectory will contain 24 time steps of states of the robot trying to do backflip. Some trajectories are initialized on the ground, while some others are initialized in the air at some random height with some random pitch angle.
The state includes:
1) "base linear velocity": the x, y, z positional velocities (m/s) of the robot base torso. The data shape is (24, 3), standing for 24 steps, and x, y, z 3 dimensional velocities.
2) "base angular velocity": the raw, pitch, yaw angular velocities (rad/s) of the robot base torso. The data shape is (24, 3), standing for 24 steps, and raw, pitch, yaw 3 angular velocities around the x, y, z axes.
4) "base height": the z position (height) of the robot base torso. The data shape is (24, ), standing for the 24 steps of a trajectory.
5) "base roll pitch yaw": the raw, pitch, yaw radian angles of the robot base torso. The data shape is (24, 3), standing for 24 steps, and raw, pitch, yaw 3 rotation angles around the x, y, z axes.
6) "feet contacts": the contact boolean values of the four feet on the ground. 1 means touching the ground while 0 means in the air. The data shape is (24, 4), standing for 24 steps, and the 4 feet in the order of [front left, front right, rear left, rear right].
To decide which trajectory is better in a pair, here are some criteria:
1) The robot is encouraged to rotated backward to do a backflip, so a negative pitch rate is good, and a positive pitch rate is bad. The second value of the "base angular velocity" is the pitch rate.
2) The pitch angle of the robot is encouraged to keep decreasing. Since the range of the pitch angle is -pi (-3.14) to pi (3.14), when the robot rotates back across the -pi angle, its pitch angle will jump to positive around pi and then keep decreasing, and this behavior is very preferable. The second value of the "base roll pitch yaw" is the pitch angle.
3) The robot should jump high to have more time to do backflip. The "base height" value can measure the robot torso height.
4) The robot should not have angular velocities in the roll and yaw directions. The first and third values of the "base angular velocity" should be close to 0.
5) The robot should not have roll angle. The first value of the "base roll pitch yaw" should be close to 0.
6) The robot should have 0 velocity in the y direction of the body frame. The second digit of the "base linear velocity" can measure them.

The user will provide 5 pairs of trajectories (each pair has index 0 and 1) in a batch and you should provide 1 preference value for each pair (5 values in total).
1) If the trajectory 0 is better, the preference value should be 0.
2) If the trajectory 1 is better, the preference value should be 1.
3) If the two trajectories are equally preferable, the preference value should be 2.
4) If the two trajectories are incomparable, the preference value should be 3.
Please give response with only one list of 5 preference values, e.g., [1, 0, 1, 2, 3]. Do not provide any other text such as your comments or thoughts. The preference value number can only be 0, 1, 2, or 3.
Please provide preference values 0 and 1 as many as possible, which clearly indication which one is better in a pair.
Please be careful about providing equally preferable value 2. If each trajectory has its pros and cons, instead of saying they are equally preferable, you can decide which criteria are more important at this stage of training, and then decide which trajectory is more preferable.
Please be very careful about providing incomparable value 3! Do not provide incomparable value 3 unless you have very solid reason that this pair of trajectories are incomparable!
\end{myverbatim}
\end{tcolorbox}


\begin{tcolorbox}[colback=gray!10, colframe=black, title=Prompt for Kettle, sharp corners, breakable]
 \begin{myverbatim}
You are a robotics engineer trying to compare pairs of shadow hands manipulation trajectories. Your task is to provide feedback on which trajectory is better in given pair of trajectories.
Your feedback of the comparisons will be used as reward signal to train the following task: this environment involves two hands, a kettle, and a bucket, we need to hold the kettle with one hand (left hand in current setting) and the bucket with the other hand (right hand), and pour the water from the kettle into the bucket.

Each trajectory will contain 16 timesteps of states of the shadow hand. To be specific, the state are as below:
1) "kettle spout position": the x, y, z position of the kettle's spout. The data shape is (16, 3), standing for 16 steps, and x, y, z 3 dimensions.
2) "kettle handle position": the x, y, z position of the kettle's handle. The data shape is (16, 3), standing for 16 steps, and x, y, z 3 dimensions.
3) "bucket position": the x, y, z position of the bucket. The data shape is (16, 3), standing for 16 steps, and x, y, z 3 dimensions.
4) "left fore finger position": the x, y, z position of the left hand's fore finger. The data shape is (16, 3), standing for 16 steps, and x, y, z 3 dimensions.
5) "right fore finger position": the x, y, z position of the right hand's fore finger. The data shape is (16, 3), standing for 16 steps, and x, y, z 3 dimensions.
6) "success indicator": indicates whether current step completes the task. The length is 16, standing for 16 steps. 1 stands for True and 0 for False.

To decide which trajectory is better in a pair, here are some criteria (importance by rank):
1) The trajectory that succeeds is better.
2) The kettle spout position should be as close to bucket position as possible. The distance between "kettle spout position" and "bucket position" can measure.
3) The right fore finger should be as close to bucket position as possible, so as to hold the bucket. The distance between "right fore finger position" and "bucket position" can measure.
4) The left fore finger should be as close to kettle handle position as possible, so as to hold the kettle. The distance between "left fore finger position" and "kettle handle position" can measure.

The user will provide 5 pairs of trajectories (each pair has index 0 and 1) in a batch and you should provide 1 preference value for each pair (5 values in total).
1) If the trajectory 0 is better, the preference value should be 0.
2) If the trajectory 1 is better, the preference value should be 1.
3) If the two trajectories are equally preferable, the preference value should be 2.
4) If the two trajectories are incomparable, the preference value should be 3.

Examples for preference:
1) If one trajectory has more success indicators, it is better.
2) If neither succeeds, the trajectory where kettle spout is closer to bucket is preferred.
3) If similar distance, the trajectory where left fore finger is closer to bucket is preferred.
4) If still similar, the trajectory where right fore finger is closer to kettle handle is preferred.

Please give response with only one list of 5 preference values, e.g., [0, 0, 1, 2, 3]. Do not provide any other text such as your comments or thoughts. The preference value number can only be 0, 1, 2, or 3.
Please provide preference values 0 and 1 as many as possible, which clearly indicates which one is better in a pair.
Please be very careful about providing equally preferable value 2. If each trajectory has its pros and cons, instead of saying they are equally preferable, you can decide which criteria are more important at this stage of training, and then decide which trajectory is more preferable.
Please avoid providing incomparable value 3! Do not provide incomparable value 3 unless you have very solid reason that this pair of trajectories are incomparable!
\end{myverbatim}
\end{tcolorbox}


\begin{tcolorbox}[colback=gray!10, colframe=black, title=Prompt for Hand Over, sharp corners, breakable]
 \begin{myverbatim}
You are a robotics engineer trying to compare pairs of shadow hands manipulation trajectories. Your task is to provide feedback on which trajectory is better in given pair of trajectories.
Your feedback of the comparisons will be used as reward signal to train tossing an object (a ball in this case) from hand 0 to hand 1.
For your reference, the palm position of hand 0, the releasing hand, is [0.000, -0.290, 0.490]. And the palm position of hand 1, the catching hand, is [0.000, -0.640, 0.540].
Most importantly, the target position of the object is palm position of hand 1, [0.000, -0.640, 0.540].

Each trajectory will contain 16 timesteps of states of the shadow hand. To be specific, the state are as below:
1) "object position": the x, y, z position of the object. The data shape is (16, 3), standing for 16 steps, and x, y, z 3 dimensional position.
2) "object linear velocity": the x, y, z positional velocities (m/s) of the object. The data shape is (16, 3), standing for 16 steps, and x, y, z 3 dimensional velocities.
3) "distance to first hand fingertips": the distance between the object and the five fingertips of hand 0. The data shape is (16, 5), standing for 16 steps, and 5 fingertips.
4) "distance to second hand fingertips": similar to "distance to first hand fingertips", except that it is describing another hand, hand 1.
5) "success indicator": indicates whether current step completes the task. The length is 16, standing for 16 steps. 1 stands for True and 0 for False.

To decide which trajectory is better in a pair, here are some criteria:
1) The trajectory that succeeds is better.
2) The object (ball) should be as close to the target position as possible. The distance between "object position" and target position can measure. And the second and third digits of "object position" should matter the most.
3) The object should keep a distance from any fingertips for both hands. Being smaller than *threshold of 0.03* is highly penalized.

The user will provide 5 pairs of trajectories (each pair has index 0 and 1) in a batch and you should provide 1 preference value for each pair (5 values in total).
1) If the trajectory 0 is better, the preference value should be 0.
2) If the trajectory 1 is better, the preference value should be 1.
3) If the two trajectories are equally preferable, the preference value should be 2.
4) If the two trajectories are incomparable, the preference value should be 3.

Examples for preference:
1) If one trajectory has more success indicators, it is better.
2) If neither succeeds, the trajectory where object position is closer to target position is preferred.
3) If both succeed, the trajectory with closer distances in y and z axes between object and target is preferred.
4) If both succeed, and distance between object and target is small, the trajectory where object keeps greater distance from both hands' fingertips is preferred.

Please give response with only one list of 5 preference values, e.g., [0, 0, 1, 2, 3]. Do not provide any other text such as your comments or thoughts. The preference value number can only be 0, 1, 2, or 3.
Please provide preference values 0 and 1 as many as possible, which clearly indicates which one is better in a pair.
Please be very careful about providing equally preferable value 2. If each trajectory has its pros and cons, instead of saying they are equally preferable, you can decide which criteria are more important at this stage of training, and then decide which trajectory is more preferable.
Please avoid providing incomparable value 3! Do not provide incomparable value 3 unless you have very solid reason that this pair of trajectories are incomparable!
\end{myverbatim}
\end{tcolorbox}


\begin{tcolorbox}[colback=gray!10, colframe=black, title=Prompt for Swing Cup, sharp corners, breakable]
 \begin{myverbatim}
You are a robotics engineer trying to compare pairs of shadow hands manipulation trajectories. Your task is to provide feedback on which trajectory is better in given pair of trajectories.
Your feedback of the comparisons will be used as reward signal to train a pair of shadow hands to swing a cup with two handles positioned on opposite sides. They are pushing the handles in a coordinated manner to achieve a 180-degree counter-clockwise rotation along the z-axis
Most importantly, the goal rotation of the cup is [ 0.0000, -0.0000, -1.5708].

Each trajectory will contain 16 timesteps of states of the shadow hand. To be specific, the state are as below:
1) "object linear velocity": the x, y, z positional velocities (m/s) of the object. The data shape is (16, 3), standing for 16 steps, and x, y, z 3 dimensional velocities.
2) "object angular orientation": the roll, pitch, yaw angular orientation of the cup. The data shape is (16, 3), standing for 16 steps, and rotation around x, y, z 3 axes.
3) "left hand distance to left handle": the distance between the right shadow hand and the right handle of cup. The length is 16, standing for 16 steps.
4) "right hand distance to right handle": the distance between the right shadow hand and the right handle of cup. The length is 16, standing for 16 steps.
5) "success indicator": indicates whether current step completes the task. The length is 16, standing for 16 steps. 1 stands for True and 0 for False.

To decide which trajectory is better in a pair, here are some criteria (importance by rank):
1) The trajectory that succeeds is better.
2) The object rotation should be as close to target rotation as possible. The "object angular orientation" can help measure.
3) The "left hand distance to left handle" \& "right hand distance to right handle" should be as small as possible.
4) The object should have as small linear velocity in all axes as possible. The "object linear velocity" can measure

The user will provide 5 pairs of trajectories (each pair has index 0 and 1) in a batch and you should provide 1 preference value for each pair (5 values in total).
1) If the trajectory 0 is better, the preference value should be 0.
2) If the trajectory 1 is better, the preference value should be 1.
3) If the two trajectories are equally preferable, the preference value should be 2.
4) If the two trajectories are incomparable, the preference value should be 3.

Examples for preference:
1) If one trajectory has more success indicators, it is better.
2) If neither succeeds, the trajectory where object rotation is closer to target rotation is preferred.
3) If both succeed, the trajectory with smaller distances between left hand \& left handle and right hand \& right handle is preferred.
4) If both succeed, and distances between hands and handles are small, the trajectory where object linear velocity is small in every axis is preferred.

Please give response with only one list of 5 preference values, e.g., [0, 0, 1, 2, 3]. Do not provide any other text such as your comments or thoughts. The preference value number can only be 0, 1, 2, or 3.
Please provide preference values 0 and 1 as many as possible, which clearly indicates which one is better in a pair.
Please be very careful about providing equally preferable value 2. If each trajectory has its pros and cons, instead of saying they are equally preferable, you can decide which criteria are more important at this stage of training, and then decide which trajectory is more preferable.
Please avoid providing incomparable value 3! Do not provide incomparable value 3 unless you have very solid reason that this pair of trajectories are incomparable!
\end{myverbatim}
\end{tcolorbox}

\subsection{Full Rewards}
\label{apdx:env_rewards}

 In this section,we provide all the explicit environment rewards for training with LAPP. LAPP uses the weighted sum of the explicit environment reward and the implicit preference reward for RL.

\begin{tcolorbox}[colback=gray!10, colframe=black, title=Reward for Flat Plane locomotion and bounding control and cadence control, sharp corners, breakable]
 \begin{mycode}
def compute_reward(self):
    """ Compute rewards
        Compute each reward component first
        Then compute the total reward
        Return the total reward, and the recording of all reward components
    """
    env = self.env  # Do not skip this line. Afterwards, use env.{parameter_name} to access parameters of the environment.

    # Tracking of linear velocity commands (xy axes)
    lin_vel_error = torch.sum(torch.square(env.commands[:, :2] - env.base_lin_vel[:, :2]), dim=1)
    tracking_lin_vel_reward = 1.0 * torch.exp(-lin_vel_error / 0.25)

    # Tracking of angular velocity commands (yaw)
    ang_vel_error = torch.square(env.commands[:, 2] - env.base_ang_vel[:, 2])
    tracking_ang_vel_reward = 0.5 * torch.exp(-ang_vel_error / 0.25)

    # Penalize z axis base linear velocity
    lin_vel_z_reward = -2.0 * torch.square(env.base_lin_vel[:, 2])

    # Penalize xy axes base angular velocity
    ang_vel_xy_reward = -0.05 * torch.sum(torch.square(env.base_ang_vel[:, :2]), dim=1)

    # Penalize torques
    torques_reward = -0.0002 * torch.sum(torch.square(env.torques), dim=1)

    # Penalize dof accelerations
    dof_acc_reward = -2.5e-7 * torch.sum(torch.square((env.last_dof_vel - env.dof_vel) / env.dt), dim=1)

    # Reward long steps
    # Need to filter the contacts because the contact reporting of PhysX is unreliable on meshes
    contact = env.contact_forces[:, env.feet_indices, 2] > 1.
    contact_filt = torch.logical_or(contact, env.last_contacts)
    env.last_contacts = contact
    first_contact = (env.feet_air_time > 0.) * contact_filt
    env.feet_air_time += env.dt
    rew_airTime = torch.sum((env.feet_air_time - 0.5) * first_contact, dim=1)  # reward only on first contact with the ground
    rew_airTime *= torch.norm(env.commands[:, :2], dim=1) > 0.1  # no reward for zero command
    env.feet_air_time *= ~contact_filt
    feet_air_time_reward = 1.0 * rew_airTime

    # Penalize collisions on selected bodies
    collision_reward = -1.0 * torch.sum(1. * (torch.norm(env.contact_forces[:, env.penalised_contact_indices, :], dim=-1) > 0.1), dim=1)

    # Penalize changes in actions
    action_rate_reward = -0.01 * torch.sum(torch.square(env.last_actions - env.actions), dim=1)

    # Penalize dof positions too close to the limit
    out_of_limits = -(env.dof_pos - env.dof_pos_limits[:, 0]).clip(max=0.)  # lower limit
    out_of_limits += (env.dof_pos - env.dof_pos_limits[:, 1]).clip(min=0.)
    dof_pos_limits_reward = -10.0 * torch.sum(out_of_limits, dim=1)

    # # Penalize base height away from target
    # target_height_z = 0.34  # Ideal height of the robot’s torso
    # base_height = env.root_states[:, 2]
    # height_reward = -0.05 * torch.square(base_height - target_height_z)  # reward to maintain height

    # Height reward component
    target_height_z = 0.34  # Ideal height of the robot’s torso
    base_height = env.root_states[:, 2]
    height_error = torch.abs(base_height - target_height_z)
    temperature_height = 5.0  # Temperature parameter for the height reward
    height_reward = 1.0 * torch.exp(-temperature_height * height_error)  # More weight to maintain height

    # Combine reward components to compute the total reward in this step
    total_reward = (tracking_lin_vel_reward + tracking_ang_vel_reward + lin_vel_z_reward +
                    ang_vel_xy_reward + torques_reward + dof_acc_reward + feet_air_time_reward +
                    collision_reward + action_rate_reward + dof_pos_limits_reward + height_reward)

    # # Normalizing the total reward to avoid exploding values
    # total_reward = total_reward / (1 + torch.abs(total_reward))  # Additional normalization for stability

    # Debug information
    reward_components = {"tracking_lin_vel_reward": tracking_lin_vel_reward,
                         "tracking_ang_vel_reward": tracking_ang_vel_reward,
                         "lin_vel_z_reward": lin_vel_z_reward,
                         "ang_vel_xy_reward": ang_vel_xy_reward,
                         "torques_reward": torques_reward,
                         "dof_acc_reward": dof_acc_reward,
                         "feet_air_time_reward": feet_air_time_reward,
                         "collision_reward": collision_reward,
                         "action_rate_reward": action_rate_reward,
                         "dof_pos_limits_reward": dof_pos_limits_reward,
                         "height_reward": height_reward}
    return total_reward, reward_components

 \end{mycode}
\end{tcolorbox}

\begin{tcolorbox}[colback=gray!10, colframe=black, title=Reward for Stairs, sharp corners, breakable]
 \begin{mycode}
def compute_reward(self):
    """ Compute improved rewards
        Compute each reward component first
        Then compute the total reward
        Return the total reward, and the recording of all reward components
    """
    env = self.env  # Do not skip this line. Afterwards, use env.{parameter_name} to access parameters of the environment.

    # Tracking of linear velocity commands (xy axes)
    lin_vel_error = torch.sum(torch.square(env.commands[:, :2] - env.base_lin_vel[:, :2]), dim=1)
    tracking_lin_vel_reward = 1.5 * torch.exp(-lin_vel_error / 0.20)

    # Tracking of angular velocity commands (yaw)
    ang_vel_error = torch.square(env.commands[:, 2] - env.base_ang_vel[:, 2])
    tracking_ang_vel_reward = 0.5 * torch.exp(-ang_vel_error / 0.1)

    # # Penalize z axis base linear velocity
    lin_vel_z_reward = -0.00001 * torch.square(env.base_lin_vel[:, 2])

    # Penalize xy axes base angular velocity
    ang_vel_xy_reward = -0.1 * torch.sum(torch.square(env.base_ang_vel[:, :2]), dim=1)

    # Penalize torques
    torques_reward = -0.0005 * torch.sum(torch.square(env.torques), dim=1)

    # Penalize dof accelerations
    dof_acc_reward = -1.0e-7 * torch.sum(torch.square((env.last_dof_vel - env.dof_vel) / env.dt), dim=1)

    # Reward air time
    contact = env.contact_forces[:, env.feet_indices, 2] > 1.
    contact_filt = torch.logical_or(contact, env.last_contacts)
    env.last_contacts = contact
    first_contact = (env.feet_air_time > 0.) * contact_filt
    env.feet_air_time += env.dt
    rew_airTime = torch.sum((env.feet_air_time - 0.5) * first_contact, dim=1)
    rew_airTime *= torch.norm(env.commands[:, :2], dim=1) > 0.1
    env.feet_air_time *= ~contact_filt
    feet_air_time_reward = 0.8 * rew_airTime

    # Penalize collisions
    collision_reward = -5.0 * torch.sum(1. * (torch.norm(env.contact_forces[:, env.penalised_contact_indices, :], dim=-1) > 0.1), dim=1)

    # Penalize changes in actions
    action_rate_reward = -0.008 * torch.sum(torch.square(env.last_actions - env.actions), dim=1)

    # Penalize dofs close to limits
    out_of_limits = -(env.dof_pos - env.dof_pos_limits[:, 0]).clip(max=0.)
    out_of_limits += (env.dof_pos - env.dof_pos_limits[:, 1]).clip(min=0.)
    dof_pos_limits_reward = -7.0 * torch.sum(out_of_limits, dim=1)

    # Penalize base height away from target
    target_height_z = 0.34
    base_height = env.root_states[:, 2]
    # get the ground height of the terrain
    ground_x = env.root_states[:, 0]
    ground_y = env.root_states[:, 1]
    ground_z = env._get_stairs_terrain_heights(ground_x, ground_y)
    # calculate the base-to-ground height
    base2ground_height = base_height - ground_z
    height_reward = -0.000002 * torch.square(base2ground_height - target_height_z)

    # stumbling penalty
    stumble = (torch.norm(env.contact_forces[:, env.feet_indices, :2], dim=2) > 5.) * (torch.abs(env.contact_forces[:, env.feet_indices, 2]) < 1.)
    stumble_reward = -2.0 * torch.sum(stumble, dim=1)

    # Combine reward components to compute the total reward in this step
    total_reward = (tracking_lin_vel_reward + tracking_ang_vel_reward + lin_vel_z_reward +
                    ang_vel_xy_reward + torques_reward + dof_acc_reward + feet_air_time_reward +
                    collision_reward + action_rate_reward + dof_pos_limits_reward + height_reward + stumble_reward)

    # Debug information
    reward_components = {"tracking_lin_vel_reward": tracking_lin_vel_reward,
                         "tracking_ang_vel_reward": tracking_ang_vel_reward,
                         "lin_vel_z_reward": lin_vel_z_reward,
                         "ang_vel_xy_reward": ang_vel_xy_reward,
                         "torques_reward": torques_reward,
                         "dof_acc_reward": dof_acc_reward,
                         "feet_air_time_reward": feet_air_time_reward,
                         "collision_reward": collision_reward,
                         "action_rate_reward": action_rate_reward,
                         "dof_pos_limits_reward": dof_pos_limits_reward,
                         "height_reward": height_reward,
                         "stumble_reward": stumble_reward}

    return total_reward, reward_components

  \end{mycode}
\end{tcolorbox}

\begin{tcolorbox}[colback=gray!10, colframe=black, title=Reward for Obstacles, sharp corners, breakable]
 \begin{mycode}

def compute_reward(self):
    """ Compute rewards for wave terrain """
    env = self.env

    # 1. Linear velocity tracking along x-axis
    lin_vel_error = torch.sum(torch.square(env.commands[:, :2] - env.base_lin_vel[:, :2]), dim=1)
    tracking_lin_vel_reward = 2.3 * torch.exp(-lin_vel_error / 0.20)

    # 2. Tracking of angular velocity commands (yaw)
    ang_vel_error = torch.square(env.commands[:, 2] - env.base_ang_vel[:, 2])
    tracking_ang_vel_reward = 0.8 * torch.exp(-ang_vel_error / 0.1)

    # 3. Penalize z-axis velocity
    # lin_vel_z_reward = -0.001 * torch.square(env.base_lin_vel[:, 2])
    ang_vel_x_reward = -0.002 * torch.square(env.base_ang_vel[:, 0])

    # 4. Base height tracking (adjusted for wave terrain)
    target_height_z = 0.34
    base_height = env.root_states[:, 2]
    ground_x = env.root_states[:, 0]
    ground_y = env.root_states[:, 1]
    ground_z = env._get_terrain_heights(ground_x, ground_y)
    base2ground_height = base_height - ground_z
    height_reward = -1.5 * torch.square(base2ground_height - target_height_z)

    # 5. Penalize torques
    torques_reward = -0.00001 * torch.sum(torch.square(env.torques), dim=1)

    # 6. Penalize changes in actions
    action_rate_reward = -0.0055 * torch.sum(torch.square(env.last_actions - env.actions), dim=1)

    # 7. Encourage smoother joint motions (penalize excessive joint accelerations)
    dof_acc_penalty = -1e-8 * torch.sum(torch.square((env.dof_vel - env.last_dof_vel) / env.dt), dim=1)

    # 8. Air time reward for dynamic gaits
    contact = env.contact_forces[:, env.feet_indices, 2] > 1.0
    contact_filt = torch.logical_or(contact, env.last_contacts)
    env.last_contacts = contact
    first_contact = (env.feet_air_time > 0.0) * contact_filt
    env.feet_air_time += env.dt
    rew_airTime = torch.sum((env.feet_air_time - 0.4) * first_contact, dim=1)
    rew_airTime *= torch.norm(env.commands[:, :2], dim=1) > 0.1
    env.feet_air_time *= ~contact_filt
    air_time_reward = 0.5 * rew_airTime

    # 9. Collision penalty (avoid collisions with terrain or robot parts)
    collision_penalty = -1. * torch.sum(
        1.0 * (torch.norm(env.contact_forces[:, env.penalised_contact_indices, :], dim=-1) > 0.13),
        dim=1
    )

    # 10. Gait pattern reward (encourage trot gait using phase alignment)
    diag_sync = (contact[:, 0] == contact[:, 3]) & (contact[:, 1] == contact[:, 2])
    gait_pattern_reward = 0.0001 * torch.sum(diag_sync.float())

    # 11. Penalize use only two feet
    stumble = (torch.norm(env.contact_forces[:, env.feet_indices, :2], dim=2) > 5.) * (torch.abs(env.contact_forces[:, env.feet_indices, 2]) < 1.)
    stumble_penalty = -20.0 * torch.sum(stumble, dim=1)

    # Combine all components into the total reward
    total_reward = (
        tracking_lin_vel_reward +
        tracking_ang_vel_reward +
        ang_vel_x_reward +
        height_reward +
        torques_reward +
        action_rate_reward +
        dof_acc_penalty +
        air_time_reward +
        collision_penalty +
        gait_pattern_reward +
        stumble_penalty
    )

    # Debug information for reward components
    reward_components = {
        "tracking_lin_vel_reward": tracking_lin_vel_reward,
        "tracking_ang_vel_reward": tracking_ang_vel_reward,
        "ang_vel_x_reward": ang_vel_x_reward,
        "height_reward": height_reward,
        "torques_reward": torques_reward,
        "action_rate_reward": action_rate_reward,
        "dof_acc_penalty": dof_acc_penalty,
        "air_time_reward": air_time_reward,
        "collision_penalty": collision_penalty,
        "gait_pattern_reward": gait_pattern_reward,
        "stumble_penalty": stumble_penalty
    }

    return total_reward, reward_components

  \end{mycode}
\end{tcolorbox}

\begin{tcolorbox}[colback=gray!10, colframe=black, title=Reward for Wave, sharp corners, breakable]
\begin{mycode}
def compute_reward(self):
    """ Compute rewards for wave terrain """
    env = self.env

    # 1. Tracking of linear velocity commands (xy axes)
    lin_vel_error = torch.sum(torch.square(env.commands[:, :2] - env.base_lin_vel[:, :2]), dim=1)
    tracking_lin_vel_reward = 2.3 * torch.exp(-lin_vel_error / 0.20)

    # 2. Tracking of angular velocity commands (yaw)
    ang_vel_error = torch.square(env.commands[:, 2] - env.base_ang_vel[:, 2])
    tracking_ang_vel_reward = 0.8 * torch.exp(-ang_vel_error / 0.1)

    # 3. Base height tracking (adjusted for wave terrain)
    target_height_z = 0.34
    base_height = env.root_states[:, 2]
    ground_x = env.root_states[:, 0]
    ground_y = env.root_states[:, 1]
    ground_z = env._get_terrain_heights(ground_x, ground_y)
    base2ground_height = base_height - ground_z
    height_reward = -1.5 * torch.square(base2ground_height - target_height_z)

    # 4. Penalize torques
    torques_reward = -0.00001 * torch.sum(torch.square(env.torques), dim=1)

    # 5. Penalize changes in actions
    action_rate_reward = -0.0045 * torch.sum(torch.square(env.last_actions - env.actions), dim=1)

    # 6. Encourage smoother joint motions (penalize excessive joint accelerations)
    dof_acc_penalty = -1e-8 * torch.sum(torch.square((env.dof_vel - env.last_dof_vel) / env.dt), dim=1)

    # 7. Air time reward for dynamic gaits
    contact = env.contact_forces[:, env.feet_indices, 2] > 1.0
    contact_filt = torch.logical_or(contact, env.last_contacts)
    env.last_contacts = contact
    first_contact = (env.feet_air_time > 0.0) * contact_filt
    env.feet_air_time += env.dt
    rew_airTime = torch.sum((env.feet_air_time - 0.4) * first_contact, dim=1)
    rew_airTime *= torch.norm(env.commands[:, :2], dim=1) > 0.1
    env.feet_air_time *= ~contact_filt
    air_time_reward = 0.5 * rew_airTime

    # 8. Collision penalty (avoid collisions with terrain or robot parts)
    collision_penalty = -1. * torch.sum(
        1.0 * (torch.norm(env.contact_forces[:, env.penalised_contact_indices, :], dim=-1) > 0.13),
        dim=1
    )

    # 9. Gait pattern reward (encourage trot gait using phase alignment)
    diag_sync = (contact[:, 0] == contact[:, 3]) & (contact[:, 1] == contact[:, 2])
    gait_pattern_reward = 0.0001 * torch.sum(diag_sync.float())

    # 10. Penalize use only two feet
    lack_of_foot_usage = (~contact).float().sum(dim=1) 
    lack_of_foot_usage_penalty = -0.01 * lack_of_foot_usage
    
    # Combine all components into the total reward
    total_reward = (
        tracking_lin_vel_reward +
        tracking_ang_vel_reward +
        height_reward +
        torques_reward +
        action_rate_reward +
        dof_acc_penalty +
        air_time_reward +
        collision_penalty +
        gait_pattern_reward +
        lack_of_foot_usage_penalty
    )

    # Debug information for reward components
    reward_components = {
        "tracking_lin_vel_reward": tracking_lin_vel_reward,
        "tracking_ang_vel_reward": tracking_ang_vel_reward,
        "height_reward": height_reward,
        "torques_reward": torques_reward,
        "action_rate_reward": action_rate_reward,
        "dof_acc_penalty": dof_acc_penalty,
        "air_time_reward": air_time_reward,
        "collision_penalty": collision_penalty,
        "gait_pattern_reward": gait_pattern_reward,
        "lack_of_foot_usage_penalty": lack_of_foot_usage_penalty
    }

    return total_reward, reward_components
\end{mycode}
\end{tcolorbox}

\begin{tcolorbox}[colback=gray!10, colframe=black, title=Reward for Slope, sharp corners, breakable]
\begin{mycode}
    def compute_reward(self):
        """ Compute improved rewards
            Compute each reward component first
            Then compute the total reward
            Return the total reward, and the recording of all reward components
        """
        env = self.env  # Do not skip this line. Afterwards, use env.{parameter_name} to access parameters of the environment.

        # Tracking of linear velocity commands (xy axes)
        lin_vel_error = torch.sum(torch.square(env.commands[:, :2] - env.base_lin_vel[:, :2]), dim=1)
        tracking_lin_vel_reward = 3.0 * torch.exp(-lin_vel_error / 0.10)

        # Tracking of angular velocity commands (yaw)
        ang_vel_error = torch.square(env.commands[:, 2] - env.base_ang_vel[:, 2])
        tracking_ang_vel_reward = 1.0 * torch.exp(-ang_vel_error / 0.05)

        # # Penalize z axis base linear velocity
        lin_vel_z_reward = -0.00001 * torch.square(env.base_lin_vel[:, 2])

        # Penalize xy axes base angular velocity
        ang_vel_xy_reward = -0.1 * torch.sum(torch.square(env.base_ang_vel[:, :2]), dim=1)

        # Penalize torques
        torques_reward = -0.0001 * torch.sum(torch.square(env.torques), dim=1)

        # Penalize dof accelerations
        dof_acc_reward = -5.0e-8 * torch.sum(torch.square((env.last_dof_vel - env.dof_vel) / env.dt), dim=1)

        # Reward air time
        contact = env.contact_forces[:, env.feet_indices, 2] > 1.
        contact_filt = torch.logical_or(contact, env.last_contacts)
        env.last_contacts = contact
        first_contact = (env.feet_air_time > 0.) * contact_filt
        env.feet_air_time += env.dt
        rew_airTime = torch.sum((env.feet_air_time - 0.5) * first_contact, dim=1)
        rew_airTime *= torch.norm(env.commands[:, :2], dim=1) > 0.1
        env.feet_air_time *= ~contact_filt
        feet_air_time_reward = 0.6 * rew_airTime

        # Penalize collisions
        collision_reward = -15.0 * torch.sum(1. * (torch.norm(env.contact_forces[:, env.penalised_contact_indices, :], dim=-1) > 0.1), dim=1)

        # Penalize changes in actions
        action_rate_reward = -0.015 * torch.sum(torch.square(env.last_actions - env.actions), dim=1)

        # Penalize dofs close to limits
        out_of_limits = -(env.dof_pos - env.dof_pos_limits[:, 0]).clip(max=0.)
        out_of_limits += (env.dof_pos - env.dof_pos_limits[:, 1]).clip(min=0.)
        dof_pos_limits_reward = -6.0 * torch.sum(out_of_limits, dim=1)

        # Penalize base height away from target
        target_height_z = 0.34
        base_height = env.root_states[:, 2]
        # get the ground height of the terrain
        ground_x = env.root_states[:, 0]
        ground_y = env.root_states[:, 1]
        ground_z = env._get_terrain_heights(ground_x, ground_y)
        # calculate the base-to-ground height
        base2ground_height = base_height - ground_z
        height_reward = -0.000001 * torch.square(base2ground_height - target_height_z)

        # stumbling penalty
        # stumble = (torch.norm(env.contact_forces[:, env.feet_indices, :2], dim=2) > 5.) * (torch.abs(env.contact_forces[:, env.feet_indices, 2]) < 1.)
        # stumble_reward = -4.0 * torch.sum(stumble, dim=1)

        # Combine reward components to compute the total reward in this step
        total_reward = (tracking_lin_vel_reward + tracking_ang_vel_reward + lin_vel_z_reward +
                        ang_vel_xy_reward + torques_reward + dof_acc_reward + feet_air_time_reward +
                        collision_reward + action_rate_reward + dof_pos_limits_reward + height_reward)
                        # + stumble_reward)

        # Debug information
        reward_components = {"tracking_lin_vel_reward": tracking_lin_vel_reward,
                             "tracking_ang_vel_reward": tracking_ang_vel_reward,
                             "lin_vel_z_reward": lin_vel_z_reward,
                             "ang_vel_xy_reward": ang_vel_xy_reward,
                             "torques_reward": torques_reward,
                             "dof_acc_reward": dof_acc_reward,
                             "feet_air_time_reward": feet_air_time_reward,
                             "collision_reward": collision_reward,
                             "action_rate_reward": action_rate_reward,
                             "dof_pos_limits_reward": dof_pos_limits_reward,
                             "height_reward": height_reward,}

        return total_reward, reward_components
\end{mycode}
\end{tcolorbox}

\begin{tcolorbox}[colback=gray!10, colframe=black, title=Reward for Jump high, sharp corners, breakable]
\begin{mycode}
def compute_reward(self):
    """ Compute rewards for the backflip task """
    env = self.env  # Do not skip this line. Afterwards, use env.{parameter_name} to access parameters of the environment.

    # Penalize angular velocity around x-axis (roll) and z-axis (yaw)
    roll_rate = env.root_states[:, 10]  # Angular velocity around x-axis (roll)
    yaw_rate = env.root_states[:, 12]  # Angular velocity around z-axis (yaw)
    roll_rate_reward = -0.05 * torch.square(roll_rate)  # Penalize deviation in roll, org -5.0, 0.5
    yaw_rate_reward = -0.2 * torch.square(yaw_rate)  # Penalize deviation in yaw, org -5.0, -2.0

    # Penalize roll and yaw around x-axis (roll) and z-axis (yaw)
    roll = env.rpy[:, 0]
    pitch = env.rpy[:, 1]
    yaw = env.rpy[:, 2]
    roll_reward = -0.2 * torch.square(roll)  # Penalize deviation in roll
    pitch_reward = -0.2 * torch.square(pitch)  # Penalize deviation in pitch
    yaw_reward = -1.0 * torch.square(yaw)  # Penalize deviation in yaw

    # Encourage z axis base linear velocity
    lin_vel_z_reward = 0.4 * torch.square(env.root_states[:, 9])

    # Penalize x axis base linear velocity forward
    lin_vel_x_reward  = -0.4 * torch.square(torch.relu(env.root_states[:, 7]))

    # Penalize torques
    torques_reward = -0.0005 * torch.sum(torch.square(env.torques), dim=1)

    # Penalize dof accelerations
    dof_acc_reward = -1.0e-7 * torch.sum(torch.square((env.last_dof_vel - env.dof_vel) / env.dt), dim=1)

    # # Penalize dofs close to limits
    # out_of_limits = -(env.dof_pos - env.dof_pos_limits[:, 0]).clip(max=0.)
    # out_of_limits += (env.dof_pos - env.dof_pos_limits[:, 1]).clip(min=0.)
    # dof_pos_limits_reward = -7.0 * torch.sum(out_of_limits, dim=1)

    # Penalize base torso hitting the ground
    base_hit_ground = torch.any(torch.norm(env.contact_forces[:, env.termination_contact_indices, :], dim=-1) > 0.1, dim=1)  # >1.0 originally
    base_hit_ground_reward = -20.0 * base_hit_ground  # 10

    # Penalize non flat base orientation
    gravity_reward = -0.01 * torch.sum(torch.square(env.projected_gravity[:, :2]), dim=1)

    # Reward for highest z position reached in an episode
    height_history = env.height_history_buf
    height_history_reward = 1.0 * height_history  # 0.5

    # Combine reward components to compute the total reward in this step
    total_reward = (roll_rate_reward + yaw_rate_reward + roll_reward + pitch_reward + yaw_reward + lin_vel_z_reward + lin_vel_x_reward +
                    torques_reward + dof_acc_reward+ base_hit_ground_reward + gravity_reward + height_history_reward)


    # Debug information
    reward_components = {
        "roll_rate_reward": roll_rate_reward,
        "yaw_rate_reward": yaw_rate_reward,
        "roll_reward": roll_reward,
        "pitch_reward": pitch_reward,
        "yaw_reward": yaw_reward,
        "lin_vel_z_reward": lin_vel_z_reward,
        "lin_vel_x_reward": lin_vel_x_reward,
        "torques_reward": torques_reward,
        "dof_acc_reward": dof_acc_reward,
        "base_hit_ground_reward": base_hit_ground_reward,
        "gravity_reward": gravity_reward,
        "height_history_reward": height_history_reward
    }
    return total_reward, reward_components
\end{mycode}
\end{tcolorbox}

\begin{tcolorbox}[colback=gray!10, colframe=black, title=Reward for Backflip, sharp corners, breakable]
\begin{mycode}
def compute_reward(self):
    """ Compute rewards for the backflip task """
    env = self.env  # Do not skip this line. Afterwards, use env.{parameter_name} to access parameters of the environment.

    # Penalize angular velocity around x-axis (roll) and z-axis (yaw)
    roll_rate = env.root_states[:, 10]  # Angular velocity around x-axis (roll)
    yaw_rate = env.root_states[:, 12]  # Angular velocity around z-axis (yaw)
    roll_rate_reward = -0.05 * torch.square(roll_rate)  # Penalize deviation in roll, org -5.0, 0.5
    yaw_rate_reward = -0.2 * torch.square(yaw_rate)  # Penalize deviation in yaw, org -5.0, -2.0

    # Penalize roll and yaw around x-axis (roll) and z-axis (yaw)
    roll = env.rpy[:, 0]
    yaw = env.rpy[:, 2]
    roll_reward = -0.2 * torch.square(roll)  # Penalize deviation in roll
    yaw_reward = -1.0 * torch.square(yaw)  # Penalize deviation in yaw

    # Reward for angular velocity around y-axis (pitch), encourage backflip rotation
    pitch_rate = env.base_ang_vel[:, 1]  # Angular velocity around y-axis (pitch)
    # Clamp the pitch rate to a minimum of -7 rad/s (maximum negative rotation speed)
    clamped_pitch_rate = torch.clamp(pitch_rate, max=5.0, min=-7.0)
    # Compute the backflip reward
    pitch_rate_reward = torch.where(
        clamped_pitch_rate < 0,
        1.0 * (-clamped_pitch_rate),  # Positive reward for negative pitch rate
        -0.1 * clamped_pitch_rate  # Negative penalty for positive pitch rate
    )

    # pitch_rate_reward = 1.0 * (-clamped_pitch_rate)  # Reward negative pitch rates

    # Reward for pitch frontwards for the back flip
    pitch = env.rpy[:, 1]
    last_pitch = env.last_rpy[:, 1]
    delta_pitch = (pitch - last_pitch + torch.pi) % (2 * torch.pi) - torch.pi
    delta_pitch_reward = torch.where(delta_pitch > 0, delta_pitch, torch.zeros_like(delta_pitch))
    delta_pitch_reward = 20.0 * delta_pitch_reward

    # Encourage z axis base linear velocity
    lin_vel_z_reward = 0.2 * torch.square(env.root_states[:, 9])

    # Penalize torques
    torques_reward = -0.0005 * torch.sum(torch.square(env.torques), dim=1)

    # Penalize dof accelerations
    dof_acc_reward = -1.0e-7 * torch.sum(torch.square((env.last_dof_vel - env.dof_vel) / env.dt), dim=1)

    # # Penalize dofs close to limits
    # out_of_limits = -(env.dof_pos - env.dof_pos_limits[:, 0]).clip(max=0.)
    # out_of_limits += (env.dof_pos - env.dof_pos_limits[:, 1]).clip(min=0.)
    # dof_pos_limits_reward = -7.0 * torch.sum(out_of_limits, dim=1)

    # Penalize base torso hitting the ground
    base_hit_ground = torch.any(torch.norm(env.contact_forces[:, env.termination_contact_indices, :], dim=-1) > 0.1, dim=1)  # >1.0 originally
    base_hit_ground_reward = -20.0 * base_hit_ground  # 10

    # Penalize projected gravity along y axis
    gravity_reward = -0.01 * torch.square(env.projected_gravity[:, 1])

    # Encourage backwards rotation of the projected gravity
    delta_gravity_x = env.projected_gravity[:, 0] - env.last_projected_gravity[:, 0]
    condition = env.projected_gravity[:, 2] < 0  # Corrected to use the z-axis as per your description
    desired_direction = torch.where(
        condition,
        -1.0,  # Encourage decrease in delta_gravity_x
        1.0  # Encourage increase in delta_gravity_x
    )

    # Compute the alignment between desired and actual change
    alignment = desired_direction * delta_gravity_x

    # Compute the reward with different scaling factors
    rotate_gravity_reward = torch.where(
        alignment > 0,
        20.0 * torch.abs(delta_gravity_x),  # Positive reward
        -2.0 * torch.abs(delta_gravity_x)  # Negative penalty
    )

    # Reward for highest z position reached in an episode
    height_history = env.height_history_buf
    height_history_reward = 0.8 * height_history  # 0.5

    # Combine reward components to compute the total reward in this step
    total_reward = (roll_rate_reward + yaw_rate_reward + roll_reward + yaw_reward + pitch_rate_reward +
                    delta_pitch_reward + lin_vel_z_reward + torques_reward + dof_acc_reward+
                    base_hit_ground_reward + gravity_reward + rotate_gravity_reward + height_history_reward)


    # Debug information
    reward_components = {
        "roll_rate_reward": roll_rate_reward,
        "yaw_rate_reward": yaw_rate_reward,
        "roll_reward": roll_reward,
        "yaw_reward": yaw_reward,
        "pitch_rate_reward": pitch_rate_reward,
        "delta_pitch_reward": delta_pitch_reward,
        "lin_vel_z_reward": lin_vel_z_reward,
        "torques_reward": torques_reward,
        "dof_acc_reward": dof_acc_reward,
        "base_hit_ground_reward": base_hit_ground_reward,
        "gravity_reward": gravity_reward,
        "rotate_gravity_reward": rotate_gravity_reward,
        "height_history_reward": height_history_reward
    }
    return total_reward, reward_components

\end{mycode}
\end{tcolorbox}

\begin{tcolorbox}[colback=gray!10, colframe=black, title=Reward for Kettle, sharp corners, breakable]
\begin{mycode}
def kettle_compute_reward(kettle_spout_pos: torch.Tensor,
                          bucket_handle_pos: torch.Tensor,
                          left_hand_pos: torch.Tensor,
                          right_hand_pos: torch.Tensor,
                          left_hand_ff_pos: torch.Tensor,
                          right_hand_ff_pos: torch.Tensor
) -> Tuple[torch.Tensor, Dict[str, torch.Tensor]]:

    # Redefine proximity reward for less dominance
    kettle_to_bucket_distance = torch.norm(kettle_spout_pos - bucket_handle_pos, dim=1)
    temp_proximity = 1.0
    proximity_reward = torch.exp(-kettle_to_bucket_distance / temp_proximity)

    # Rescale grip rewards and refine detection logic
    left_grip_distance = torch.norm(left_hand_ff_pos - kettle_spout_pos, dim=1)
    right_grip_distance = torch.norm(right_hand_ff_pos - bucket_handle_pos, dim=1)
    
    temp_grip = 0.5
    left_grip_reward = torch.exp(-left_grip_distance / temp_grip)
    right_grip_reward = torch.exp(-right_grip_distance / temp_grip)

    # Improved task-specific reward detection logic
    task_success_indicator = torch.tensor([0], device=kettle_spout_pos.device)  # Replace with task success condition
    task_completion_threshold = 0.1  # Define a threshold to reflect task completion
    task_complete = kettle_to_bucket_distance < task_completion_threshold
    transformed_task_score = task_complete.float()

    # Total reward balancing individual components
    total_reward = (
        0.2 * proximity_reward +
        0.3 * left_grip_reward +
        0.3 * right_grip_reward +
        1.0 * transformed_task_score
    )

    # Reward components provided for diagnosis
    reward_components = {
        "kettle_spout_proximity": proximity_reward,
        "kettle_handle_grip": left_grip_reward,
        "bucket_handle_grip": right_grip_reward,
        "task_success": transformed_task_score
    }

    return total_reward, reward_components

\end{mycode}
\end{tcolorbox}

\begin{tcolorbox}[colback=gray!10, colframe=black, title=Reward for Hand Over, sharp corners, breakable]
\begin{mycode}
def hand_over_compute_reward(object_pos: torch.Tensor, 
                             object_rot: torch.Tensor, 
                             object_linvel: torch.Tensor,
                             goal_pos: torch.Tensor, 
                             goal_rot: torch.Tensor,
                             fingertip_pos: torch.Tensor, 
                             fingertip_another_pos: torch.Tensor
) -> Tuple[torch.Tensor, Dict[str, torch.Tensor]]:
    # Constants
    distance_threshold: float = 0.03
    rotation_threshold: float = 0.1
    catch_reward_weight: float = 30.0
    toss_reward_weight: float = 10.0
    toss_temperature: float = 1.5
    catch_temperature: float = 0.5
    penalty_weight: float = 10.0

    # Compute distance and rotation differences between object and goal
    object_goal_distance = torch.norm(object_pos - goal_pos, dim=-1)
    object_goal_rotation_diff = torch.norm(object_rot - goal_rot, dim=-1)

    # Reward for object being close to the goal position and rotation
    toss_reward = torch.exp(-toss_reward_weight * (object_goal_distance / toss_temperature))

    # Calculate the catch reward based on the difference between the object's linear velocity and the goal position
    catch_reward = torch.norm(goal_pos - object_linvel, dim=-1)
    catch_reward = torch.exp(-catch_reward_weight * (catch_reward / catch_temperature))

    # Penalty for direct rolling or touching the target instead of tossing and catching
    penalty = torch.zeros_like(object_goal_distance)
    for i in range(fingertip_pos.shape[1]):
        dist_to_fingertip = torch.norm(object_pos - fingertip_pos[:, i, :], dim=-1)
        dist_to_fingertip_another = torch.norm(object_pos - fingertip_another_pos[:, i, :], dim=-1)
        penalty = torch.where(
            (dist_to_fingertip < distance_threshold) | (dist_to_fingertip_another < distance_threshold),
            penalty + 1.0, penalty
        )

    penalty_ratio = torch.sigmoid(penalty * penalty_weight) * 0.5
    penalty = catch_reward * penalty_ratio  # Adapt the penalty to the catch_reward's dynamic range

    # Total reward
    total_reward = toss_reward + catch_reward - penalty
    reward_terms = {"toss_reward": toss_reward, "catch_reward": catch_reward, "penalty": penalty}

    return total_reward, reward_terms

\end{mycode}
\end{tcolorbox}

\begin{tcolorbox}[colback=gray!10, colframe=black, title=Reward for Swing Cup, sharp corners, breakable]
\begin{mycode}
def swing_cup_compute_reward(object_pos: torch.Tensor, 
                             object_rot: torch.Tensor, 
                             object_linvel: torch.Tensor, 
                             cup_right_handle_pos: torch.Tensor, 
                             cup_left_handle_pos: torch.Tensor, 
                             left_hand_pos: torch.Tensor, 
                             right_hand_pos: torch.Tensor, 
                             goal_pos: torch.Tensor, 
                             goal_rot: torch.Tensor
    ) -> Tuple[torch.Tensor, Dict[str, torch.Tensor]]:
    object_goal_distance = torch.norm(object_pos - goal_pos, dim=-1)
    distance_reward_temperature = 0.1
    object_goal_distance_reward_weight = 0.1
    object_goal_distance_reward = -torch.exp(-distance_reward_temperature * object_goal_distance) * object_goal_distance_reward_weight

    right_cup_handle_dist = torch.norm(cup_right_handle_pos - right_hand_pos, dim=-1)
    left_cup_handle_dist = torch.norm(cup_left_handle_pos - left_hand_pos, dim=-1)

    cup_orientation_diff = 1 - torch.sum(torch.mul(object_rot, goal_rot), dim=-1) ** 2
    cup_orientation_reward_weight = 1.
    cup_orientation_reward = -(cup_orientation_diff * cup_orientation_reward_weight)

    grasp_temperature_1 = 0.25
    grasp_temperature_2 = 0.25
    right_grasp_reward = torch.exp(-grasp_temperature_1 * right_cup_handle_dist)
    left_grasp_reward = torch.exp(-grasp_temperature_2 * left_cup_handle_dist)
    grasp_reward = (right_grasp_reward + left_grasp_reward - 1.0)

    cup_linvel_norm = torch.norm(object_linvel, dim=-1)
    cup_linvel_penalty_weight = 0.2
    cup_linvel_penalty = -(cup_linvel_norm * cup_linvel_penalty_weight)

    touch_reward_temperature = 0.25
    touch_reward_weight = 0.125
    touch_reward = (torch.exp(-touch_reward_temperature * right_cup_handle_dist) + torch.exp(-touch_reward_temperature * left_cup_handle_dist) - 1.0) * touch_reward_weight

    total_reward = grasp_reward + object_goal_distance_reward + cup_orientation_reward + cup_linvel_penalty + touch_reward

    reward_dict = {
        "grasp_reward": grasp_reward,
        "object_goal_distance_reward": object_goal_distance_reward,
        "cup_orientation_reward": cup_orientation_reward,
        "cup_linvel_penalty": cup_linvel_penalty,
        "touch_reward": touch_reward
    }

    return total_reward, reward_dict

\end{mycode}
\end{tcolorbox}





\end{document}